\documentclass{ieeeaccess}
\usepackage{pifont}
\newcommand{\xmark}{\ding{55}}
\newcommand{\vmark}{\ding{51}}
\usepackage{cite}
\usepackage{amsmath,amssymb,amsfonts}
\usepackage{algpseudocode}
\usepackage{graphicx}
\usepackage{subcaption}
\usepackage{caption,setspace}
\usepackage{textcomp}
\usepackage{listings}
\usepackage{longtable}
\usepackage{array,ragged2e}
\newcolumntype{P}[1]{>{\RaggedRight\arraybackslash}p{#1}}
\usepackage[table]{xcolor}

\definecolor{lightyellow}{rgb}{1, 1, 0.75}
\definecolor{green}{rgb}{0, 1, 0}

\newcommand{\comment}[1]{}
\newcommand{\etal}{\textit{et al. }}

\newcommand{\COMMENT}[1]{$\triangleright$  #1}

\def\BibTeX{{\rm B\kern-.05em{\sc i\kern-.025em b}\kern-.08em
    T\kern-.1667em\lower.7ex\hbox{E}\kern-.125emX}}
\begin{document}
\history{Date of publication xxxx 00, 0000, date of current version xxxx 00, 0000.}
\doi{10.1109/ACCESS.2017.DOI}

\title{Fast Low-parameter Video Activity Localization in Collaborative Learning Environments}
\author{
  \uppercase{Venkatesh Jatla}\authorrefmark{1},
  \uppercase{Sravani Teeparthi}\authorrefmark{1},
  \uppercase{Ugesh Egala}\authorrefmark{1},
  \uppercase{Sylvia Celed\'{o}n Pattichis}\authorrefmark{2}, and
  \uppercase{Marios S. Pattichis}\authorrefmark{1}, \IEEEmembership{Senior Member, IEEE}
}
\address[1]{
  Image and Video Processing and Communications Lab,
  Dept. of Electrical and Computer Engineering,
  University of New Mexico,
  Albuquerque, NM 87131, USA
  (email: venkatesh369@unm.edu, pattichi@unm.edu, steeparthi@unm.edu, ugeshe@unm.edu)
}
\address[2]{
  Department of Curriculum and Instruction,
  The University of Texas at Austin,
  Austin,
  Texas, 78712-1293
  (email: sylvia.celedon@austin.utexas.edu)
}

\tfootnote{This work was supported in part by the National Science Foundation under Grant No.1949230}

\markboth
{Author \headeretal: Preparation of Papers for IEEE TRANSACTIONS and JOURNALS}
{Author \headeretal: Preparation of Papers for IEEE TRANSACTIONS and JOURNALS}

\corresp{Corresponding author: Prof. Marios S. Pattichis (e-mail: pattichi@unm.edu).}

\begin{abstract}
  Research on video activity detection has primarily focused on
  identifying well-defined human activities in short video segments.
  The majority of the research on video activity recognition is
  focused on the development of large parameter systems that require
  training on large video datasets.  This paper develops a
  low-parameter, modular system with rapid inferencing capabilities
  that can be trained entirely on limited datasets without requiring
  transfer learning from large-parameter systems.  The system can
  accurately detect and associate specific activities with the
  students who perform the activities in real-life classroom videos.
  Additionally, the paper develops an interactive web-based
  application to visualize human activity maps over long real-life
  classroom videos.
  
  Long-term video activity detection in real-life classroom videos
  present unique challenges, such as the need to detect multiple
  simultaneous activities, rapid transitions between activities,
  long-term occlusions, durations exceeding 15 minutes, and numerous
  individuals performing similar activities in the background.
  Moreover, subtle hand movements further complicate the need to
  differentiate between actual typing and writing activities as opposed
  to unrelated hand movements.
  
  The system processes the input videos using fast activity
  initializations and current methods for object detection to
  determine the location and the the person performing the activities.
  These regions are then processed through an optimal low-parameter
  dyadic 3D-CNN classifier to identify the activity.  The proposed
  system processes 1 hour of video in 15 minutes for typing and 50
  minutes for writing activities.
  
  The system uses several methods to optimize the inference pipeline.
  For each activity, the system determines an optimal low-parameter 3D
  CNN architecture selected from a family of low-parameter
  architectures.  The input video is broken into smaller video regions
  that are transcoded at an optimized frame rate.  For inference, an
  optimal batch size is determined for processing input videos faster.
  Overall, the low-parameter separable activity classification model
  uses just 18.7K parameters, requiring 136.32 MB of memory and running
  at 4,620 (154 x 30) frames per second.  Compared to current methods,
  the approach used at least 1,000 fewer parameters and 20 times less
  GPU memory, while outperforming in both inference speed and
  classification accuracy.
\end{abstract}


\titlepgskip=-15pt

\maketitle

\section{Introduction}
Over the past decade, considerable advancements have been made in
detecting a limited number of well-defined human activities in videos,
with deep learning techniques enabling accurate detection and
classification of these activities \cite{yao2019review}. This research
has primarily benefited large corporations like YouTube \cite{yt-8m},
as it allows them to effectively organize and enhance their video
recommendation algorithms. However, despite their successess, these
methods still face challenges, including (i) connecting activities to
the person performing them over extended periods, (ii) detecting
activities with limited training data, (iii) establishing a fast
inferencing pipeline for activity detection, and (iv) presenting
detected activities in a practical and interactive manner that helps
users identify important events based on the observed activities.

\begin{figure*}
  \centering  

  \begin{subfigure}{0.47\textwidth}
    \includegraphics[width=\linewidth]{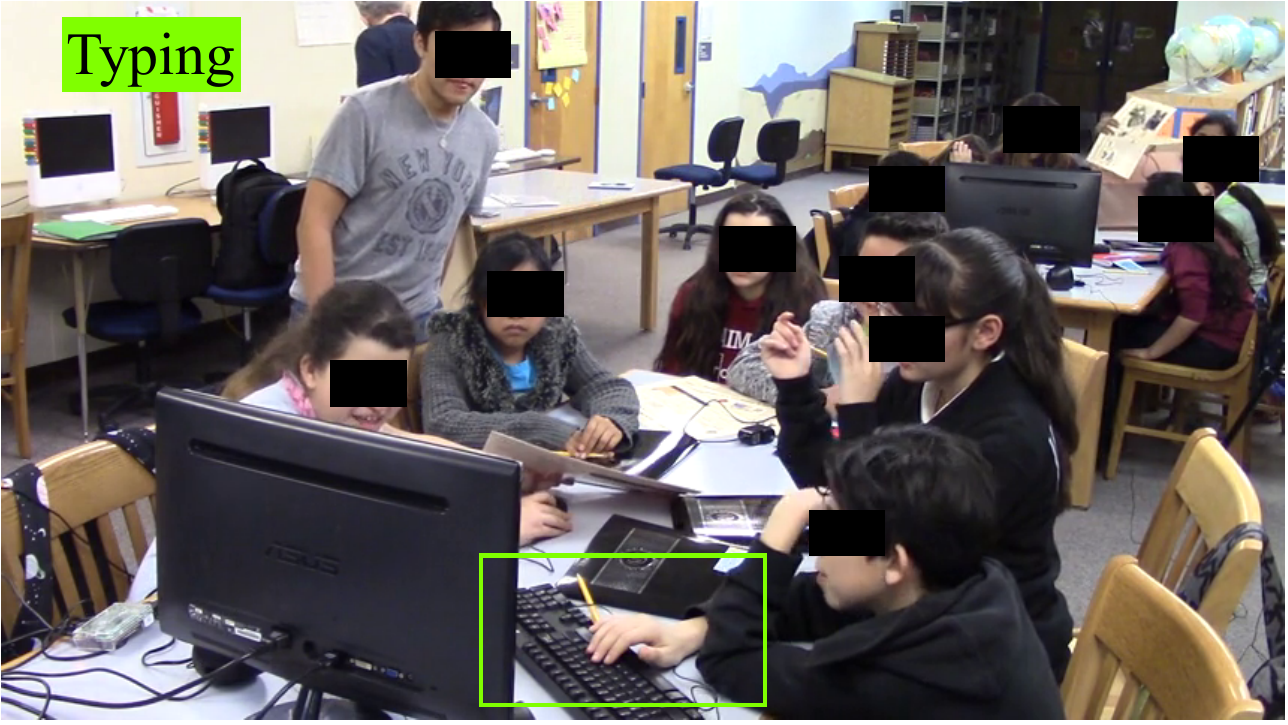}
    \caption{Typing activity in collaborative learning environment. The keyboard is partially visible and the students very close to each other.}
    \label{subfig:ty-det}
  \end{subfigure}~~  
  \begin{subfigure}{0.47\textwidth}
    \includegraphics[width=\linewidth]{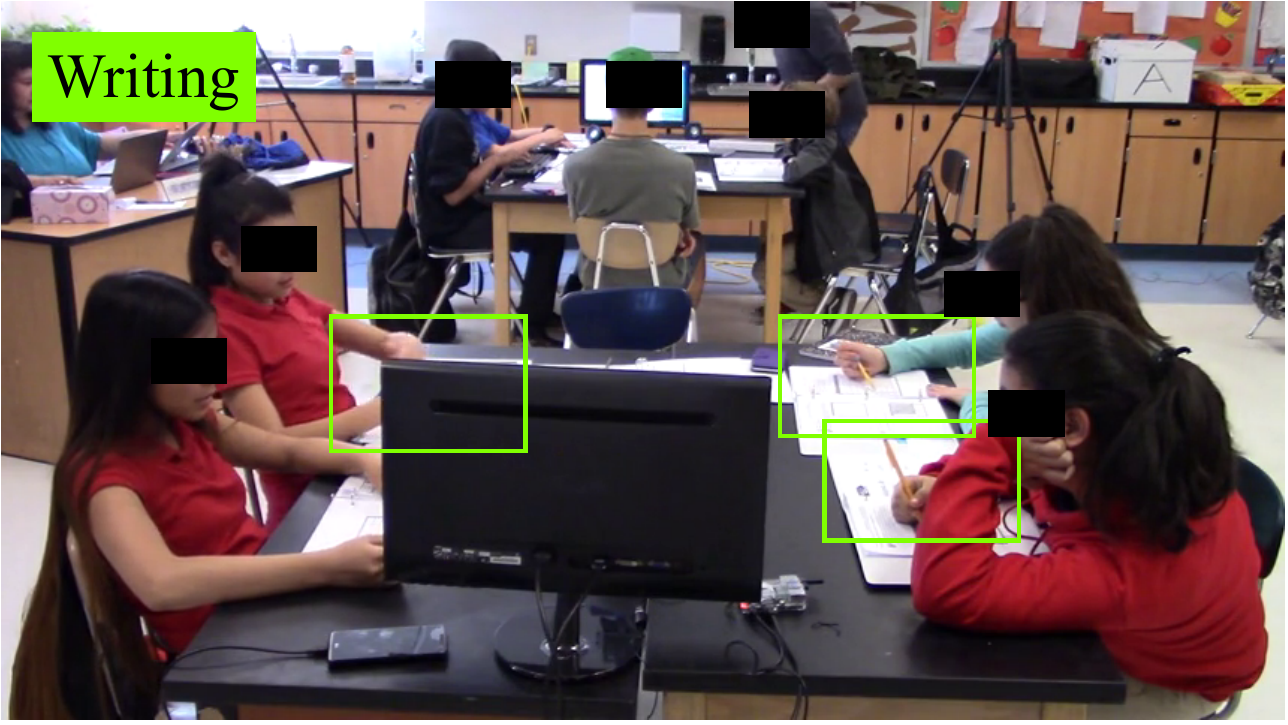}
    \caption{Writing activity in collaborative learning environment. Multiple writing activities with complete or partial occlusion are happening in this example.}
    \label{subfig:wr-det}
  \end{subfigure}

  \begin{subfigure}{0.97\textwidth}
    \includegraphics[width=\linewidth]{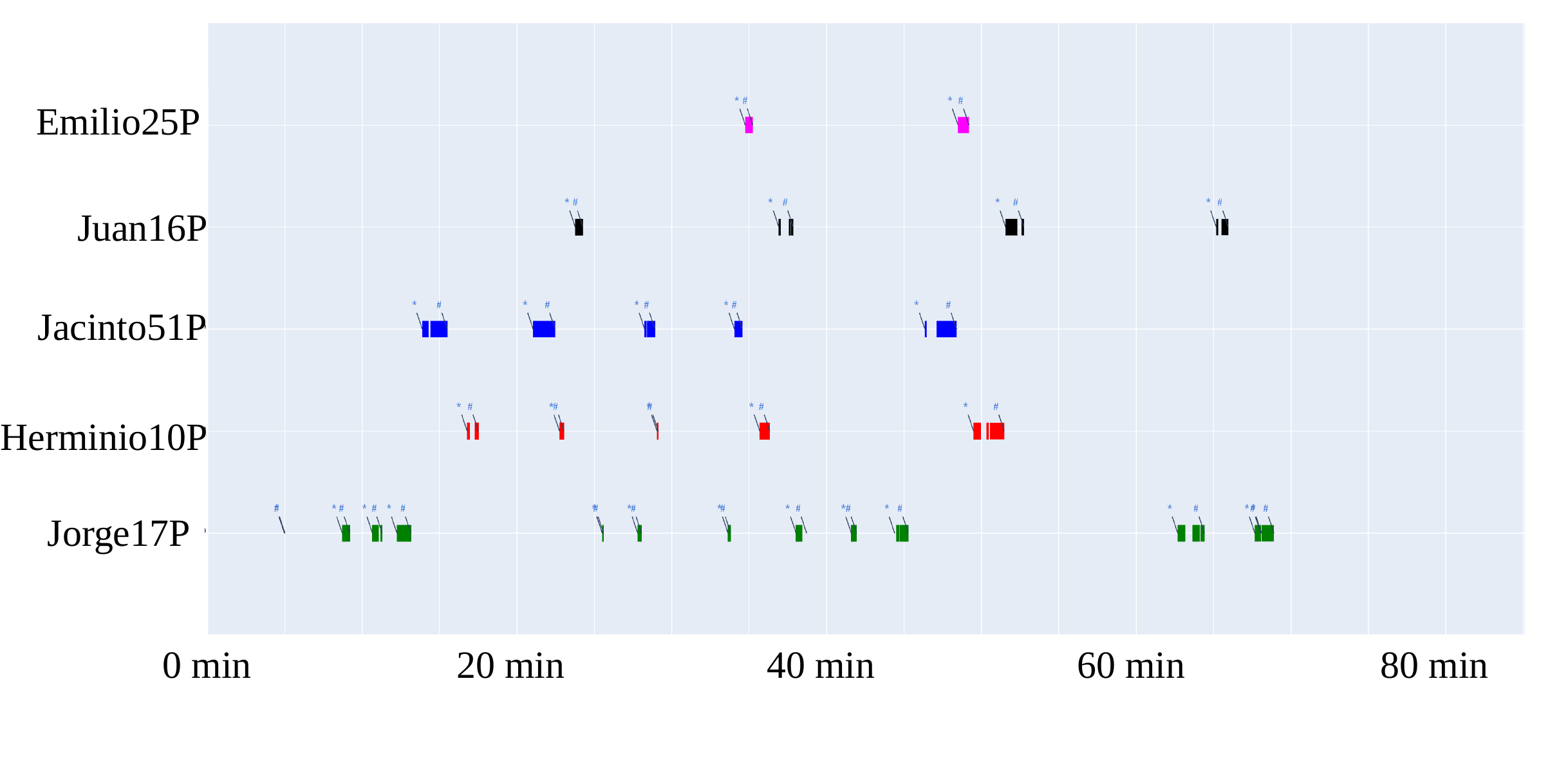}
    \caption{An activity map showing typing activity for a 1 hour 23 minute session. The asterisks are web-links that point to corresponding time in the video.}
    \label{subfig:act-map}
  \end{subfigure}

  \caption{\textbf{Typing and writing activities and expected visualization. The interactive activity map with the activities associated with the person helps the user to get a better understanding of the detected activities.}}
  \label{fig:our-act-det}
\end{figure*}

In contrast to standard activity detection problems, our research
requires a fast, modular long-term activity detection system and a
practical method for visualizing the results. We demonstrate an
example of such detection and visualization in Figure
\ref{fig:our-act-det}. Detection examples shown in Figure
\ref{subfig:ty-det} and \ref{subfig:wr-det} highlight how our dataset
and activities, such as typing and writing, differ from standard
activities and datasets.

Our activities involve subtle movements and can be performed in close
quarters by multiple people simultaneously, as illustrated in the
writing example in Figure \ref{subfig:wr-det}. Additionally, our
activities do not have clear beginnings and endings. A typing
activity, for example, can start without any prior hand movements. In
contrast, standard activities like soccer and playing guitar typically
have distinct starts and finishes. This presents a significant
challenge for us when it comes to locating the start and end times of
the activities.

Furthermore, to meaningfully interpret these activities, we need to
associate them with the person performing them for an extended
duration. Current state-of-the-art methods primarily focus on
detecting important activities locally, without considering the need
to associate activities with individuals over extended
periods. Detecting writing or typing activities for a short duration
is insufficient for providing useful information. It is necessary to
study these activities within the context of at least an entire
session (longer than 1 hour).

We accomplish this by summarizing the detections using an interactive
graph based on web technologies, as shown in Figure
\ref{subfig:act-map}. This interactive graph has a hierarchical
design, allowing users to zoom in and out, analyze specific time
segments (e.g., 0 to 20 minutes), and examine activities in
detail. Additionally, this visualization integrates with the website
(https://aolme.unm.edu/). The asterisk (*) at the beginning of each
activity serve as links to the corresponding timestamps in the video.

Understanding the learning process of students is a primary objective
of the AOLME project. Consequently, the interactions between students,
facilitators, and lessons are crucial. To illustrate this, students
engage in learning and interact in activities like typing (Figure
\ref{subfig:ty-det}), writing (Figure \ref{subfig:wr-det}), and
pointing at objects. It is essential to analyze and visualize these
activities within the context of sessions and, by extension,
groups. This analysis provides valuable information, such as the time
a student spends solving a problem in a session, whether a particular
student uses the keyboard more than others, or if a session encourages
students to work on paper or a computer, among other insights.

The primary motivation of this paper is to develop a
low-parameter, modular system with rapid inferencing capabilities,
capable of being trained on limited datasets to accurately detect and
associate specific activities, such as typing and writing, within
videos over extended periods of time. Additionally, we aim to create
an interactive web-based application to facilitate the user's further
examination and understanding of these activities.

The contributions of this paper include the development
of a fast, separable, low-parameter, and memory-efficient model
using 3D-CNNs for detecting writing and typing activities in
collaborative learning environments. While standard approaches
focus on modeling complex spatio-temporal features using highly
complex models for a general understanding of video content, the
paper employs separable models to detect the presence or
absence of specific activities within small regions of a video. These
regions are defined based on the outputs of well-studied and
established object detectors.

Additionally, the paper contributes a modular and fast
inferencing activity detection system. The system features a highly
modular design, enabling the incorporation of new research as long as
it provides spatial region coordinates of potential activity. For the
current application, the paper developed object detection
method for detecting keyboards and human hands. This approach can be
expanded by integrating more efficient and faster object detectors or
by detecting objects essential to an activity not yet explored in this
paper.

To ensure that the total time taken for inferring an activity is
within a reasonable duration (less than 2 hours for a 1-hour video),
the paper employs fast tracking and projection-based techniques
to further accelerate the module responsible for proposing activity
regions. Additionally, the paper develops optimal
batch-based inferencing to speed up the classification of proposal
regions, achieving a 9 $\times$ speedup compared to processing a
single proposal region at a time.

In addition to inferencing speed, the models used in the system can be
trained using a limited dataset. This is made possible by
employing transfer learning for standard object detectors and adopting
a low-parameter approach for classifiers. As a result, I was able to
create ground truth labels to train our system within a year. The
initial estimate for the amount of video activity ground truth needed
to train a complex end-to-end system was much higher. Furthermore, we
cannot use transfer learning for activity recognition within the
proposed activity regions, as standard video activity detection
datasets rely on person detectors, as surveyed by Elahe \etal
\cite{vahdani2022deep}, to propose regions of interest for human
activity. In contrast, given our focus on hand activities, we train a
hands detector directly, without requiring person detection first.

State-of-the-art activity detection systems do not thoroughly examine
the activity association problem in long videos. In our case, it is
crucial to associate and study activities within the context of a
video with a minimum duration of 1 hour. To provide this capability
for users, we employ interactive activity maps to summarize
and display our activity detections. These maps are based on web
application technologies and integrate into the website hosting the
videos. The web application features a hierarchical design and offers
links for easy viewing of the detected activities within the video.

The remainder of this paper is organized into 5 sections. Section
\ref{background} provides an a brief overview of standard activity
recognition datasets and methods. We also provide an overview of AOLME
dataset with emphasis on testing sessions.

In Section \ref{methodology}, we will describe the proposed system in
detail. We begin by providing an overview of the system's design and
components, followed by an in-depth discussion of the procedure for
optimizing a family of dyadic CNN architectures. The goal is to offer
a comprehensive understanding of the system's structure, function, and
performance, as well as the optimization methods employed to maximize
its efficiency and inference speed.

The training and testing procedures are descrived in Section
\ref{method_training_and_testing}, we will present efficient activity
labeling procedures and methods for extracting representative samples
from the labeled data to train our system. By outlining these
processes, we aim to demonstrate how our system effectively utilizes
the available data to optimize its performance in detecting and
associating activities within videos, even when working with limited
datasets.

In Secton \ref{results}, we showcase the results of each module within
our system and provide examples of using the complete system for
activity detection, ultimately generating interactive activity
maps. By demonstrating the outcomes and practical applications of our
system, we aim to highlight its effectiveness and speed in detecting
and associating activities within videos, as well as its ability to
present these findings in a user-friendly, interactive manner.

We conclude and provide insignts into future work in Section
\ref{conclusion_and_future_work}. These insights aim to provide a
foundation for further exploration and development, ensuring the
continued improvement of AOLME activity detection.

\section{Background}
\label{background}
In this section, we will start by looking at common activity
recognition datasets and frameworks in section \ref{sec:bg-HAR}. Then,
in section \ref{sec:bg-AOLME}, we will introduce the Advancing
Out-of-School Learning in Mathematics and Engineering research study
(AOLME) dataset and describe its unique challenges that make it
different from previously considered datasets.

\subsection{Human Actvity Recognition}
\label{sec:bg-HAR}
A human activity recognition system is designed to identify and
classify the actions performed by one or more individuals in a
video. These videos may include humans engaging in various activities,
with a background that may also be populated. The system must also
account for challenges such as differing video durations, scaling,
zooming, viewpoint changes, scene changes, and camera movements.

For a human activity recognition system to operate effectively, it
must accurately identify and extract the relevant spatiotemporal
features while disregarding any background elements that are not
related to the activity of interest performed by the person or persons
being observed. In section \ref{subsec:bg-har-dataset}, we will
present an overview of commonly used Human Activity Recognition (HAR)
datasets. Then, in section \ref{subsec:bg-har-systems}, we will
provide a summary of the activity recognition systems that are trained
on these datasets, which we will use as a standard of comparison
against our proposed system.

\subsubsection{Human activity recognition datasets}
\label{subsec:bg-har-dataset}

Common human activity recognition datasets are developed with the aim
of training models that can accurately classify and categorize videos
found on video streaming platforms, such as YouTube
\cite{yt-8m}. These datasets typically comprise a large collection of
popular human activities, such as ``Playing Guitar'', ``Cricket
Shot'', ``Soccer Penalty'',``Knitting'', and ``Sumo Wrestling''
(activities from UCF101 \cite{ucf101}). 

Upon closer inspection, it becomes evident that the videos in these
standard datasets primarily focus on individuals performing the
activity, with clear temporal differences between the actions
observed. For instance, running and jumping are easily distinguishable
activities. In contrast, activities in AOLME, such as writing versus
playing with a pencil, are much harder to differentiate. Furthermore,
our activities are carried out by multiple students seated in close
proximity, leading to a high degree of overlap between
activities. While overlapping activities are unwanted in standard
activity detection problems, they are desirable in our dataset as they
represent collaboration between the students.

\subsubsection{Human activity recognition systems}
\label{subsec:bg-har-systems}

Here we present the Human Activity Recognition (HAR)
systems that we use as a benchmark to compare against our proposed
low-parameter system. To ensure a fair comparison, we will retrain these
systems using our dataset and compare their performance against our
approach.

\textbf{Temporal Segment Network (TSN)} \cite{tsn} is a framework
designed for video-based action recognition that is centered around
the concept of long-range temporal structural modeling. This approach
combines a sparse temporal sampling strategy with video-level
supervision, resulting in high performance on datasets such as HMDB51
(69.4\%) and UCF101 (94.2\%). This approach relies on capturing the
temporal characterastics of an activity by sampling at regular
intervals. To implement TSN, a video is first divided into K segments
of equal duration.  The network then models a sequence of sparse
frames sampled from each segment and subsequently aggregates the
information obtained from these frames to make a final prediction
regarding the action being performed in the video. The sparse sampling
approach is effective when the activities being analyzed have distinct
temporal structures. However, for our dataset, it is difficult to
differentiate writing from its absence using sparse sampling, as there
are no clear temporal differences between the two. Therefore, a
complete modeling of the activity is required in order to effectively
differentiate between the two.

\textbf{Two-Stream Inflated 3D ConvNet (I3D)} \cite{i3d} is a powerful
framework used for video-based action recognition. It leverages
successful 2D image classification architectures and inflates them to
3D to learn spatio-temporal features from video data. The filters and
pooling kernels of very deep image classification ConvNets are
expanded into 3D, and the resulting inflated layers are inserted
between the original 2D layers, with weights shared between them. This
approach enables I3D to learn seamless spatio-temporal feature
extractors from video data, while leveraging the powerful
representations learned from 2D image data.

To further improve performance, I3D employs two input streams: one for
RGB input and the other for optical flow input. Each stream is
initialized with the weights of the corresponding 2D image
classification network and then inflated to 3D. I3D has achieved
impressive results on benchmark datasets, including 80.9\% accuracy on
HMDB-51 and 98.0\% accuracy on UFC-101.

Overall, I3D is a highly effective framework for video-based action
recognition, leveraging successful 2D image classification
architectures and inflating them to 3D to learn spatio-temporal
features from video data. Our method takes inspiration from I3D, but
we have chosen not to employ transfer learning from image
datasets. While the primary focus of our dataset is human activity
modeling, our study specifically targets writing and typing
activities. These activities may appear similar when observed within
the context of a person, as both are performed while seated at a table
and involve similar body movements, with the exception of hand
movements.

\textbf{Temporal Shift Module (TSMs)} \cite{lin2019tsm} is a highly
efficient and performant model that achieves 3D CNN-level performance
while maintaining the complexity of a 2D CNN. By moving a portion of
the channels along the temporal axis, TSM facilitates communication
between neighboring frames and enables efficient temporal
modeling. This feature, along with its support for both offline and
online video recognition, make TSM a versatile and powerful model for
analyzing videos.

In offline tests, TSM achieved impressive results, with 74.1\%
accuracy on Kinetics, 95.9\% on UCF101, and 73.5\% on HMDB51. Online,
TSM was able to achieve 74.3\%, 95.5\%, and 73.6\% on the same
datasets respectively. As a result, TSM is a highly effective and
efficient tool for video analysis tasks.

\textbf{SlowFast} \cite{feichtenhofer2019slowfast} is a video analysis
model that comprises of a Slow pathway and a Fast pathway. The Slow
pathway operates at a lower frame rate and captures spatial semantics,
while the Fast pathway operates at a higher frame rate and captures
motion at a finer temporal resolution.

SlowFast models have demonstrated strong performance in both action
classification and detection in video, with significant improvements
attributed to the SlowFast concept. The Slow pathway in a SlowFast
network is designed to have a low frame rate and lower temporal
resolution, while the Fast pathway has a high frame rate and greater
temporal resolution.

Overall, the SlowFast model architecture provides a powerful and
effective means of capturing spatio-temporal features from video, with
the Slow and Fast pathways working together to achieve impressive
results in video analysis tasks.

\subsection{Group interactions video dataset in AOLME}
\label{sec:bg-AOLME}

\begin{figure*}[!t]
  \centering  

  \begin{subfigure}{0.45\textwidth}
    \includegraphics[width=\linewidth]{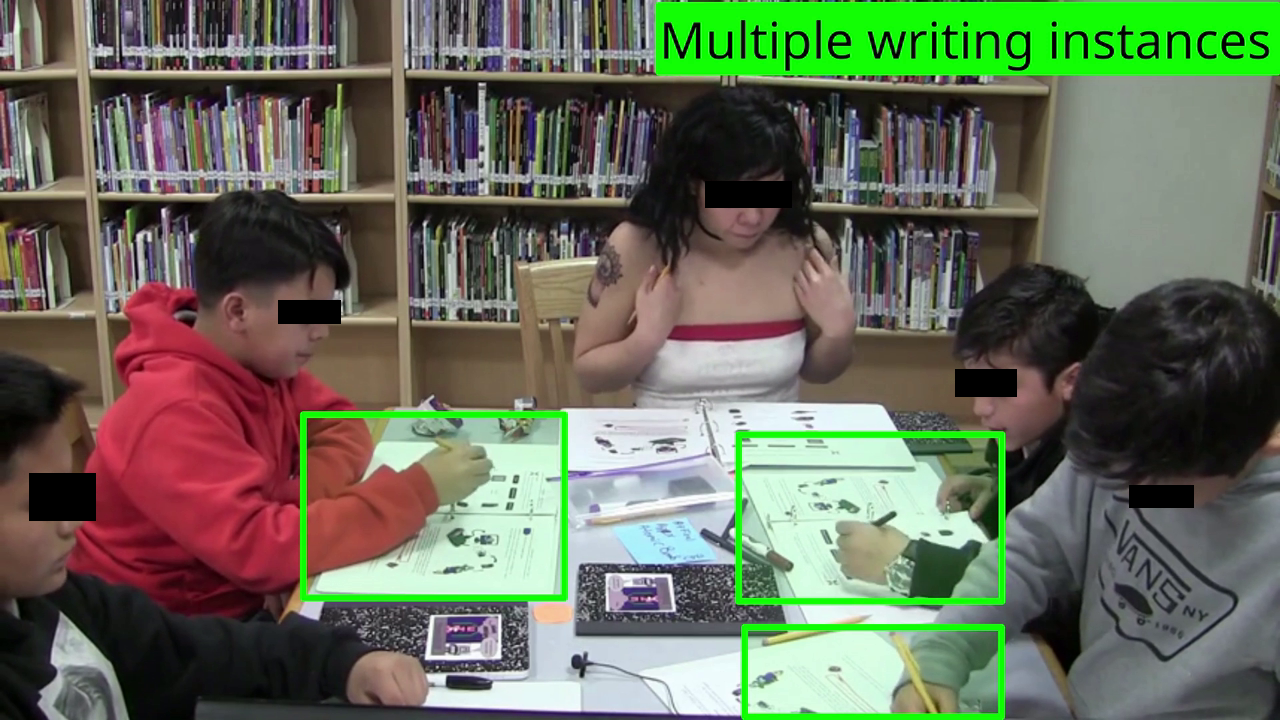}
    \caption{Video with camera very near to the table with multiple writing activities.}
    \label{subfig:multiple-activities}
  \end{subfigure}~~  
  \begin{subfigure}{0.45\textwidth}
    \includegraphics[width=\linewidth]{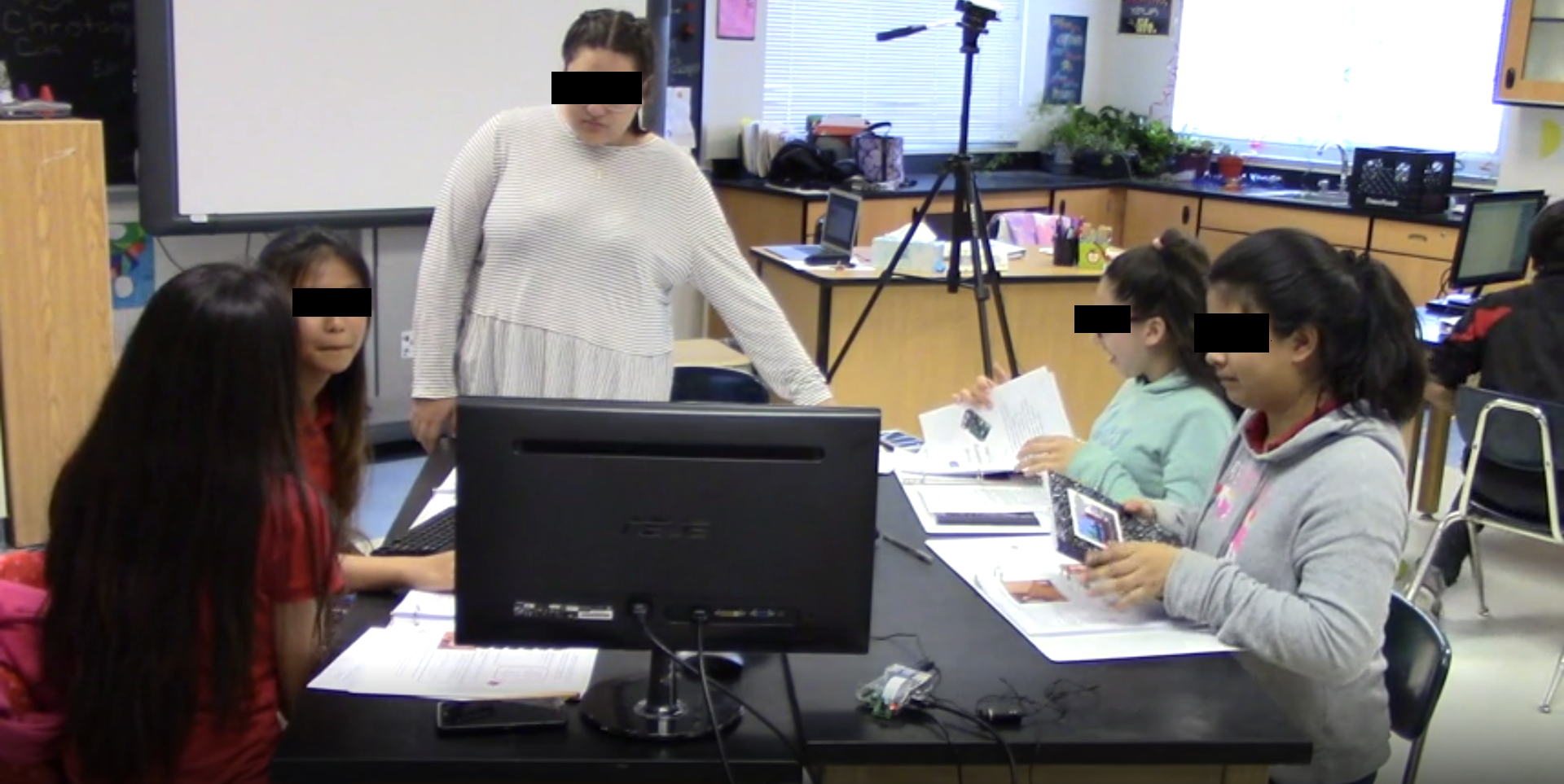}
    \caption{Video with camera very far from the table.}
  \end{subfigure}
  
  \begin{subfigure}{0.45\textwidth}
    \includegraphics[width=\linewidth]{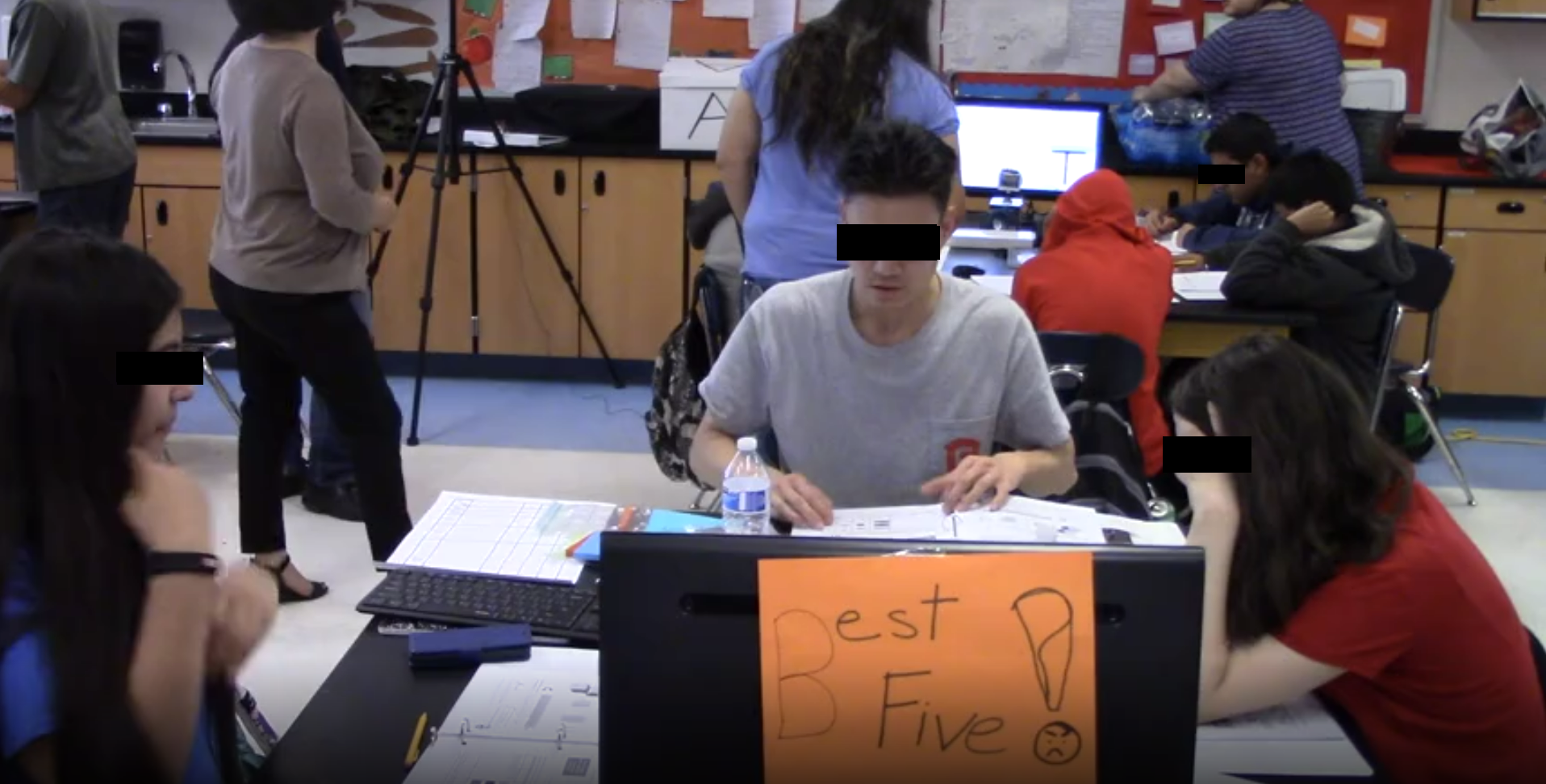}
    \caption{Video with group interactions against a dark table.}
  \end{subfigure}~~
  \begin{subfigure}{0.45\textwidth}
    \includegraphics[width=\linewidth]{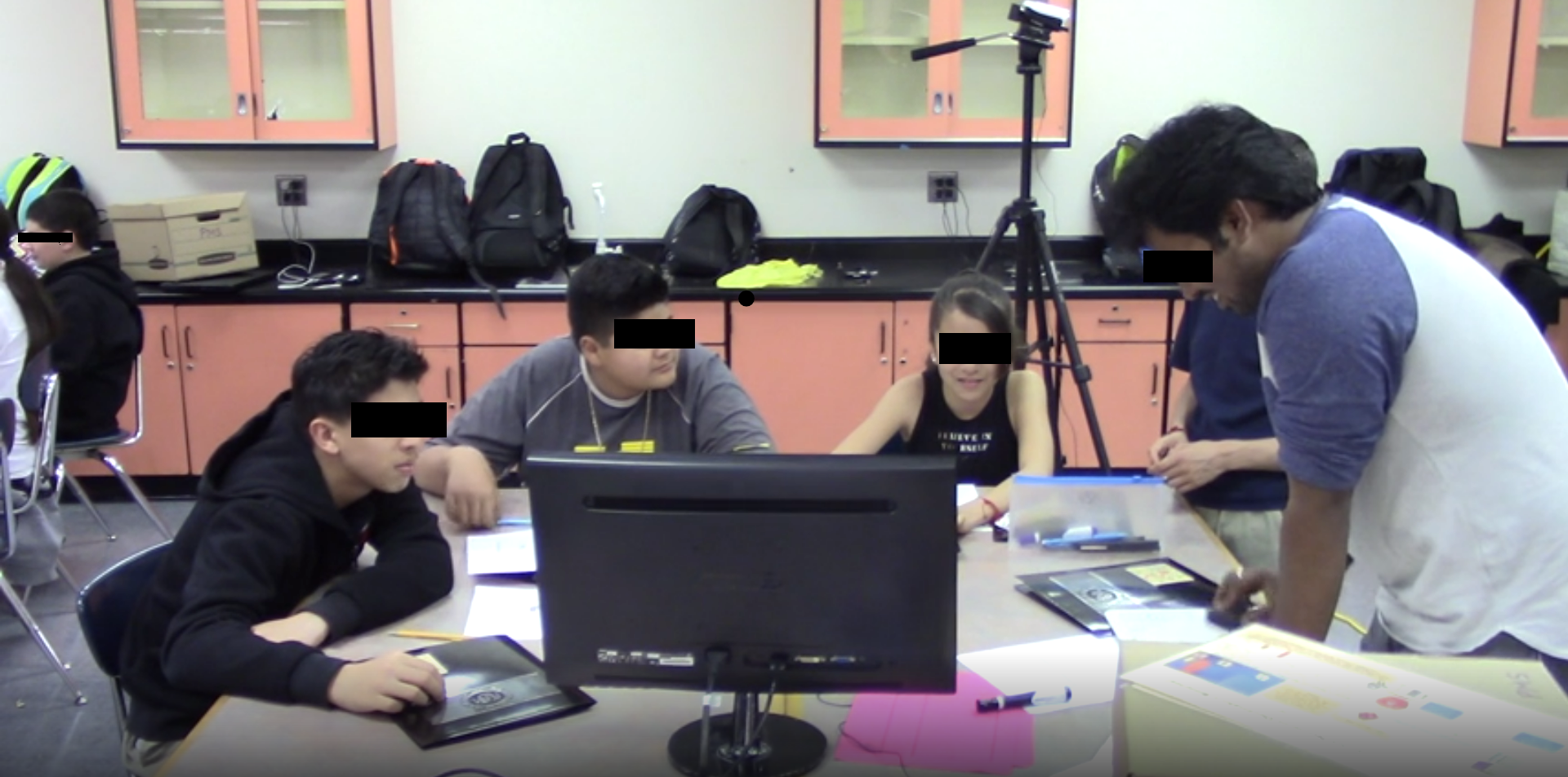}
    \caption{Video with group interactions against a white table.}
  \end{subfigure}

  \begin{subfigure}{0.45\textwidth}
    \includegraphics[width=\linewidth]{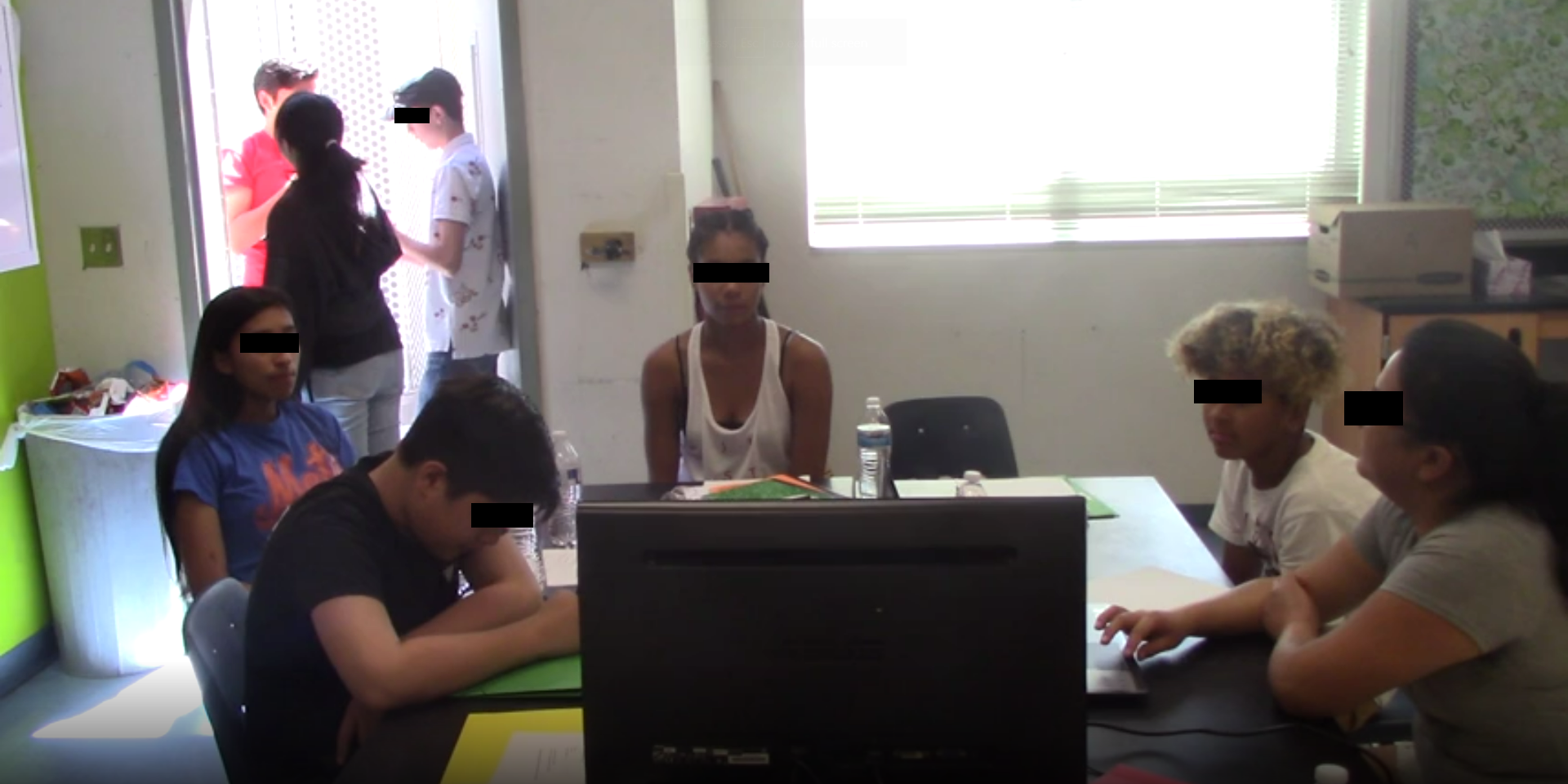}
    \caption{Video with very bright natural light in the background.}
  \end{subfigure}~~
  \begin{subfigure}{0.45\textwidth}
    \includegraphics[width=\linewidth]{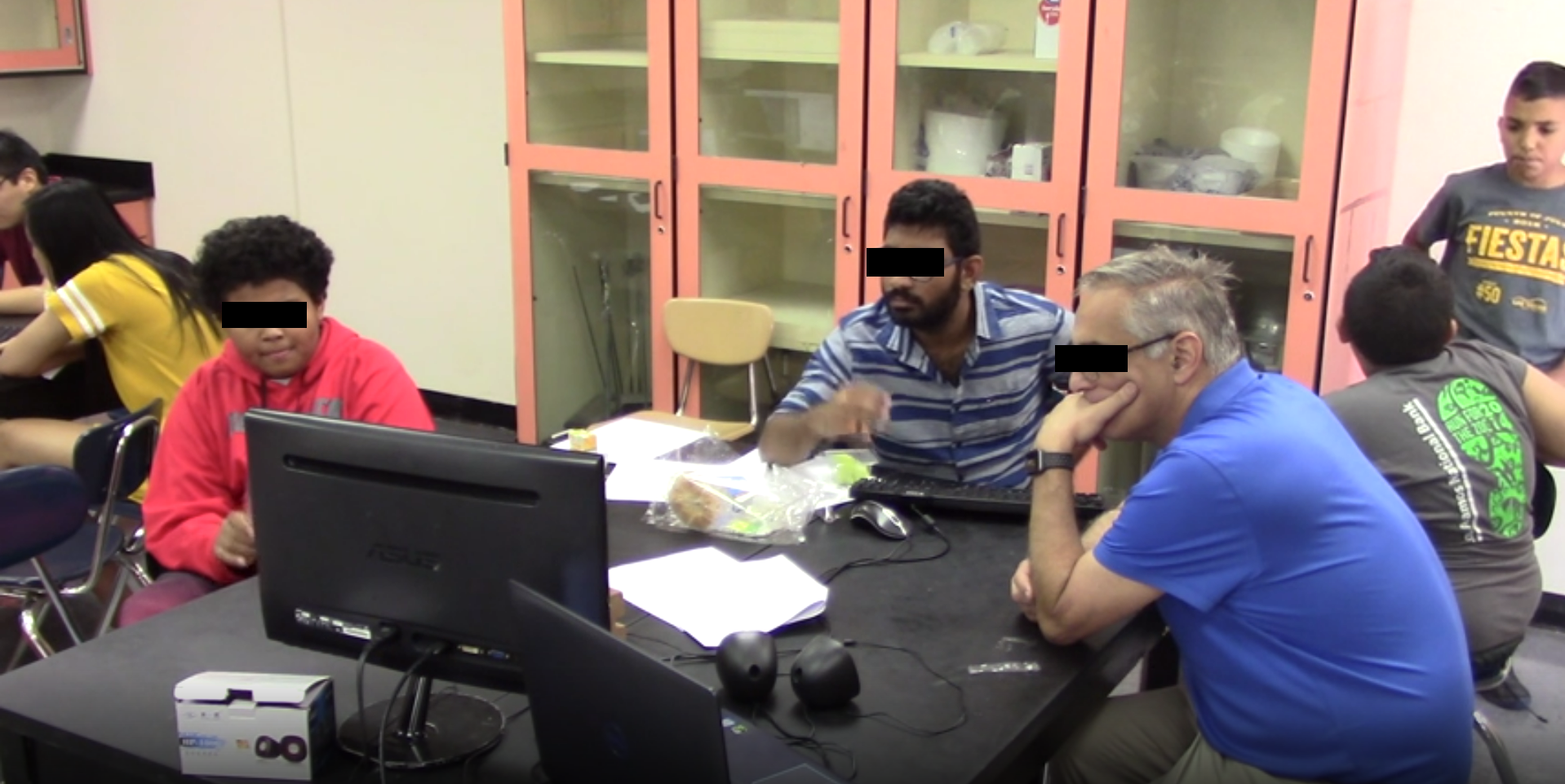}
    \caption{Video under uniform artificial lighting.}
  \end{subfigure}

  \begin{subfigure}{0.45\textwidth}
    \includegraphics[width=\linewidth]{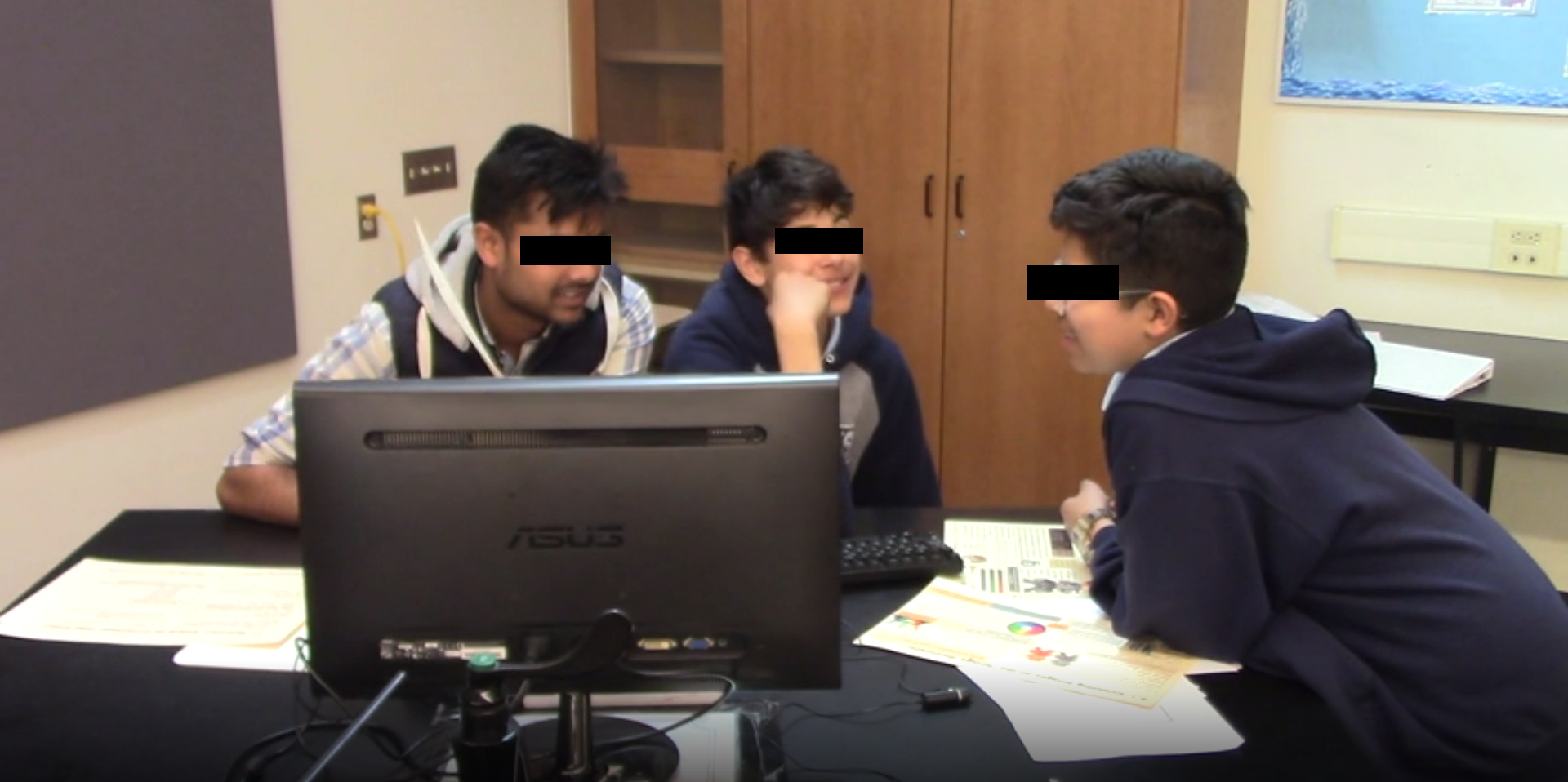}
    \caption{Video having male students sitting to the right side of the table.}
  \end{subfigure}~~
  \begin{subfigure}{0.45\textwidth}
    \includegraphics[width=\linewidth]{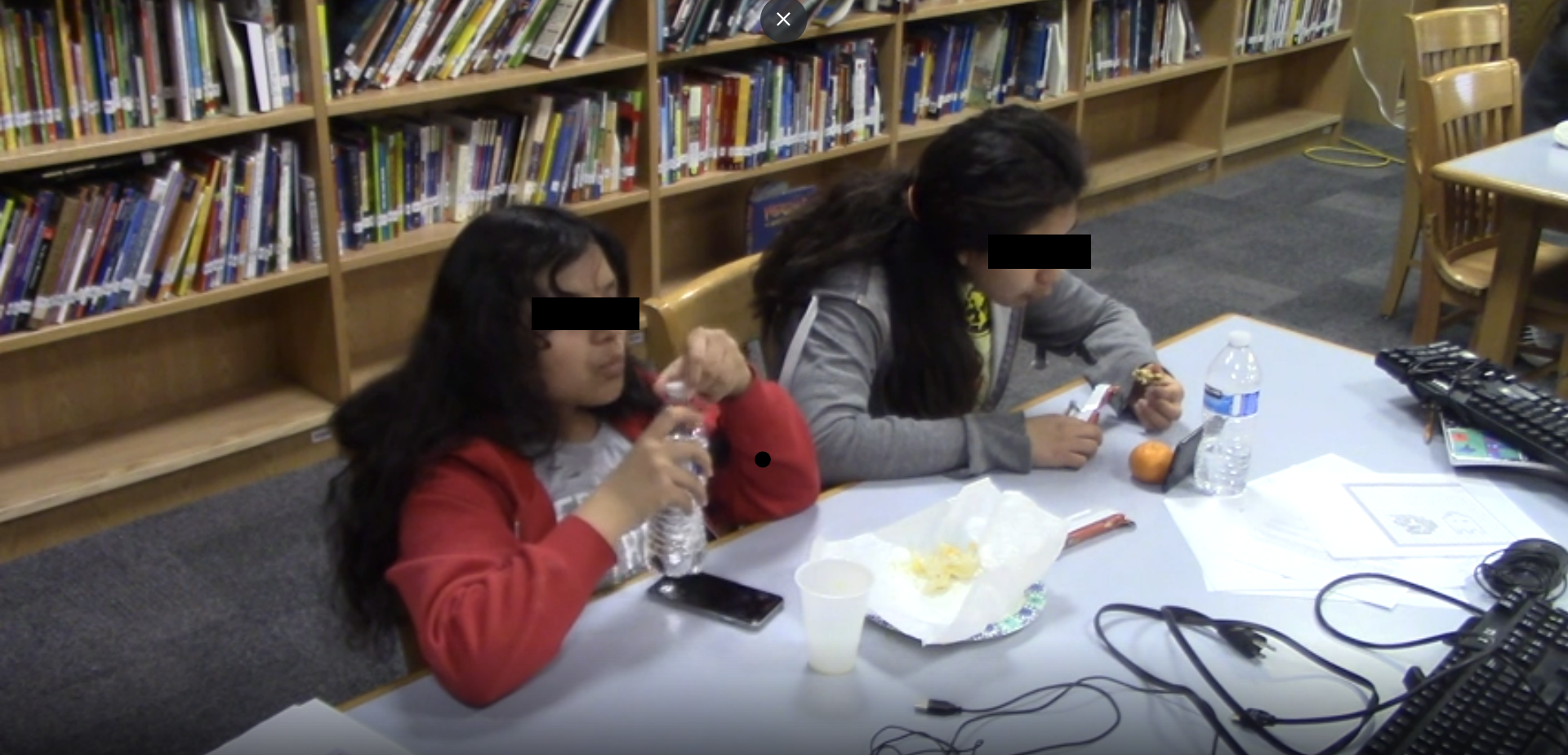}
    \caption{Video having female students sitting to the left side of table.}
  \end{subfigure}

  \caption{\textbf{Figure showing variability in AOLME group interaction videos.}}
  \label{fig:variability-in-group-interactions}
\end{figure*}

This section introduces and compares our dataset, which consists of
group interaction videos collected as part of the AOLME project. AOLME
is an interdisciplinary project undertaken by the Department of
Electrical and Computer Engineering and the Department of Language,
Literacy and Sociocultural Studies at a university in Southwest
(UNM), and involved the collection of approximately 2,218 hours of
multimedia data over three years. The multimedia data includes group
interaction videos, audio recordings, screen recordings of laptops
using Active Presenter \cite{ActivePresenter}, and screen recordings
of Raspberry Pi \cite{raspberry_pi} using external cameras.

For this paper, we focus on analyzing the group interaction
videos, refer figure \ref{fig:variability-in-group-interactions} to
detect instances of typing and writing. We compare our dataset to
standard datasets commonly used in human activity recognition, which
typically feature well-separated actions performed by
individuals. However, our dataset presents unique challenges due to
the close proximity of multiple individuals performing subtle actions,
making accurate classification and categorization more difficult.

\subsubsection{Group interactions video dataset naming convention}
AOLME project spans several years and groups, with $987$ hours of
group interactions videos. We organize the videos into cohorts,
levels, school and groups. We use cohort-1, cohort-2, and cohort-3 to
indicate the years 2017, 2018 and 2019. Each cohort contains different
levels of AOLME implementation. Each level has two schools and each
school has several student groups. Each group contains $2$ to $5$
students, a facilitator and a co-facilitator.

Each student group has typically about ten to twelve sessions per
level. A sessions typically lasts $45$ minutes to $90$ minutes. For ease
of recognition, the videos are labeled as C1L1P-A, Mar02 which means
Cohort 1, Level 1, Rural, Group A on March 2nd (single session). This
notation is followed throughout this thesis document in the later
sections to present results.

\begin{table*}[t]
  \caption{\textbf{Table comparing video characteristics of AOLME and public datasets used in testing 
      Activity Recognition algorithms.
      We use ~~\xmark~~ to denote the absence of a video property and ~~\vmark~~ to denote the presence of a video property.}}
  \label{tab:our_dataset_vs_public}
  \centering
  \begin{tabular}{p{3.5 cm}| p{2cm}p{2cm}p{2.5cm}p{2.5cm}|p{2cm}}
    \hline
    \textbf{Video activity detection problems}                       & \multicolumn{4}{c|}{\textbf{Public datasets}} & \textbf{AOLME}                                              \\
                                                                     & UCF101                                        & HMDB51      & Kinetics-400    & Acitivity-Net      &        \\
                                                                     & \cite{ucf101}                                 & \cite{hmdb} & \cite{kinetics} & \cite{activitynet} &        \\
    \hline
    \hline
    
    Multiple simultaneous activities          & \cellcolor{red!25}No    & \cellcolor{red!25}No    & \cellcolor{red!25}No    & \cellcolor{red!25}No    & \cellcolor{green!25}Yes \\
                                                                     &                         &                         &                         &                         &                         \\
    Very fast transition between activities   & \cellcolor{red!25}No    & \cellcolor{red!25}No    & \cellcolor{red!25}No    & \cellcolor{red!25}No    & \cellcolor{green!25}Yes \\
                                                                     &                         &                         &                         &                         &                         \\
    Various camera angles                     & \cellcolor{green!25}Yes & \cellcolor{green!25}Yes & \cellcolor{green!25}Yes & \cellcolor{green!25}Yes & \cellcolor{green!25}Yes \\
                                                                     & \cellcolor{green!25} & \cellcolor{green!25} & \cellcolor{green!25} & \cellcolor{green!25} & \cellcolor{green!25} \\

                                                                     &                         &                         &                         &                         &                         \\
    Long term occlusion                       & \cellcolor{red!25}No    & \cellcolor{red!25}No    & \cellcolor{red!25}No    & \cellcolor{red!25}No    & \cellcolor{green!25}Yes \\
                                                                     & \cellcolor{red!25}    & \cellcolor{red!25}    & \cellcolor{red!25}    & \cellcolor{red!25}    & \cellcolor{green!25} \\

                                                                     &                         &                         &                         &                         &                         \\
    Similar looking activiities & \cellcolor{red!25}No & \cellcolor{red!25}No & \cellcolor{red!25}No & \cellcolor{red!25}No & \cellcolor{green!25}Yes    \\
                                                                     &                         &                         &                         &                         &                         \\
    Duration $\ge $ 15 min                    & \cellcolor{red!25}No    & \cellcolor{red!25}No    & \cellcolor{red!25}No    & \cellcolor{red!25}No    & \cellcolor{green!25}Yes \\
                                                                     & \cellcolor{red!25}    & \cellcolor{red!25}    & \cellcolor{red!25}    & \cellcolor{red!25}    & \cellcolor{green!25} \\
    
    \hline
  \end{tabular}
\end{table*}

Table \ref{tab:our_dataset_vs_public} provides a comparative analysis
of our group interaction video dataset against commonly used public
datasets, highlighting key differences. One significant distinction is
the duration of our videos, which typically ranges from 1 to 1.5
hours. As a result, the activities we are studying are scattered
across these long-duration videos, requiring a carefully designed
context-based approach to detect them. Additionally, our dataset
features multiple activities occurring simultaneously in close
proximity, whereas standard datasets usually focus on a single
activity with others occurring in the background. For example, in
figure \ref{subfig:multiple-activities}, multiple writing activities
are taking place in close proximity, highlighting the spatial
closeness challenge. In addition to spatial closeness, our dataset
poses challenges as activities can transition rapidly between one
another, making it difficult to create a reliable ground truth and
design an activity detection system.

In addition to spatial and temporal challenges, our group interaction
videos can also involve long-term occlusions. The camera is often
positioned behind a monitor, obscuring the activities of students
seated close to the screen. Unlike standard datasets that usually
ignore occluded regions, our approach aims to capture any visible
activities in such areas. Therefore, our activity detection system
must consider the possibility of occlusions and incorporate methods to
handle such situations accurately. Overall, these unique features of
our dataset pose significant challenges for activity detection and
require innovative solutions.

\section{Methodology}
\label{methodology}
\subsection{Video Activity Recognition and Visualization System}
\label{sec:system}

\begin{figure*}[t]
  \includegraphics[width=\linewidth]{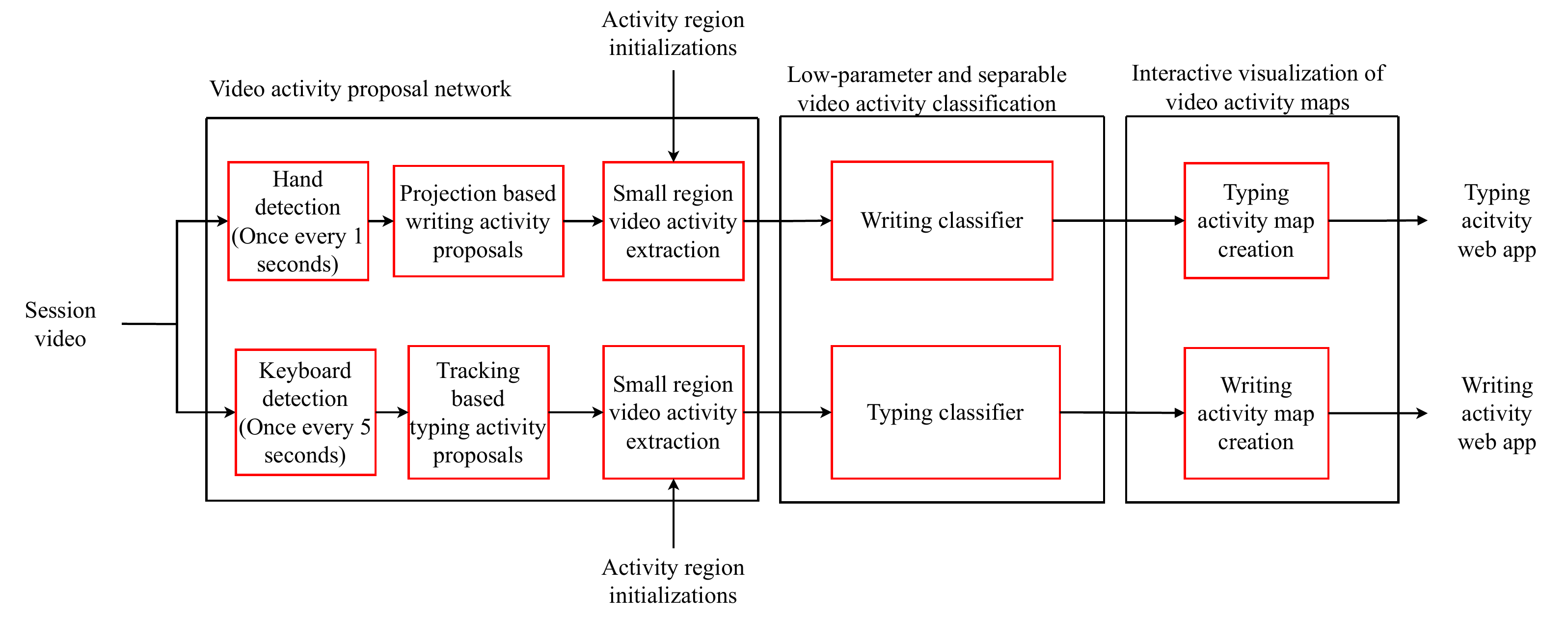}
  \caption{\textbf{System diagram of activity detection system for typing and
      writing in AOLME group interaction videos.}}
  \label{fig:activity-detection-system-overview}
\end{figure*}

This section presents a high-level description of the video activity
recognition and visualization system and provides a concise summary of
its design characteristics. The system is designed to detect and
accurately quantify typing and writing activities in AOLME group
interaction videos. A top-level diagram of the system is shown in
Figure \ref{fig:activity-detection-system-overview}.

We will describe the system in terms of three separate stages. First,
the video activity segment proposal network generates candidate video
segments of possible human activity of a specific type. In our case,
we generate proposals for writing and typing acitivities. Second, for
each type of activity, we use separate low-parameter video segment
classifiers to determine whether the activity is taking place. Third,
the interactive visualization stage uses the activity detection
results to generate an interactive visualization of the differenty
types of activities.

We note that the video activity segment proposal network is designed
to reduce computational complexity while improving detection accuracy
of the overall approach. Here, we use object detection over a single
frame, sampled every few seconds (refer to diagram) to localize each
activity.  Here, we incorporate reliable detection of the keyboards
and human hands used in typing and writing. Then, we track each object
over a short video segment before we consider object detection
again. Here, the idea is that object detection alone is
computationally expensive while not capturing the physical
characteristics of object motions. Instead, object motion is covered
through object tracking. Furthermore, after a short period, we perform
object detection and restart the process to avoid long-term failures
from tracker failures.

In order to associate human activity with specific students, we
require method initialization by specifying activity regions over the
table. The basic idea here is to segment the table into regions, where
each region is associated with a specific student. The initialization
avoids developing person tracking through occlusions. We refer to the
fast face recognition by Tran \etal\cite{Tran2021} for methods for
addressing such issues.  We note that the video initialization
requires minimal human input in the sense that marking a single frame
can be applied over very long video sessions.

We use separable, low-parameter video classifiers for detecting each
activity. Each classifier is built using a 3D-CNN.  The separable
design was found to perform much better than the use of common
features with two separate classes. The individual 3D CNNs were thus
trained for a specific activity, within the context of the object
detected for the activity. Here, we note that the context was derived
from the fact that each video segment is associated with a specific
object detected for the activity. For writing, we classify hand
movements associated with hand detection. For typing, we classify
motions associated with the keyboard. Here, we note that there was no
need to require both hand detection and keyboard detection for
recognizing typing. A reliable keyboard detector proved sufficient for
identifying keyboard activities without the need for hand detection
complicated by occlusions associated with the presence of the
keyboard.

Once the proposals have been classified, we perform a post-processing
operation to clean and create a long-term interactive activity
visualization. This involves filtering out any false positives and
generating a visualization that shows the activities that occurred
over a longer period of time. This visualization can be used to gain a
better understanding of how users interact with each other during
group interactions, and to identify patterns and trends in their
behavior.

Overall, the combination of the low parameter classifier and the
post-processing operation enables our activity detection system to
detect typing and writing activities in AOLME group interaction
videos, providing valuable insights into user interactions and
behavior.

\textbf{Fundamental characteristics of our system design:}
We summarize three essential characteristics of our design that
enabled us to train the system with limited datasets, fast inference,
and system tuning based on visualizing the results from the different
stages.

\textbf{System component training using limited datasets:}
Our approach requires significantly fewer data to train because the
components of the system that require training themselves need very
little data. Specifically, we employ effective object detections for
hands and keyboards to minimize the need for new datasets. We begin
with pre-trained models and utilize transfer learning to train our
object detector, using only 700 images for detecting keyboards and 305
images for detecting hands.

The object tracking and table region labels provide us with
spatiotemporal regions potentially containing the current group's
typing and writing activities. As we have already filtered regions
that do not belong to the table of the current group, our classifiers
only need to distinguish between typing and no-typing or writing and
no-writing. This simplifies the problem significantly, and we can
solve it using a low-parameter model that can capture subtle temporal
changes.

\textbf{Fast inference:}
The  time  consuming  parts  of our  framework  inference  are  object
detection and  activity classification. An hour  of group interactions
video  takes between  3  and  15 minutes  through  the video  activity
proposal  detection  stage.  The  performance of  the  video  activity
classifiers is  heavily dependent on  the the number of  video segment
proposals. In our experiments we are able to process an hour of region
proposals  in less  than  20 and  40 minutes  for  typing and  writing
respectively.  The  second  stage  inference  speed  up  is  primarily
achieved due to batch based inference  made possible due to low memory
requirement of  our low-parameter model.  We also would like  to point
our that  we use  NVIDIA Quadro RTX  5000 GPU.

\begin{figure*}[t]
  \centering  

  \begin{subfigure}{0.47\textwidth}
    \includegraphics[width=\linewidth]{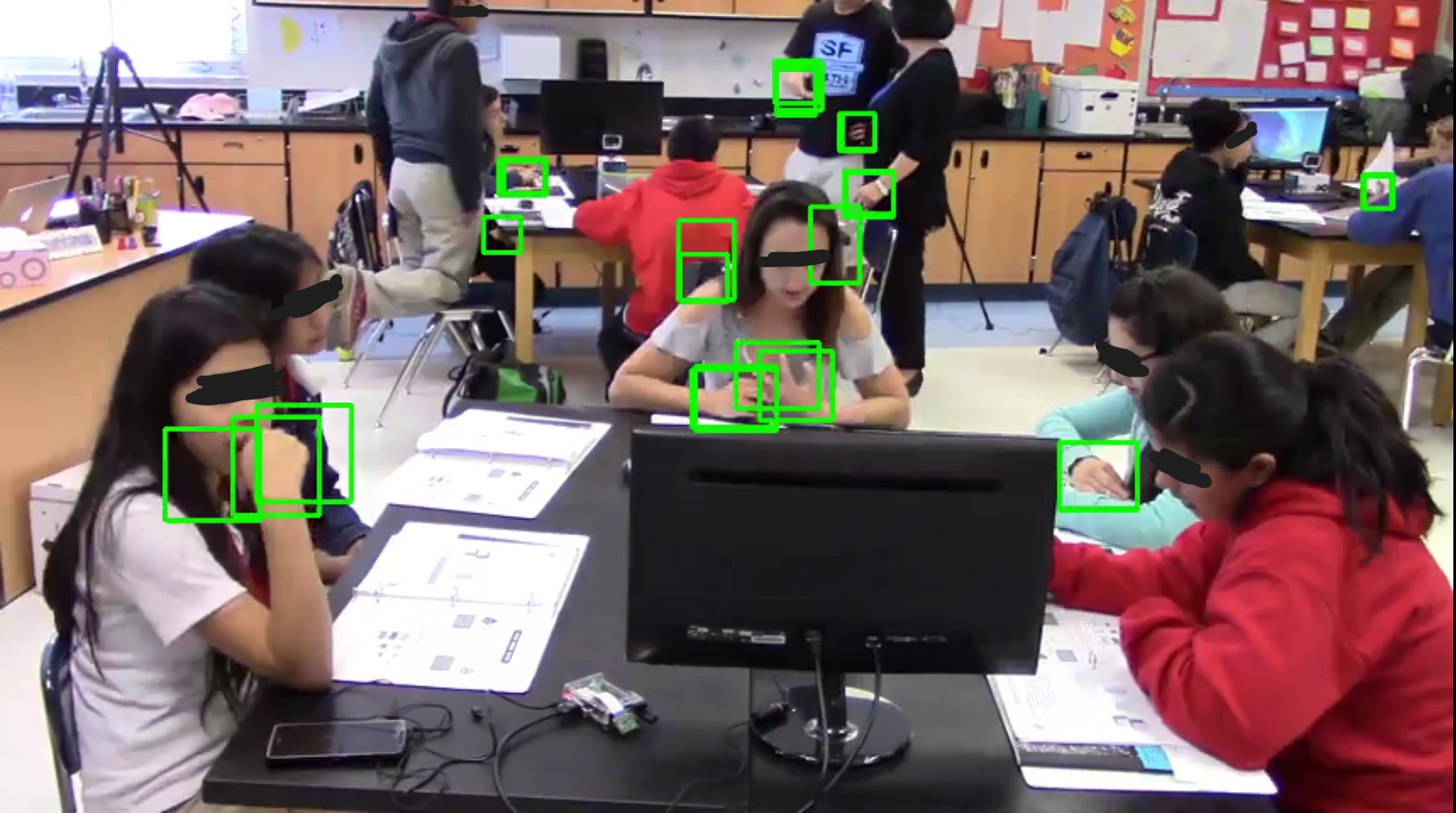}
    \caption{Hand region tracking prior to projection-based post-processing revealing a significant number of false positives.}
    \label{fig:method-modular-design-a}
  \end{subfigure}
  ~~  
  \begin{subfigure}{0.47\textwidth}
    \includegraphics[width=\linewidth]{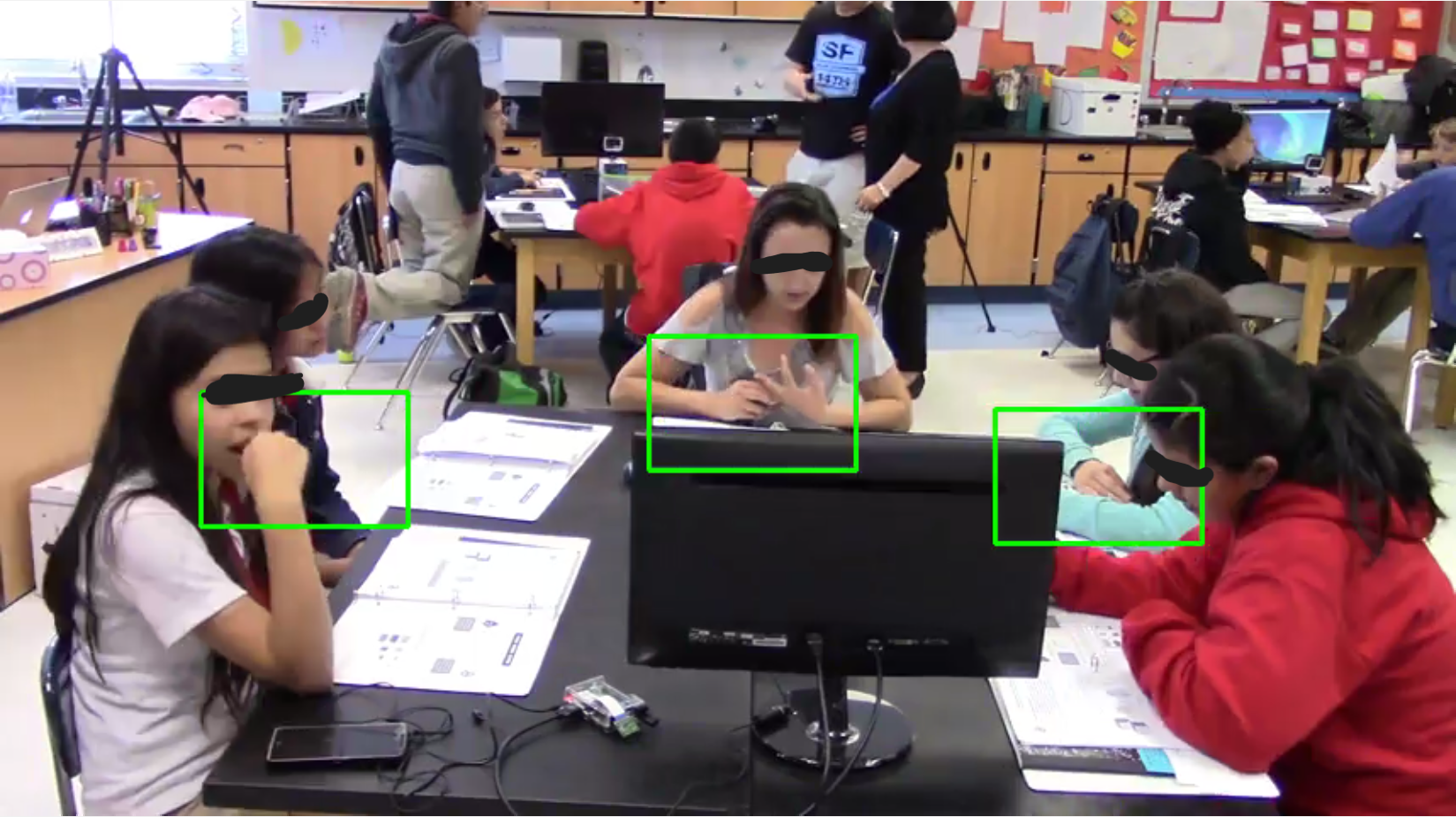}
    \caption{Hand detections after projection-based post-processing exhibited a notable reduction in the number of false positives.}
  \end{subfigure}

  \caption{\textbf{The modular two-stage approach generates intermediate results that can be analyzed and visualized, enabling us to identify opportunities for improving the activity region proposals until they provide satisfactory qualitative and quantitative performance.}}
  \label{fig:method-modular-design}
\end{figure*}

\textbf{System tuning using a modular design:}
The modularity of the our system allows for greater transparency and
interpretability, enabling us to gain insights into the system's
internal workings and how it arrives at its final output. An example
scenario in which the modular design helped improve performance is
illustrated in Figure \ref{fig:method-modular-design}. During the
initial stages of development, we employed hand detections without
post-processing to identify active regions on the table, which
resulted in many false video segment proposals for writing. By
visually analyzing the hand detections, we identified an excess of
false positives. We then applied projection-based post-processing
techniques, to be described below, to reduce the number of false positives, resulting in
improved system performance. Thus, modular design enabled us to isolate and address issues within each module.

\subsection{Fast object tracking}
\label{method-fast-object-tracking}

Fast object tracking uses Faster-RCNN for detecting objects at regular
intervals. Faster-RCNN proved very effective at object detection and
there was no need to consider alternative methods.

Once the objects are detected, the system employs different tracking
strategies based on their movement characteristics. The keyboard,
which has less movement, is tracked using a fast but moderately
accurate traditional tracker such as KCF \cite{KCF}. We use keyboard
detection on only one frame every 5 seconds reducing inference
time. The keyboard location in between these detections is calculated
using the KCF tracker.

In contrast, hands change position more rapidly, making them more
challenging to track. To address this, the system employs a temporal
projection-based strategy. The strategy is to collect hand detections
on one frame per second for 12 seconds and only consider regions that
have consistently higher detections. This helps to stabilize the
regions on the table where the hands are placed, enabling more
accurate hand tracking on the table for current group.

\subsection{Low parameter separable activity classifier optimization}
\label{sec:method-activity-classifiers}
This section describes the design and optimization of low-parameter
classification models. As mentioned earlier, we use independent models
for detecting typing and writing. This requires careful modeling of
small and subtle temporal variations in the video data. We developed an
optimization framework that finds the optimal architecture from a
hierarchy of low-parameter 3D-CNN architectures.

During the initial exploratory phase, we created 3D-ConvNet models
capable of classifying multiple activities. However, we discovered
that such models added unnecessary complexity without significantly
enhancing performance. Therefore, we developed two distinct binary
classification architectures to effectively model each activity. This
approach enabled us to streamline the models while enhancing their
overall performance.

This section is divided into two subsections. In Section
\ref{subsec:3D-cnn-models}, we present the architecture of our
low-parameter 3D-ConvNet models. In Section
\ref{subsec:model-optimization}, we describe the optimization procedure
used to determine the optimal number of dyads and input frame rate.

\subsubsection{3D-ConvNets models architecture}
\label{subsec:3D-cnn-models}

\begin{figure*}[t]
  \includegraphics[width=\linewidth]{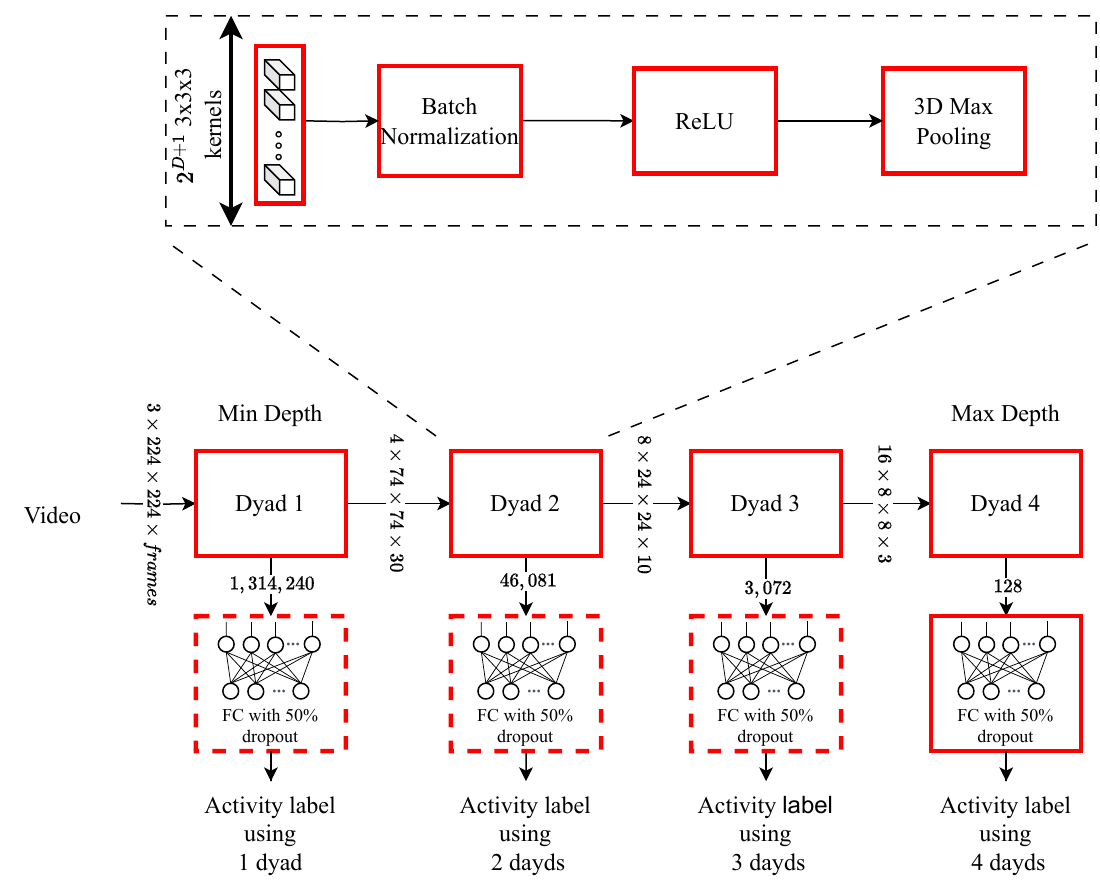}
  \caption{\textbf{Family of dyadic architectures, $\mathcal{A}$, produced by
      varying depth.}}
  \label{fig:model-architecture}
\end{figure*}

In this section, we present a family of low-parameter neural network
architectures that are designed for the effective modeling of specific
activities. These architectures demonstrate a dyadic structure, where
the addition of each dyad leads to an increased depth ($D$) of the
overall architecture. Figure \ref{fig:model-architecture} displays the
internal components of each dyad, which include 3D-ConvNet kernels,
batch normalization, ReLU activation, and 3D max-pooling.

At the beginning of each dyad, there are 3D-ConvNet kernels, whose
number depends on the depth ($D$) of the dyad. Our empirical findings
suggest that NVIDIA GPUs can efficiently handle 8 kernels at a fast
rate. However, the starting dyad has only 4 kernels, while the
subsequent dyads have a multiple of 8 kernels. Specifically, a dyad at
depth $D$ has $2^{D+1}$ 3D-ConvNet kernels, as shown in figure
\ref{fig:model-architecture}. After the convolutional layer, the
output features are passed through batch normalization and ReLU
activation before going through 3D-MaxPooling.

In addition to supporting different depths, our family of
architectures also accommodates input videos with varying frame rates,
$fr$. To facilitate this, we modify the kernel size of the first dyad
max pooling layer to $3\times 3\times d_{fr}$. The rest of the
architecture always use a $3\times 3\times 3$ max pooling kernels. The
value of $d_{fr}$ depends on the input video frame rate, which is given
by $(3\times fr)/30$. For example a video having frame rate ($fr$)
equal to 10 goes through a max pooling kernel of size
$3 \times 3\times 1$ kernel at its first depth. This modification
allows us to keep the rest of the architecture unchanged. This also
ensures equal number of total parameters for architectures having same
depth for different video frame rates as shown in Figure
\ref{fig:model-architecture}.

The maximum depth, $D_{max}$, of our architecture depends on the size
of the input video. For our dataset videos, which have a size of
$3 \times 224 \times 224 \times fr$, the maximum depth that can be
supported is 4. Therefore, in total, we have 12 models, and we need to
choose the optimal one among them.

\subsubsection{Model frame rate and depth
  optimization}
\label{subsec:model-optimization}

In this section, we describe the procedure for selecting the optimal
model from the family of 12 models described in Section
\ref{subsec:3D-cnn-models}. Let $A_{D, fr}$ denote a particular, fixed
neural network architecture having depth $D$, $D \in \{1, 2, 3, 4\}$,
and input video frame rate of $fr$, $fr \in \{10, 20, 30\}$. Let
$W_{D, fr}$ be the weights associated with $A_{D, fr}$. Let the family
of neural networks we are considering for optimization be denoted
using $\mathcal{A} \in \{A_{1, 10}, A_{1, 20}, \dots, A_{4, 30}\}$. To
find the optimal architecture, $A^*_{D, fr} \in \mathcal{A}$, we must
first determine the optimal weights, $W^*_{D, fr}$, for each
architecture, $A_{D, fr}$, and then determine the optimal
architecture, $A^*_{D, fr}$, that gives the best performance.

To obtain optimal weights $W^*_{D, fr}$ for each model, we partition
our data into training, validation, and testing sets, denoted as
$\mathcal{F}$, $\mathcal{V}$, and $\mathcal{T}$, respectively. We use
the training set $\mathcal{F}$ to compute the fit, as given by:
equation \ref{eq:train}.

\begin{equation}
  W^*_{D, fr} = \arg\min_{W_{D, fr}}F(W_{D, fr}, A_{D, fr}, \mathcal{F}).
  \label{eq:train}
\end{equation}

To prevent overfitting, we use the validation
set $\mathcal{V}$ to perform early stopping with a patience of 5
epochs based on validation loss. We train the model using the training
set for a minimum of 50 epochs and set the maximum number of epochs to
100. However, training typically stops around 60 epochs, as we rarely
need the full 100 epochs.

After training each model, we obtain the optimal weights for each
model, denoted as $W^*_{D, fr}$. We select the optimal model along
with its corresponding optimal weights based on the model's
performance on the validation set, denoted as $\mathcal{V}$, as given
by: equation \ref{eq:opt-model}.

\begin{equation}
  A^*_{D, fr} = \arg\min_{A_{D, fr} \in \mathcal{A}}F(W^*_{D, fr}, A_{D, fr}, \mathcal{V}).
  \label{eq:opt-model}
\end{equation}

\subsection{Long term interactive activity
  visualization}
\label{sec:long-term-interactive-act-viz}

\begin{figure*}
  \centering  

  \begin{subfigure}{0.97\textwidth}
    \includegraphics[width=\linewidth]{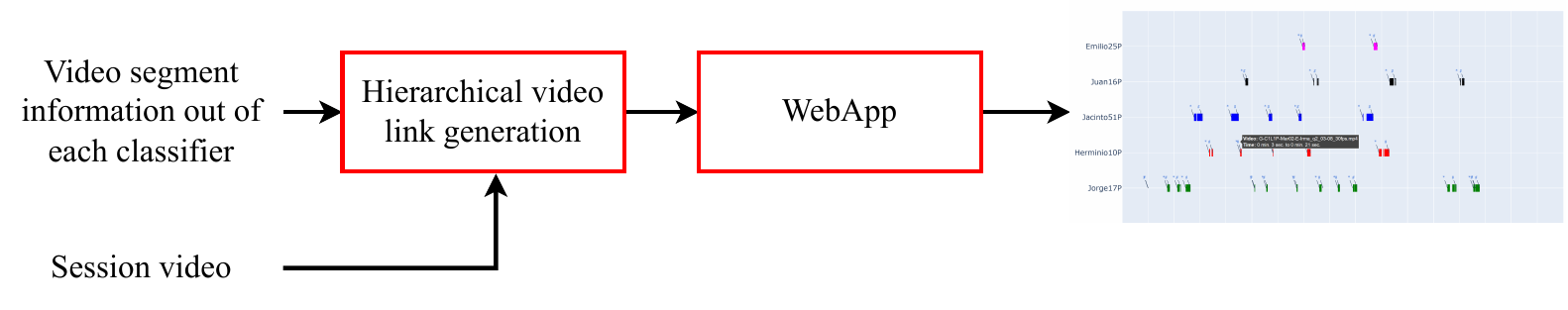}
    \caption{Interactive activity map visualization system diagram.}
    \label{subfig:int-act-sys-diag}
  \end{subfigure}

  \begin{subfigure}{0.97\textwidth}
    \includegraphics[width=\linewidth]{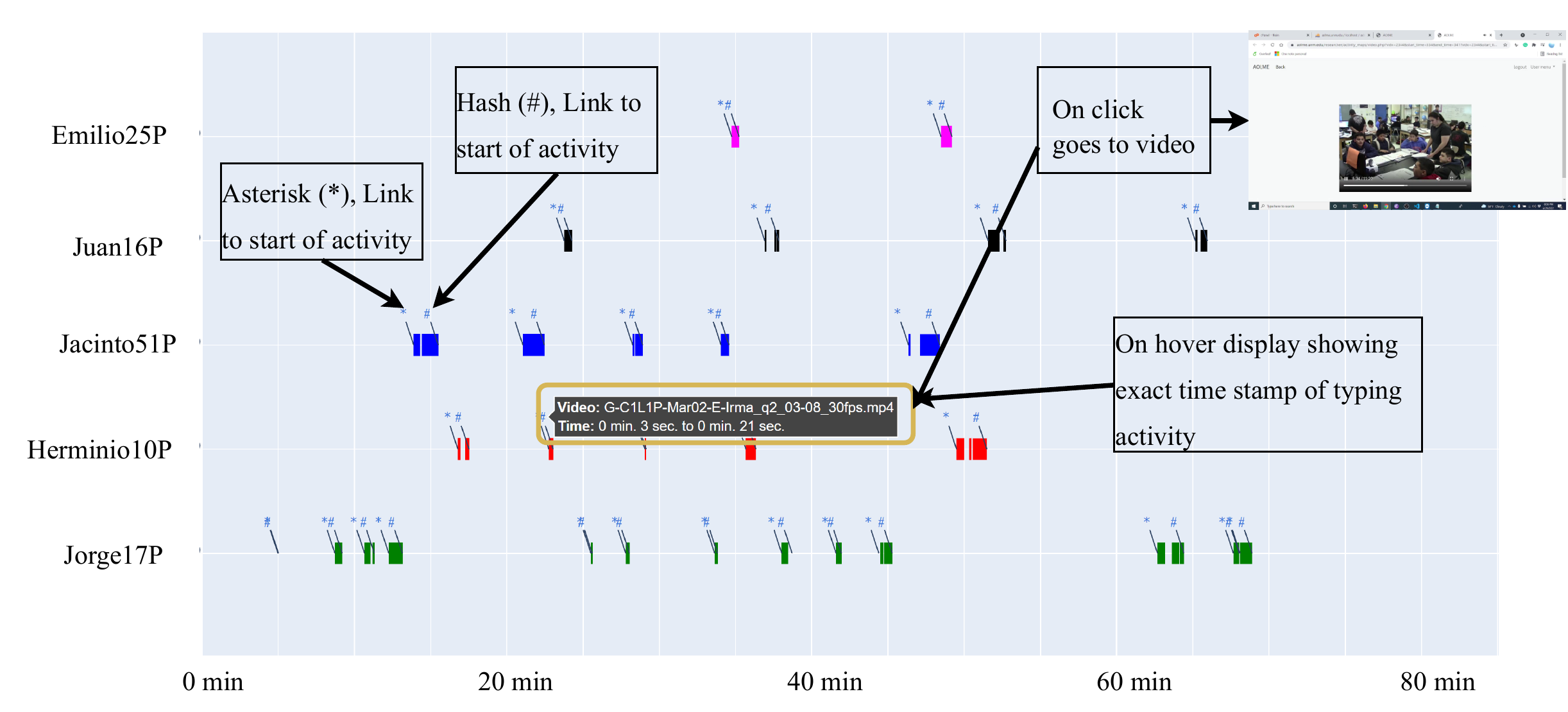}
    \caption{Interactive typing activity map for C1L1P-E, Mar 02 session. It supports ``on hover'', ``zoom'', ``selection'' and ``clickable'' events. A user can use mouse to hover over the asterisks (*) and hash (\#), which display exact location in the video. These symbols also serve as weblinks (requires AOLME account), displaying the activity in a web browser.}
    \label{subfig:act-map}
  \end{subfigure}

  \caption{\textbf{System diagram and example of interactive activity map.}}
  \label{fig:method-int-act-map}
\end{figure*}

Our activity detection system is capable of classifying small
3-second proposal regions in videos as either having typing or
writing activity or not having it. However, displaying these
activity detections in a user-friendly way is crucial for users to
gain insights and draw inferences. To provide a seamless user
experience, we design interactive activity map generation system as
shown in figure \ref{subfig:int-act-sys-diag}.

The output video activity classification, along with the person's
pseudonym, time interval, and spatial coordinates, are processed to
create web links. These web links are used to display the results of
our basic grouping of activity, which groups together activities
from the same person that are less than 3 seconds in duration.

We then plot these grouped time activities and mark the starting and
ending points with web links. When a user hovers over these links,
they can view the activity time interval. Clicking on the links
loads the video hosted on our AOLME server, allowing users to view
the activity in question.

To access the videos via web links, users must first register with the
AOLME website as the data is protected. However, once registered,
multiple users can access the activity maps at any time using only a
browser. This method of sharing has an added advantage of being easily
accessible and available to users anytime and anywhere, as long as
they have internet access.

We provide an example of our interactive activity map in Figure
\ref{subfig:act-map}. We also display activity detections in
a way that allows users to interact with the visualizations, zooming
in and out of the plot, and hovering over individual points to view
specific details. This interactivity enables users to easily explore
the activity detections and gain a better understanding of the
underlying data.

Furthermore, we integrate the activity detections with the video,
enabling users to watch the video at specific times of interest. By
integrating the activity detections and the video, users can quickly
and easily navigate to specific points in the video where activity is
detected, allowing them to see the activity in context and draw more
meaningful insights.

\section{Method training and  testing}
\label{method_training_and_testing}
This section provides an overview of the procedures and protocols
followed to train and test our video activity recognition and
visualization system. It is organized into three sections. Section
\ref{sec:trn-tst-sessions} describes the process of preparing and
partitioning the group interaction videos into testing and training
sessions. We also ensure that the testing sessions were the same when
testing different stages of the system, helping us to evaluate the
system's performance accurately. In section \ref{sec:trn-RPN}, we
explain in detail the training process for our object detector, which
involved using ground truth data from the training sessions.

Moving onto section \ref{sec:trn-act-classifier}, we discuss the
protocols used to create a dataset for training our activity
recognition system. Here, we developed a fast activity labeling
procedure that accurately labeled typing, no-typing, writing, and
no-writing in group interaction videos. We also explain the procedure
followed to create representative samples from the labeled ground
truth, which helped us to train the activity classification system. By
detailing these procedures and protocols, we hope to provide insight
into the training of our video activity recognition and visualization
system and contribute to the wider field of research in this area.

\subsubsection{Session based partitioning of group \\
  interaction videos.}
\label{sec:trn-tst-sessions}

This section provides a summary of the pre-processing of group
interaction videos. The primary objective of this pre-processing
is to standardize the video frame rate and resolution, which ensures
that the videos are consistent and comparable.

Subsequently, in section \ref{subsec:testing-sessions}, we explain why we
split the group interactons video dataset at the session level and
present the sessions that were collaboratively selected with education
researchers for testing. This approach allowed us to evaluate the
performance of our system accurately and ensured that the testing
sessions were not used at any stage of our system before testing.

\textbf{Preparing group interaction videos:}
In order to use the group interaction videos for training and testing
our system, we needed to pre-process them. The videos were captured at
a high resolution of $1920 \times 1080$ at either 30 or 60 frames per
second (FPS). However, storing, streaming, and analyzing videos at
such high quality is inefficient due to constraints in terms of
bandwidth and memory. To address this issue, we transcoded the videos
to a lower resolution of $858 \times 480$ at 30 FPS. With this we
achieved a minimum of 5 times to a maximum of 10 times video data
compression.

The transcoding process made sure that the audio quality is preserved.
This is ideal for audio-related research \cite{sanchez2021bilingual,
  Antonio2022}. In addition to video and audio researchers the
compressed videos are deliverd to educaitonal researchers through a
web application to study group interacitons,
\cite{lopezleiva2020participation, celedon2022fake,
  yanguas2022middle}. To carry out the transcoding process we utalized
ffmpeg library \cite{ffmpeg}.  The process is finely tuned through
specific commands to meet research requirements. These commands
facilitate the adjustment of video and audio parameters, such as
resolution, bitrates, keyframe insertion \cite{esakki2021adaptive} and
frame rates, to create optimized outputs that balance the need for
streaming and video analysis. Please refer to Appendix
\ref{sec:video_compression} for more details.

\textbf{Collaborative selection of testing sessions:\label{subsec:testing-sessions}}
Section \ref{sec:bg-AOLME} illustrates that we can partition the group
interaction videos into training and testing datasets based on video
intervals, session, or group. We chose to split the dataset at the
session level since the AOLME curriculum teaches a concept within one
session, and this session could involve typing or writing activities
as the primary activity. For example, during the the initial phase of
the project, students may use paper and color pencils to design before
implementing it programmatically. This implies that
writing is the primary activity at the beginning of the project phase
while typing becomes primary towards the end. By dividing our dataset
at the session level, we could select sessions that were focused on
typing or writing, enabling us to accurately test these activities.

Moreover, a session has the advantage of having consistent lighting,
seating arrangements, and student attire.  Splitting the dataset using
video intervals could result in training and testing datasets that are
too similar to each other. A system that performs well on this type of
data splitting may not perform well on new videos. Therefore, to
ensure that our system is robust and performs well on new sessions, we
decided to use session-based splitting.

To enhance the relevance and accuracy of our evaluation, we
collaborated closely with the education department to carefully select
the group interaction sessions used for testing. This collaborative
effort led to the identification of sessions that were of particular
interest to the education department and are summarized in Table
\ref{tab:edu-important-sessions}. We then evaluated our complete
system, as outlined in Section \ref{sec:system}, using these
sessions. By doing so, we were able to fine-tune our framework and
optimize its performance on similar sessions, thereby improving its
robustness and accuracy. In summary, this collaborative approach
ensured that our research had practical applications and implications,
making it more impactful and meaningful.

\begin{table}[!t]
  \centering
  \caption{\textbf{ A table presenting AOLME small group interaction sessions
      identified as important by education researchers. We chose our
      training dataset representative of the identified sessions. For
      example, we chose multiple sessions from cohort 1, level 1, group
      C which is given great importance by education researchers.}}
  \label{tab:edu-important-sessions}
  \begin{tabular}{l l l l}
    \hline
    \textbf{Group} & \textbf{Date} & \textbf{\# students} & \textbf{Duration} \\
    \hline                                                   
    \hline                                                   
    C1L1P-B        & Mar 02        & 4                    & 1 hr. 22 min.     \\
    C1L1P-C        & Mar 30        & 4                    & 1 hr. 36 min.     \\ 
    C1L1P-C        & Apr 13        & 4                    & 1 hr. 43 min.     \\
    C1L1P-C        & Apr 06        & 4                    & 1 hr. 28 min.     \\
    C1L1P-E        & Mar 02        & 5                    & 1 hr. 25 min.     \\
    C2L1P-B        & Feb 23        & 5                    & 1 hr. 38 min.     \\
    C2L1P-C        & Apr 12        & 4                    & 1 hr. 56 min.     \\
    C2L1P-D        & Mar 08        & 3                    & 1 hr. 36 min.     \\
    C2L1P-E        & Apr 12        & 4                    & 1 hr. 51 min.     \\
    C2L1W-B        & Feb 27        & 4                    & 1 hr. 23 min.     \\
    C3L1P-C        & Apr 11        & 5                    & 1 hr. 37 min.     \\
    C3L1P-D        & Feb 21        & 4                    & 1 hr. 36 min.     \\
    C3L1W-D        & Mar 19        & 3                    & 1 hr. 21 min.     \\
    \hline
  \end{tabular}
\end{table}

\subsubsection{Activity region initialization and  labeling
  procedures }
\label{sec:ty-wr-labeling}

This section describes the procedures utilized for labeling activities
(typing, writing) and activity regions. The primary emphasis of the
design is to expedite the labeling process. To achieve this
objective, we have implemented a two-pass approach. Firstly, we
observe a session at a very high playback speed to generate time
intervals that identify inactive regions, such as students not present
in video and camera transitions (zooming, panning, and changes in
location). Secondly, we employ these timestamps to label the
activities at a speed of 30x, which equates to labeling one frame
every second. This approach has led to significant improvements in
labeling efficiency, reducing the time required to initialize and
label activities in a 1-hour video from 12 hours to 1.5 hours, while
simultaneously maintaining the quality of ground truth labels.

Sections \ref{sec:act-reg-init} and \ref{sec:ty-wr-labeling-procedure}
describe the activity region initialization and activity (typing and
writing) labeling procedures respectively. In describing the procedures
we use the symbol $S$ to represent group interaction sessions that
necessitate labeling. We have divided these sessions into two distinct
groups: (1) sessions that necessitate labeling for the entire duration
of the session, referred to as $S^*$, and (2) sessions from which we
may select 10 to 15 minutes to label. The first group ($S^*$) is used
for validating and testing our framework and primarily comprises
sessions identified by education researchers, as illustrated in Table
\ref{tab:edu-important-sessions}. On the other hand, the second group
comprises representative samples of the first group and is mainly
utilized for training purposes.

\textbf{Activity region initialization \label{sec:act-reg-init}}\\

\begin{figure*}[t]
  \centering  

  \begin{subfigure}{0.95\linewidth}
    \includegraphics[width=\linewidth]{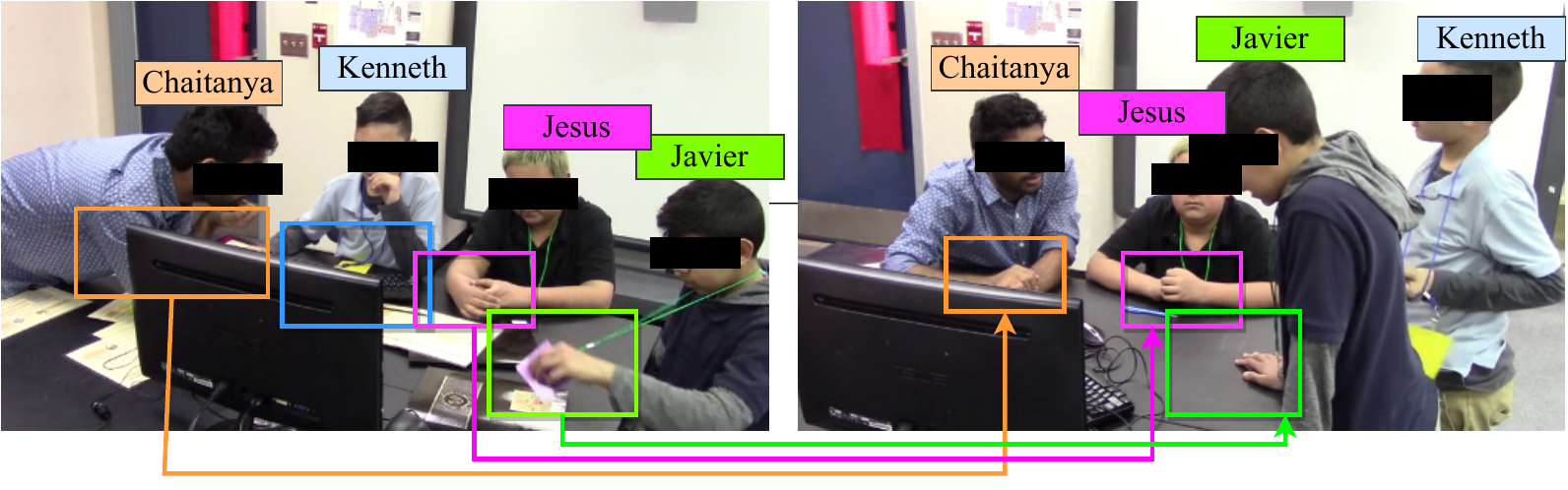}
    \caption{Evolution of activity regions in 8 seconds. As shown by the labeled table regions, the pink region remained consistent over time, the brown and green regions changed in shape and position. Additionally, the blue region assigned to Kenneth was not present in the later image, suggesting that he did not use the table for more than five minutes.}
    \label{fig:activity-region-evolves-a}
  \end{subfigure}

  \begin{subfigure}{0.95\linewidth}
    \includegraphics[width=\linewidth]{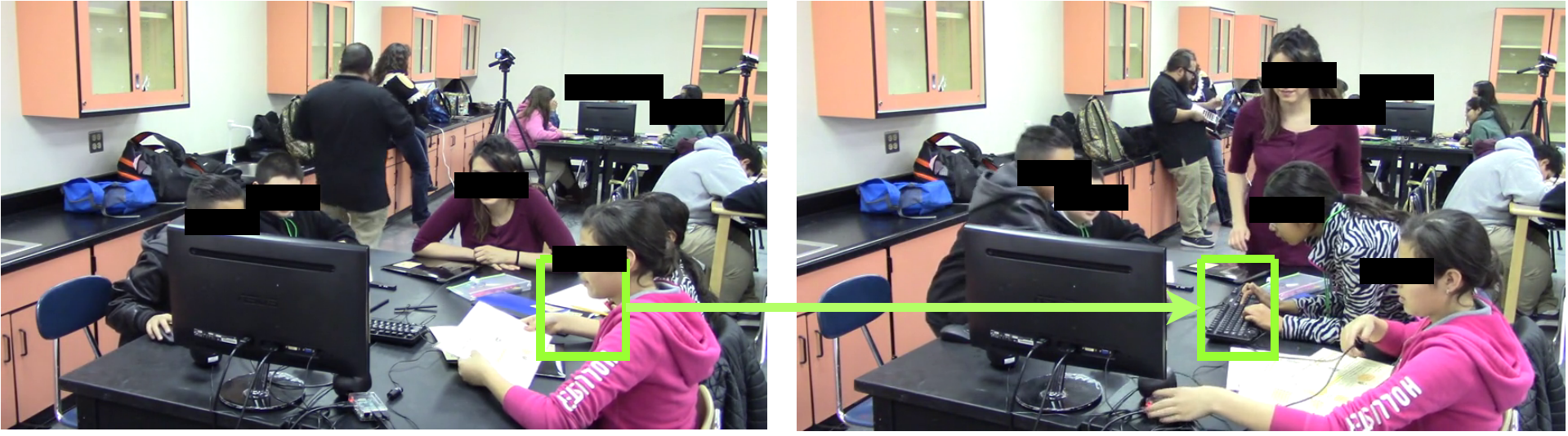}
    \caption{A student leans into the keyboard for typing, moving the activity region towards the keyboard as shown with the green arrow.}
    \label{fig:activity-region-evolves-b}
  \end{subfigure}

  \caption{Depicting the evolution of activity initializations in AOLME group interaction videos.}
  \label{fig:activity-region-evolves}
\end{figure*}

\begin{figure}[!h]
  \includegraphics[width=0.47\textwidth]{ 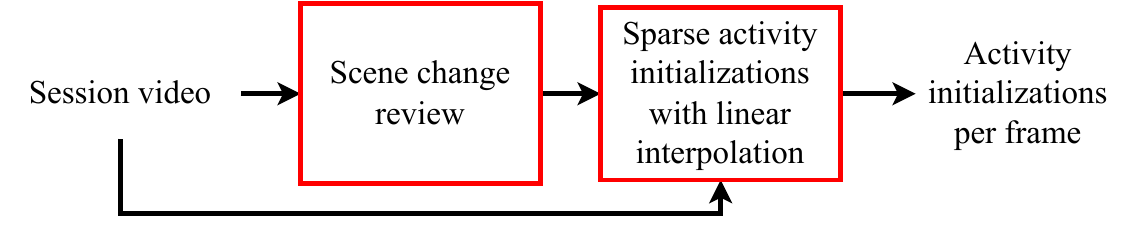 }
  \caption{\textbf{Activity region initilization using two passses.}}
  \label{fig:act-init-sys-diag}
\end{figure}

The purpose of this section is to provide a technical outline of the
procedure for initializing activity regions in the context of group
interaction videos. The goal is to initialize the video with
rectangular regions that are labeled with the name of the person
sitting closest to it. This initialization process will occur in two
passes as depicted in Figure \ref{fig:act-init-sys-diag}.

The initial review process for labeling the ground truth involves a
quick review of the session video to identify time intervals without
changes in camera angles or individuals' seating arrangements. Camera
adjustments are typically made 3 to 4 times throughout a session,
while seating arrangements tend to remain constant. A detailed
procedure is provided in Figure \ref{fig:act-reg-init}. Typically,
the first pass takes around 1 to 2 minutes per session to complete,
based on our experience.

The second pass of the labeling process involves reviewing the session
video at a faster playback rate of 30 times the normal speed. Although
seating arrangements generally remain consistent, activity regions can
shift over time. For example, if a student leans in to type on a
keyboard, the activity region will move towards the keyboard as
depicted in figure figure \ref{fig:activity-region-evolves-b}. To
address this, we mark the start and end of each activity and use
linear interpolation to label the frames in between. This method
allows us to accurately label activity regions without expending
excessive time and effort on every second of the video. Overall, our
efficient and accurate labeling process enables us to create
high-quality training datasets for our video activity recognition
system. We use Figures \ref{fig:act-reg-init} and
\ref{fig:act-reg-init-procedure-second-pass} to further expand on our
activity region initialization procedure.

\textbf{Typing and writing activity labeling
  procedure \label{sec:ty-wr-labeling-procedure}}\\

\begin{figure}[!h]
  \includegraphics[width=0.95\linewidth]{ 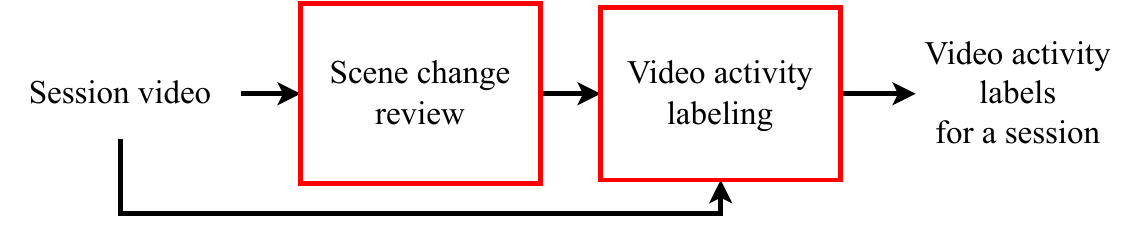}
  \caption{\textbf{Activity labeling procedure using two passses.}}
  \label{fig:act-labels-sys-diag}
\end{figure}

This section outlines the procedure for labeling typing and writing in
group interaction videos. We use a two-pass approach to speed up the
activity labeling process, similar to the activity initialization
process. In the first pass, we identify time intervals where there are
no changes in camera position or seating arrangement, using the scene
change detection process outlined in the previous section and Figure
\ref{fig:gt-initial-review}. We depict our two pass activity labeling
procedure in Figure \ref{fig:act-labels-sys-diag}.

As depicted in the figure, in the second pass, we carefully identify the
typing and writing activities that meet our predefined criteria. Our
criteria require that the activity should be at least 3 seconds in
duration and 50\% visible. For typing activities, we include instances
where typing occurs on an external keyboard or a laptop
keyboard. Similarly, for writing activities, we typically observe the
use of paper and pen, but we also include instances where dry erase
boards and markers are used.

The labeling procedure is depicted in Figures
\ref{fig:act-labeling-procedure}, and we define $S$ and $S^*$ as
previously described in Section \ref{sec:ty-wr-labeling}. The output
of the labeling procedure is a set of labeled activity instances
(typing or writing), denoted by $A = \{a_0, a_1, a_2, \dots,
a_n\}$. Each element in $A$, $a_i$, corresponds to an activity, and we
mark the spatiotemporal information by utilizing frame numbers,
rectangular coordinates, activity label, and the pseudonym of the
person performing the activity. Each activity instance, $a_i$ has same
spatial location. However, an issue arises with no-writing and
no-typing activities, as they do not belong to a specific person. In
this case, we utilize \textit{``Kidx''} instead of a pseudonym.

\subsubsection{Training keyboard and hand detector}
\label{sec:trn-RPN}

Our video activity proposal network relies on accurate identification
of regions containing keyboards and hands. To achieve this, we use the
Faster-RCNN object detection framework \cite{ren2016faster} to detect
hands and keyboards in the video. The framework is trained on video frames
(images) extracted from the same sessions used to train
the activity classifier. We used two different approaches to create
ground truth data for keyboards and hands from the typing and writing
ground truth data, respectively.

To create keyboard detection dataset, we extract two frames with
corresponding bounding boxes every minute from the typing and
no-typing ground truth. This approach is effective for three main
reasons: (1) there is only one keyboard present per group, (2) the
no-typing instances also include the keyboard, and (3) keyboards in
the background are not visible. We extract frames from a total of
33, 4, and 7 sessions for training, validation, and testing,
respectively, resulting in 700, 100, and 648 keyboard samples.

The hand detection dataset creation differs from the keyboard
detection dataset creation. We cannot use the activity labels from
writing and no-writing for hand detection, as they do not label all
hand instances in the video frames. This results in unlabelled hand
instances in the background and improper training of the object
detector. To address this, we use the online labeling tool,
makesense.ai \cite{make-sense}, to label all hand instances in video
frames. We extract frames from a total of 33, 4, and 7 sessions for
training, validation, and testing, resulting in 305, 100, and 313
frames with hand labels. As there are often more than one hand per
frame, the number of hand instances are 1803, 714, and 2031 for
training, validation, and testing, respectively.

\subsubsection{Training activity classifier}
\label{sec:trn-act-classifier}

The procedures outlined in the previous section, Section
\ref{sec:ty-wr-labeling-procedure}, are utilized to manually label
typing and writing in group interaction videos. By implementing the
two-pass labeling approach, we were able to review and label a total
of 43 sessions (approximately 75 hours) and 30 sessions (approximately
50 hours) for typing and writing in less than 200 hours. Without this
approach, the labeling process would take over 1500 hours (with each
hour of labeling requiring 12 hours).

\begin{figure*}[t]
  \subfloat[Spatiotemporal samples demonstrating the variability in the size, shape and background of typing
  activity. From left to right we show (1) full keyboard visibility, (2) partial keybaord visibility, (3) typing
  on a laptop, and (4) change in background (table color).
  ]{
    \includegraphics[width=0.95\linewidth]{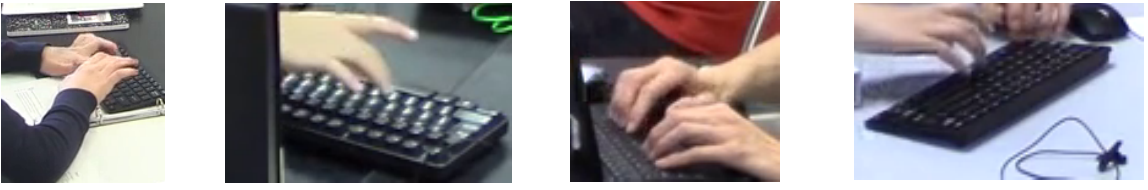} 
  }
  
  \subfloat[Spatiotemporal samples demonstrating the variability in the size, shape and background of no-typing
  activity. From left to right we show (1) no-typing with hands, (2) patial keyboard visibility, (3) full keyboard
  visibility, and (4) change in background (table color).
  ]{
    \includegraphics[width=0.95\linewidth]{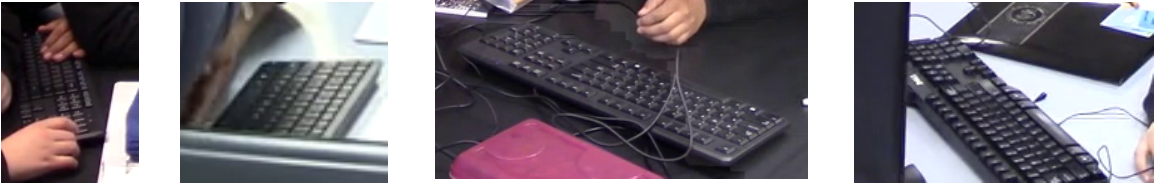}
  }

  \subfloat[Spatiotemporal samples demonstrating the variability in the size, shape and backgorund of writing
  activity. From left to right we show (1) full hand visibility, (2, 3 and 4) partial hand visibility, and
  (5) writing on dry erase board.
  ]{
    \includegraphics[width=0.95\linewidth]{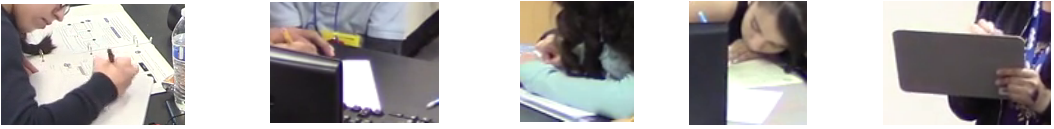} 
  }
  
  \subfloat[Spatiotemporal samples demonstrating the variability in the size, shape and backgorund of no-writing
  activity. From left to right we show (1 and 2) presense of hand and paper, (3) presense of pen and paper and
  (4) presense of hand holding the pen and paper.
  ]{
    \includegraphics[width=0.95\linewidth]{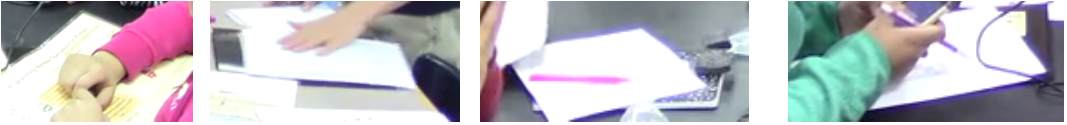} 
  }
  \caption{\textbf{A figure illustrating the spatiotemporal regions extracted
      from the ground truth labels for typing, no-typing, writing, and
      no-writing, which demonstrates the variability in size, shape and
      occlusion.}}
  \label{fig:trimmed-videos}
\end{figure*}

In summary, we generated a total of 627 typing (266 minutes), 645
no-typing (694 minutes), 1199 writing (480 minutes), and 798
no-writing (1440 minutes) spatiotemporal samples, as presented in
Tables \ref{tab:gt-typing} and \ref{tab:gt-writing}. After examining
the ground truth labels, we discovered that the minimum typing and
writing samples have a duration of 3.03 and 3.13 seconds,
respectively. This has led us to design our classifiers utilizing
3-second video samples (refer to Section
\ref{sec:sampling-procedure}). Figure \ref{fig:trimmed-videos}
demonstrates the variability in the samples with respect to occlusion,
background and camera position.

The remainder of this section is organized into two sections. In
Section \ref{sec:act-class-data-part}, we present the partitioning of
the samples into training, validation, and testing sets. Following
this section, we describe the additional data cleanup strategies that
we implemented to extract clean and representative training samples from
the activity labels in section \ref{sec:sampling-procedure}.

\textbf{Activity classificatiion data partitioning \label{sec:act-class-data-part}}\\
To support robust classification at the session level, we partitioned
the ground truth labels at the session level into training,
validation, and testing sets. We also ensured that the sessions
identified as important by education researchers were primarily
utilized for validation and testing purposes. We summarize the data
splitting in Tables \ref{tab:gt-typing} and \ref{tab:gt-writing}.

\begin{table*}[t]
  \centering
  \setlength{\tabcolsep}{0.13cm}
  \caption{\textbf{The following table summarizes the ground truth labels for
      typing and non-typing activities. The sessions that possess ground
      truth for the entire duration are denoted in boldface. We use a
      \colorbox{lightyellow}{yellow} and \colorbox{green}{green}
      background to label sessions utilized for
      \colorbox{lightyellow}{validation} and \colorbox{green}{testing},
      respectively.}}
  \label{tab:gt-typing}
  \begin{tabular}{p{2.3cm} p{4.5cm} l l l l}
    \hline
    \textbf{Group} & \textbf{Dates}                                                        & \multicolumn{2}{c}{\textbf{No. Samples}} & \multicolumn{2}{c}{\textbf{Duration}}                       \\
                   &                                                                       & \textbf{typing}                          & \textbf{no-typing} & \textbf{typing}   & \textbf{no-typing} \\
    \hline
    \hline
    
    C1L1P-A             & Apr. 06, Apr. 13, Feb. 16, Mar. 02, \colorbox{lightyellow}{Mar. 09}, \colorbox{lightyellow}{Apr. 20}, Feb. 25                                                                               & 33           & 39           & 17 min.           & 37 min.           \\
    C1L1P-B             & Apr. 27, Mar. 09, May 06, Mar. 02, Mar. 30, \colorbox{lightyellow}{Apr. 06}, May 11, May 04                                                                                               & 55           & 40           & 28 min.           & 67 min.           \\
    C1L1P-C             & \colorbox{lightyellow}{Feb. 25}, \colorbox{lightyellow}{Mar. 09}, \colorbox{lightyellow}{Apr. 20}, \textbf{May 04}, \textbf{Apr. 13}, Mar. 02, \colorbox{green}{\textbf{Mar. 30}}, Feb. 16 & 132          & 84           & 74 min.           & 78 min.           \\
    C1L1P-D             & Apr. 06, \colorbox{lightyellow}{Mar. 09}                                                                                                                                                  & 7            & 0            & 3 min.            & 0 min.            \\
    C1L1P-E             & Feb. 25, \colorbox{green}{\textbf{Mar. 02}}                                                                                                                                          & 51           & 50           & 23 min.           & 39 min.           \\
    C1L1W-A             & Feb. 28, Mar. 28, \colorbox{lightyellow}{Feb. 21}, Apr. 25, Mar. 07                                                                                                                       & 30           & 18           & 23 min.           & 26 min.           \\
    C1L1W-B             & \colorbox{lightyellow}{May 06}                                                                                                                                                            & 9            & 2            & 6 min.            & 2 min.            \\
    C1L1W-C             & Feb. 21                                                                                                                                                                                 & 3            & 2            & 2 min.            & 6 min.            \\
    C1L1W-D             & Feb. 28                                                                                                                                                                                 & 4            & 2            & 3 min.            & 2 min.            \\
    C2L1P-B             & \colorbox{green}{\textbf{Feb. 23}}                                                                                                                                                   & 56           & 67           & 12 min.           & 82 min.           \\
    C2L1P-C             & Apr. 12                                                                                                                                                                                 & 2            & 5            & 1 min.            & 11 min.           \\
    C2L1P-D             & \textbf{Mar. 08}                                                                                                                                                                        & 58           & 52           & 18 min.           & 1 min.            \\
    C2L1W-A             & Apr. 10                                                                                                                                                                                 & 6            & 0            & 1 min.            & 0 min.            \\
    C2L1W-B             & \textbf{Feb. 27}                                                                                                                                                                        & 75           & 80           & 24 min.           & 49 min.           \\
    C3L1P-C             & \textbf{Apr. 11}                                                                                                                                                                        & 13           & 43           & 6 min.            & 95 min.           \\
    C3L1P-D             & \colorbox{green}{\textbf{Feb. 21}}                                                                                                                                                   & 44           & 89           & 9 min.            & 69 min.           \\
    C3L1W-D             & \textbf{Mar. 19}                                                                                                                                                                        & 49           & 72           & 10 min.           & 66 min.           \\
    \hline
    \textbf{Training}   & \textbf{30}                                                                                                                                                                             & \textbf{405} & \textbf{398} & \textbf{180 min.} & \textbf{454 min.}      \\
    \textbf{Validation} & \textbf{9}                                                                                                                                                                              & \textbf{72}  & \textbf{45}  & \textbf{44 min.}  & \textbf{50 min.}  \\
    \textbf{Testing}    & \textbf{4}                                                                                                                                                                              & \textbf{150} & \textbf{202} & \textbf{42 min.}  & \textbf{190 min.} \\
    \textbf{Total}      & \textbf{43}                                                                                                                                                                             & \textbf{627} & \textbf{645} & \textbf{266 min.} & \textbf{694 min.} \\
    \hline
    \hline
  \end{tabular}
\end{table*}

\begin{table*}[t]
  \centering
  \setlength{\tabcolsep}{0.09cm}
  \caption{\textbf{The following table summarizes the ground truth labels for
      writing and non-writing activities. The sessions that possess ground
      truth for the entire duration are denoted in boldface. We use a
      \colorbox{lightyellow}{yellow} and \colorbox{green}{green}
      background to label sessions utilized for
      \colorbox{lightyellow}{validation} and \colorbox{green}{testing},
      respectively.}}
  \label{tab:gt-writing}
  \begin{tabular}{p{2.3cm} p{4.5cm} l l l l}
    \hline
    \textbf{Group} & \textbf{Dates}                                                 & \multicolumn{2}{c}{\textbf{No. Samples}} & \multicolumn{2}{c}{\textbf{Duration}}                        \\
                   &                                                                & \textbf{writing}                         & \textbf{no-writing} & \textbf{writing} & \textbf{no-writing} \\
    \hline
    \hline
    
    C1L1P-B             & Mar. 03                                                                                                                                                                                         & 57            & 75           & 30 min.           & 170 min.           \\
    C1L1P-C             & \colorbox{green}{\textbf{Mar. 30}}, \colorbox{lightyellow}{Apr. 06}, \textbf{Apr. 13}, Feb. 16, \colorbox{lightyellow}{Feb. 25}, \colorbox{lightyellow}{Mar. 09}, \textbf{Apr. 20}, May 04, May 11 & 364           & 165          & 133 min.          & 461 min.           \\
    C1L1P-D             & Mar. 09, \colorbox{lightyellow}{Mar. 02}, \colorbox{lightyellow}{\textbf{Mar. 30}}, Apr. 06                                                                                                         & 127           & 1            & 45 min.           & 8 min.             \\
    C1L1P-E             & \colorbox{green}{\textbf{Mar. 02}}                                                                                                                                                           & 60            & 140          & 52 min.           & 216 min.           \\
    C1L1W-A             & Feb. 14, \colorbox{lightyellow}{Feb. 21}, Feb. 28, Apr. 04                                                                                                                                        & 155           & 0            & 33 min.           & 0 min.             \\
    C2L1P-B             & \colorbox{green}{Feb. 23}                                                                                                                                                                    & 17            & 56           & 8 min.            & 170 min.           \\
    C2L1P-C             & Apr. 12                                                                                                                                                                                         & 88            & 35           & 58 min.           & 73 min.            \\
    C2L1P-D             & \textbf{Mar. 08}                                                                                                                                                                                & 14            & 2            & 4 min.            & 1 min.             \\
    C2L1P-E             & Apr. 12                                                                                                                                                                                         & 38            & 128          & 19 min.           & 129 min.           \\
    C2L1W-A             & Feb. 20, Apr. 10                                                                                                                                                                                & 116           & 0            & 20 min.           & 0 min.             \\
    C2L1W-B             & Feb. 27                                                                                                                                                                                         & 11            & 0            & 3 min.            & 0 min.             \\
    C3L1P-C             & \colorbox{green}{\textbf{Apr. 11}}                                                                                                                                                           & 109           & 176          & 50 min.           & 164 min.           \\
    C3L1P-D             & Feb. 21, Feb. 14                                                                                                                                                                                & 25            & 16           & 8 min.            & 69 min.            \\
    C3L1W-D             & \textbf{Mar. 19}                                                                                                                                                                                & 18            & 4            & 9 min.            & 11 min.            \\
    \hline
    \textbf{Training}   & \textbf{20}                                                                                                                                                                                     & \textbf{727}  & \textbf{311} & \textbf{261 min.} & \textbf{590 min.}  \\
    \textbf{Validation} & \textbf{6}                                                                                                                                                                                      & \textbf{189}  & \textbf{89}  & \textbf{74 min.}  & \textbf{186 min.}  \\
    \textbf{Testing}    & \textbf{4}                                                                                                                                                                                      & \textbf{283}  & \textbf{398} & \textbf{145 min.} & \textbf{664 min.}  \\    
    \textbf{Total}      & \textbf{30}                                                                                                                                                                                     & \textbf{1199} & \textbf{798} & \textbf{480 min.} & \textbf{1440 min.} \\
    \hline
    \hline

  \end{tabular}
\end{table*}

\begin{figure}[t]
  \centering
  \includegraphics[width=\linewidth]{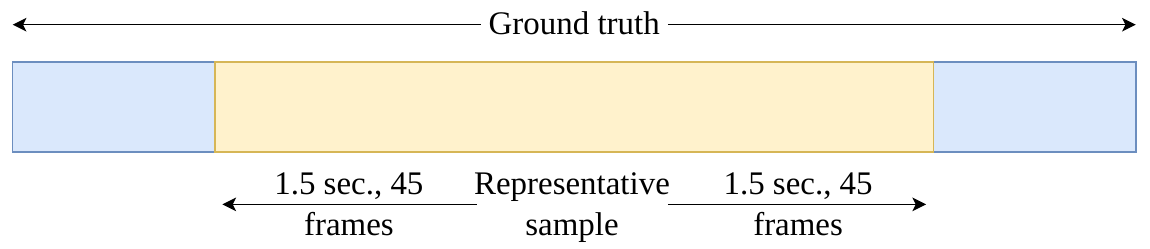}
  \caption{\textbf{Representative sample extraction form activity labels. Here
      we show the process of extracting representative sample from
      ground truth labels.}}
  \label{fig:representative-sample-extraction}
\end{figure}

\textbf{Sampling procedure \label{sec:sampling-procedure}}\\
The labeled samples presented in Tables \ref{tab:gt-typing} and
\ref{tab:gt-writing} can vary in duration from a minimum of 3 seconds
to a maximum of 284 seconds. If we extract our training dataset by
temporally segmenting the ground truth at every 3 seconds, we will
have a total of 3600 typing, 9080 no-typing, 5220 writing, and 11800
no-writing samples. However, typically, all 3-second segments
extracted from the ground truth have very similar features. To
expedite the training process, we extract a representative 3-second
sample from the activity labels, with the sample being extracted from
the middle, as illustrated in Figure
\ref{fig:representative-sample-extraction}. This method of extracting
representative samples not only speeds up the training process, but
also prevents the inclusion of the beginning and end of ground truth
labels, which generally do not contain the activity of interest.

After the extraction of representative samples, we conduct a cleaning
process to ensure that the samples accurately represent the relevant
activity instances. Any samples that do not exhibit proper activity of
interest are removed. Following this cleaning process, we are left
with a total of 324 typing, 320 no-typing, 407 writing, and 191
no-writing representative samples.

\begin{figure}
  \begin{algorithmic}[1]
    \Procedure{Scene change review}{}
      \State \COMMENT \textbf{Input:} \textit{Video session and time interval.}
      \State \COMMENT \textbf{Output:} \textit{Active time intervals, $T^*$.}
      \State
      \State $i = 0$
      \State Initialize to no time-stamps $T^* = \{\}$
      \While{reviewing the video at $> 300 \times$ speed}
        \State Mark starting time stamp $T^*_{s_i}$
        \State Mark ending time stamp $T^*_{e_i}$
        \State Add $(T^*_{s_i}, T^*_{e_i})$ to $T^*$
        \State $i = i + 1$
      \EndWhile
      \State
      \State \Return $T^* = \{(T^*_{s0}, T^*_{e0}), (T^*_{s1}, T^*_{e1}), \dots, (T^*_{sp}, T^*_{ep})\}$
    \EndProcedure
  \end{algorithmic}
  \caption{\textbf{Initial review is conducted at a very high playback speed.
      The primary objective of this stage is to identify the timestamps
      that require labeling.}}
  \label{fig:gt-initial-review}
\end{figure}

\begin{figure}[]
  \begin{algorithmic}[1]
    \Procedure{Activity initializations}{}
      \State \COMMENT \textbf{Input:} \textit{Session video, $S_i$, with selected time interval $T$}
      \State \COMMENT \textbf{Output:} \textit{Activity initializations per frame, $AI = [AI_{1}, AI_{1}, \dots, AI_{m}]$ for}
      \State \COMMENT \hspace{1.8cm}\textit{all frames. Where, $AI_{m}$ contains rectangular coordinates and}
      \State \COMMENT \hspace{1.9cm}\textit{corresponding person pseudonym of $m$th frame.}
      \State
      \State $T^*$ $=$ \Call{Scene change review}{$S_i$, $T$}
      \State $m = 0$
      \For{each scene change in $S_i$}
        \If {first frame}
          \State $AI_0$ $=$ \textbf{Annotate} table region.
        \ElsIf{Camera change \textbf{or} Person position change}
          \State $AI_m$ $=$ \textbf{Annotate} table region.
        \Else
          \State $AI_m = \{\}$, \COMMENT \textit{Skip table labeling} \label{step:skipped-manual-table-labeling}
        \EndIf
        \State $m = m + 1$
        \State $AI$ $=$ Use \textbf{liner interpolation} to evolve the size and shape of
        \State \hspace{1.2cm} bounding boxes.
      \EndFor
      \State
      \State \Return $AI$
    \EndProcedure
  \end{algorithmic}
  \caption{\textbf{The procedure for annotating table regions. We reduce
      manual labeling time by updating labels only when there is a
      change in camera position or people in the group.}}
  \label{fig:act-reg-init-procedure-second-pass}
\end{figure}

\begin{figure}[]
  \begin{algorithmic}[1]
    
    \State \COMMENT \textbf{Input:} \textit{Session, $S_i$.}
    \State \COMMENT \textbf{Output:} \textit{Activity initializations of session $S_i$, $AI_i$}
    \State
    \For{each $S_i$ is a testing set}
      \If{$S_i \in S^*$}
        \State \COMMENT \textit{Provide ground truth for entire session.}
        \State $dur =$ duration of $S_i$.
        \State $T = \{(0, dur)\}$
        \State $AI_i$ $=$ \Call{Activity initializations}{$S_i$, $T$}
      \Else
        \State \COMMENT \textit{Labeling 10 to 15 minutes in a session.}
        \State $T$ $=$ \textbf{Sample 10 to 15 minutes intervals} from $S_i$ with
        \State \hspace{0.9cm}typing and writing activities.
        \State $AI_i$ $=$ \Call{Activity initializations}{$S_i$, $T$}
      \EndIf
    \EndFor
    \State \Return $AI_i$
  \end{algorithmic}
  \caption{\textbf{Activity region initialization
      procedure.}}
  \label{fig:act-reg-init}

\end{figure}

\begin{figure}[t]
  \begin{algorithmic}[1]
    \Procedure{Video activity labeling}{$S_i$, $T$}
      \State \COMMENT \textbf{Input:} \textit{Session, $S_i$, and corresponding time intervals $T$.}
      \State \COMMENT \textbf{Output:} \textit{A set of spatiotemporal activity labels, $A$, within the time}
      \State \COMMENT \phantom{\textbf{Output:}} \textit{interval $T$.}
      \State
      \State $T^* =$ \Call{Scene change review}{T}
      \State $i = 0$
      \For{each time interval in $T^*$}
        \For{each activity in the time interval}
          \State $a_{i} =$ \textbf{Label} activity with bounding box and person pseudonym.
          \State $i = i + 1$
        \EndFor
      \EndFor
      \State
      \State \Return $A = \{a_{0}, a_{1}, a_{2}, \dots, a_{n}\}$
    \EndProcedure
  \end{algorithmic}
  \caption{\textbf{Procedure to label typing and writing activities in the
      time intervals we get after initial review (refer to figure
      \ref{fig:gt-initial-review})}}
  \label{fig:label-activity-procedure}
\end{figure}

\begin{figure}[t]
  \begin{algorithmic}[1]
    \State \COMMENT \textbf{Input:} \textit{Session, $S_i$}
    \State \COMMENT \textbf{Output:} \textit{Set of activity (typing or writing) instances,  $A=\{a_0, a_1, a_2, \dots, a_n\}$}.
    \State
    \For{each session $S_i$}
      \If{$S_i$ in testing set}
        \State \COMMENT \textit{Labeling complete duration of session.}
        \State $D_i =$ duration of $S_i$.
        \State $T \hspace{0.1cm}= \{(0, D_i)\}$
        \State $A \hspace{0.1cm}=$ \Call{Video activity labeling}{$S_i$, $T$}
      \Else
        \State \COMMENT \textit{Labeling 10 to 15 minutes in a session.}
        \State $T$ $=$ \textbf{Sample 10 to 15 minutes intervals} from $S_i$ which has
        \State \hspace{0.9cm}typing/writing activity.
        \State $A$ $=$ \Call{Video activity labeling}{$S_i$, $T$}
      \EndIf
    \EndFor
    \State
    \State \Return $A= \{a_{0}, a_{1}, a_{2}, \dots, a_{n}\}$
  \end{algorithmic}
  \caption{\textbf{The procedure for labeling typing and writing activities
      for a session involves completely labeling sessions that belong to
      the testing set. However, for sessions from the training and
      validation sets, only a 15 to 20 minute sessions is selected for
      labeling.}}
  \label{fig:act-labeling-procedure}
\end{figure}


\section{Results}
\label{results}

This section presents the results of our video activity detection
system, which was designed to detect typing and writing in group
interaction videos from AOLME. We organize this section into three
sections. First, in Section \ref{sec:results-act-prop-net}, we present
the results of the video activity proposal network. Specifically, we
provide the results of our keyboard and hand detector, as well as the
small video activity proposals we extract using detections and
activity initializations. In Section
\ref{sec:results-video-act-calss}, we compare the performance of our
low-parameter seperable activity classifiers against State of the Art
(SOTA) activity classification systems that classify typing from
no-typing and writing from no-writing. In this section we also present
the results of optimizing our family of seperable activity
classifiers.

The experiments were conducted using an Intel Xeon CPU running at 2.10
GHz and 128 GB of RAM. The system also included an Nvidia Quadro RTX
5000 GPU with 16 GB of video memory, which is considered to be
lower-end according to standard benchmarks.

\subsection{Video activity proposal network}
\label{sec:results-act-prop-net}

As described in Section \ref{sec:system}, we extracted several small
spatiotemporal regions from the session video using a combination of
object detectors (keyboard and hands), tracking, and projections,
along with activity initializations. In this section, we first
summarize the results of our keyboard detection with tracking and hand
detection with projections. We then present the results of using the
detections to filter the activity region proposals from the activity
region initializations.

\subsubsection{Keyboard tracking and hand projections results}

\begin{table*}[t]
  \centering
  \caption{\textbf{The table below shows the reduction in the number of hand
      detections achieved using our hand projection-based approach for
      removing background hand detections. The sessions used in this
      analysis are taken from training sessions and span across several
      years, demonstrating the robustness of our approach.}}
  \label{tab:pf}
  \begin{tabular}{  p {2cm}  p{2cm}  p{2cm} p{4cm} p{3cm} }
    \hline
    \textbf{Group} & \textbf{Date} & \textbf{Naive} & \textbf{Using Projections} & \textbf{\% Reduction} \\
    \hline
    \hline
    C1L1P-C        & Mar30         & 55914          & 9804                       & \textbf{82.5}         \\ 
    C1L1P-C        & Apr13         & 34665          & 8028                       & \textbf{76.8}         \\
    C1L1P-E        & Mar02         & 50312          & 9968                       & \textbf{80.0}         \\
    C2L1P-B        & Feb23         & 48073          & 9924                       & \textbf{79.3}         \\
    C2L1P-D        & Mar08         & 31875          & 7724                       & \textbf{75.7}         \\
    C3L1P-C        & Apr11         & 36757          & 9536                       & \textbf{74.0}         \\
    C3L1P-D        & Mar19         & 57319          & 9536                       & \textbf{83.3}         \\
    \hline
  \end{tabular}	
\end{table*}

\begin{figure}
  \subfloat[Successful typing region proposal using keyboard tracking when keyboard is partially visible.]{
    \includegraphics[width=0.47\linewidth]{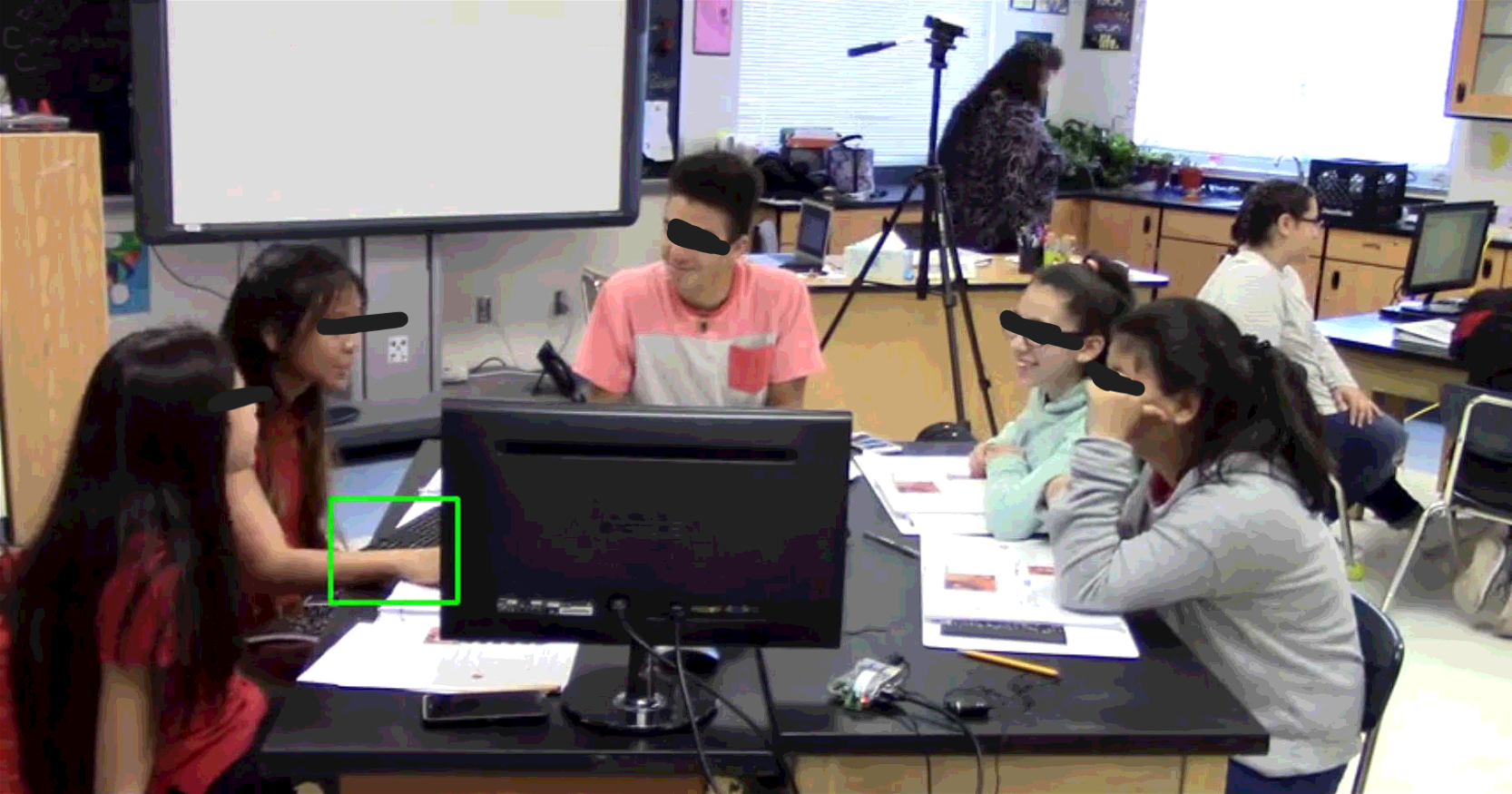}
  }~~
  \subfloat[Successful typing region proposal using keyboard tracking when keyboard is fully visible.]{
    \includegraphics[width=0.47\linewidth]{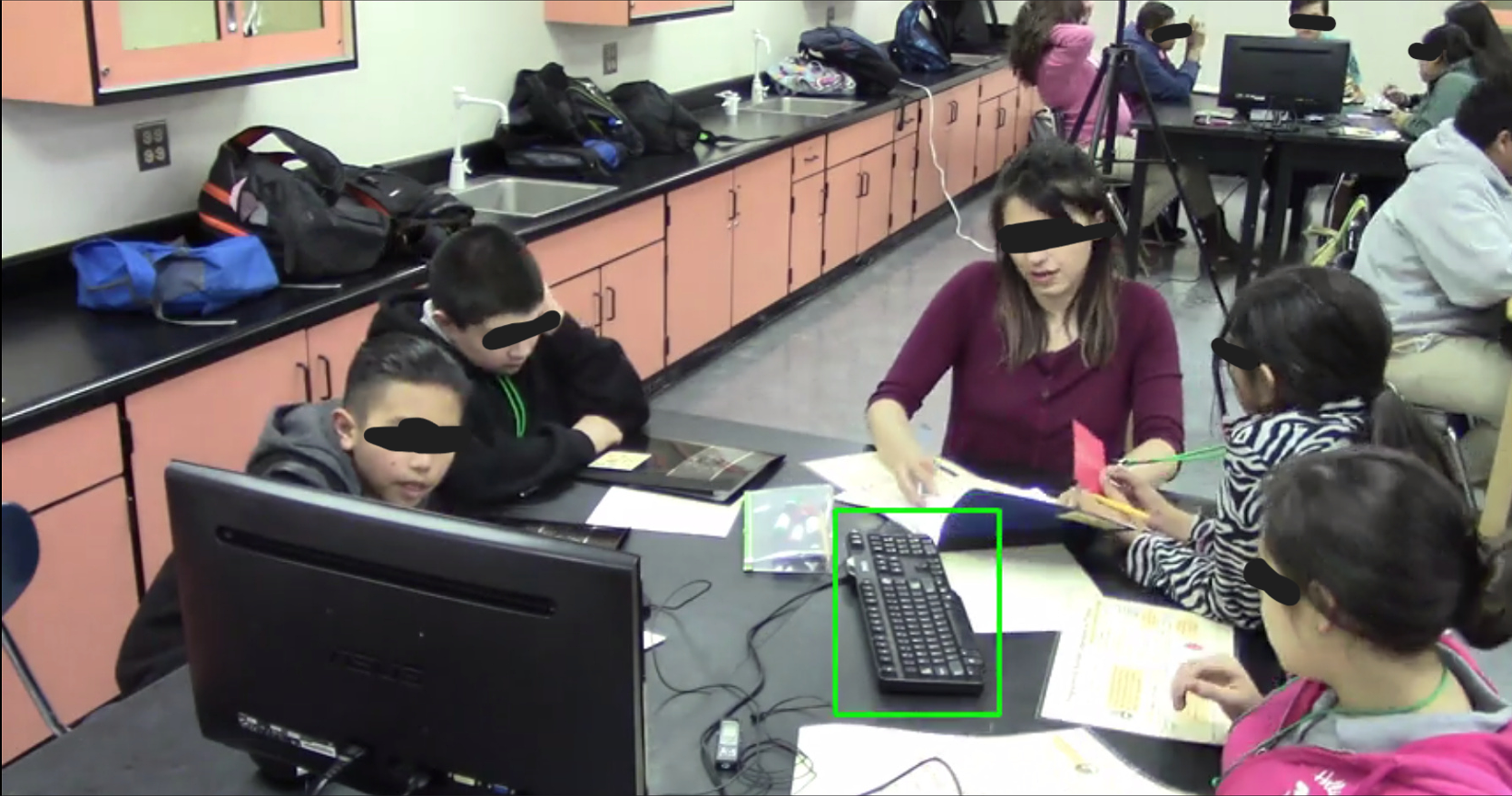}
  }
  
  \subfloat[Failure to detect typing region when keyboard is tilted and the keys are not visible.]{
    \includegraphics[width=0.47\linewidth]{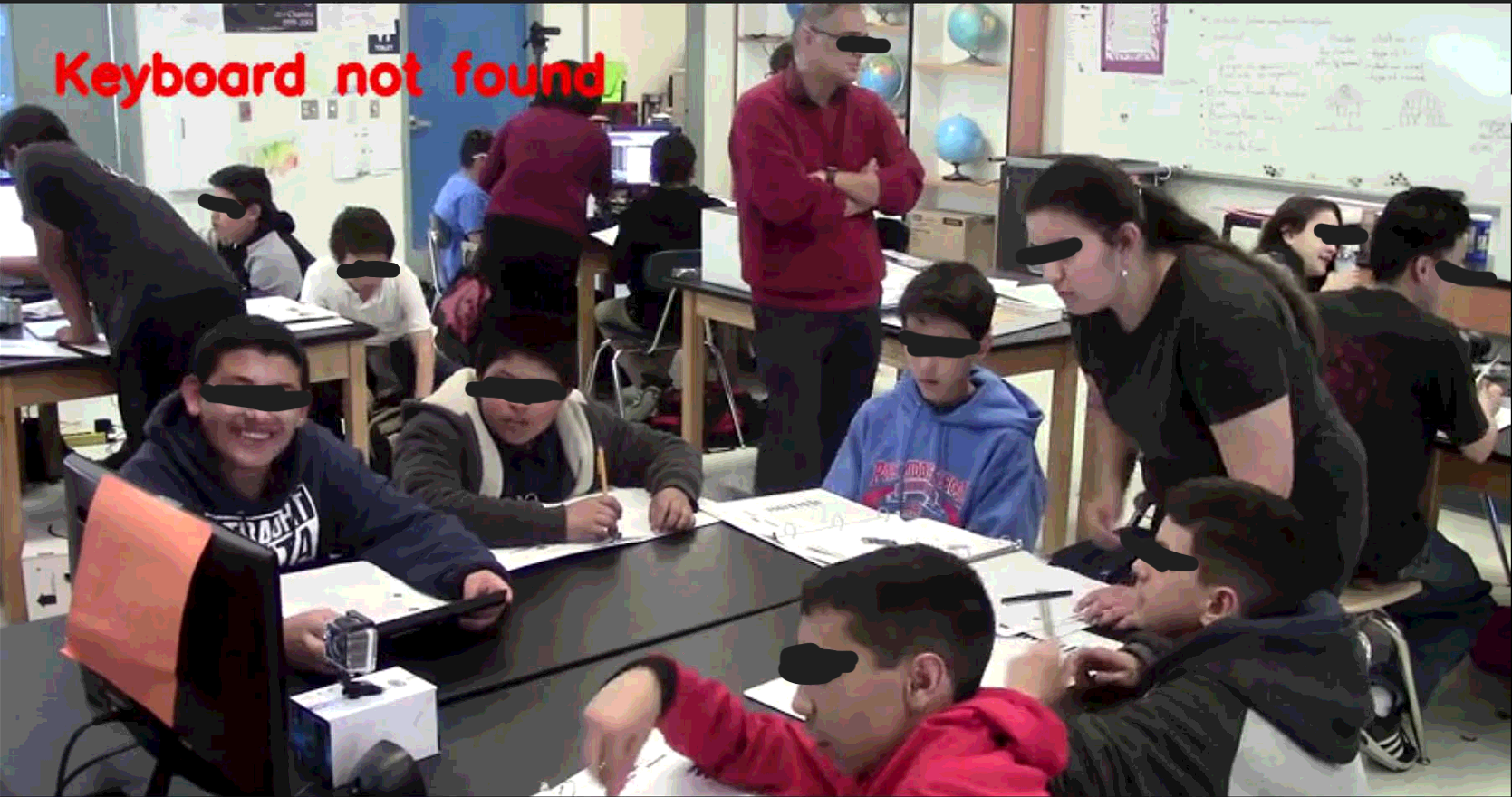}
  }~~
  \subfloat[False positive detection of book that has similar markings as keyboard.]{
    \includegraphics[width=0.47\linewidth]{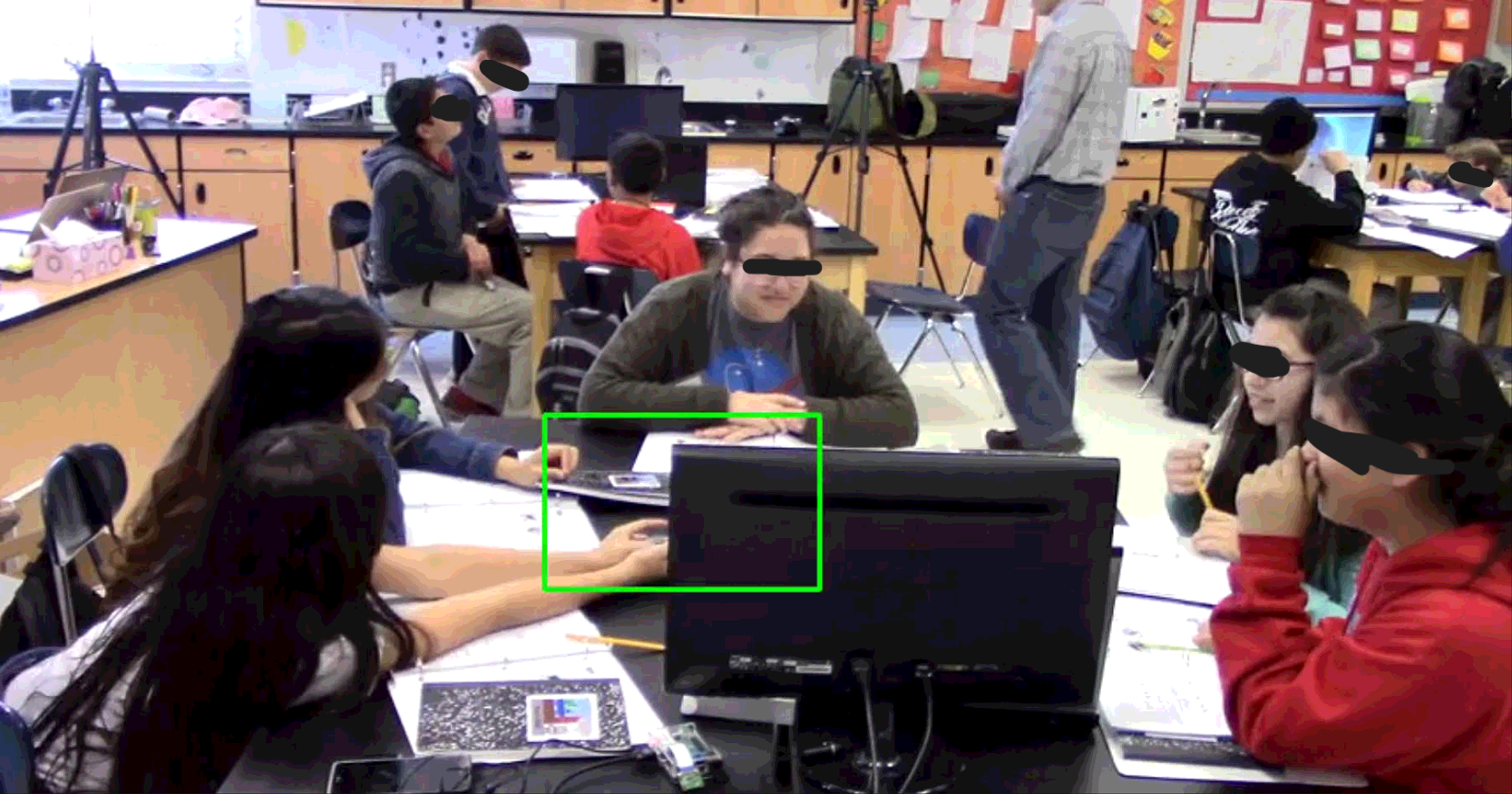}
  }
  \caption{\textbf{Frames demonstrating both successful and unsuccessful cases
      of keyboard tracking for proposing typing regions.}}
  \label{fig:kb-det-cases}
\end{figure}

\begin{figure}
  \centering  

  \begin{subfigure}{0.97\linewidth}
    \includegraphics[width=\linewidth]{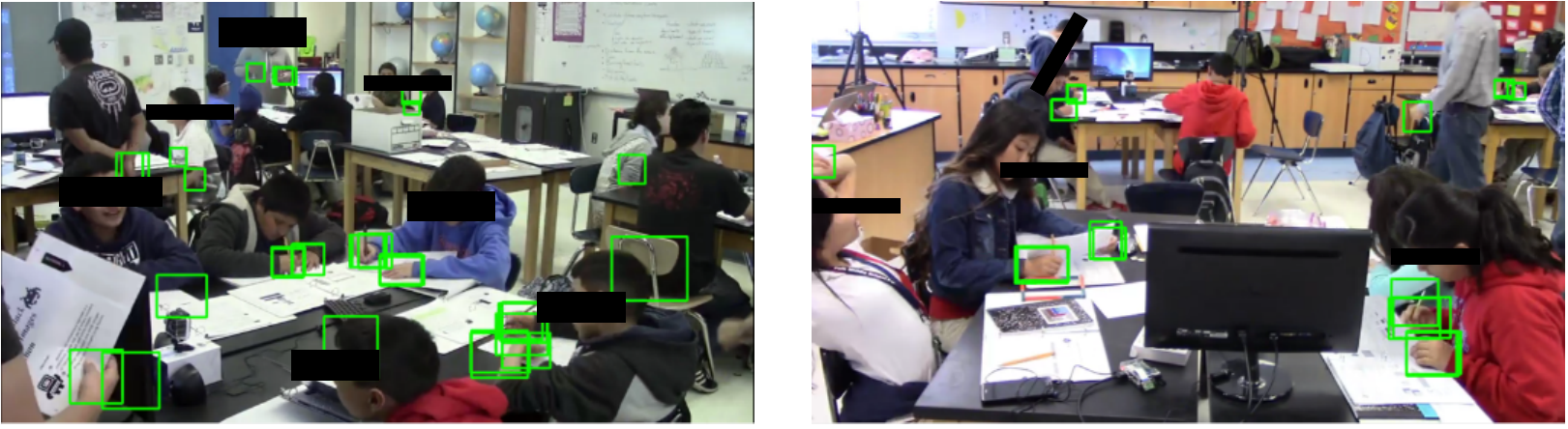}
    \caption{The images demonstrate false positive hand detections in
      the background before projection based
      filtering.}
    \label{subfig:hand-det-only}
  \end{subfigure}
  \vspace{0.5cm}  

  \begin{subfigure}{0.97\linewidth}
    \includegraphics[width=\linewidth]{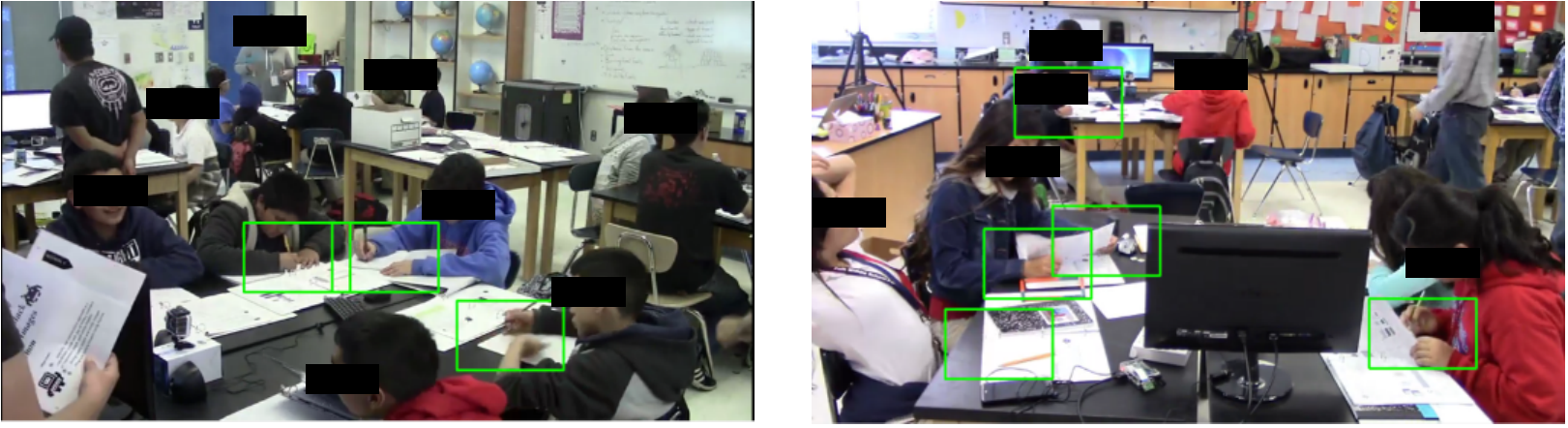}
    \caption{The following demonstrate reduction in background false positive hand detections by using our projection based filtering.}
    \label{fig:hand-proj}
  \end{subfigure}

  \caption{\textbf{The images demonstrate the effectiveness of our projection-based filtering technique for hand detection. As can be seen in the images, the false positives in the background are greatly reduced, resulting in an increase in the accuracy of hand detection in the current group.}}
  \label{fig:hand-dets}
\end{figure}

Both keyboard tracking and hand projections use Faster R-CNN for
detection and employ different post-processing techniques to improve
speed and accuracy, as described in Section
\ref{method-fast-object-tracking}. For keyboard tracking, we used KCF,
a very fast tracking method like KCF to track the keyboard for 5
seconds before reinitializing, which provided a significant boost in
speed with minimal impact on accuracy. On the other hand, hand
projections used 12-second projections to eliminate false positives
and improve performance.

The high average precision (AP) of 0.92 at 0.5 intersection over union
(IOU) achieved by our keyboard detection on the testing set
demonstrates the effectiveness of our system (see Section
\ref{sec:trn-RPN} for details). Following detection, our system
deploys a rapid ($159\times$ real-time) object tracker for five
seconds. Utilizing only keyboard detections, we attain a speed of
$4.7\times$ real-time, and by combining detections with tracking, we
achieve a $22\times$ real-time speed with only a slight decrease in
accuracy. For instance, when testing a session using both detections
and tracking, the accuracy dropped merely from 0.84 IOU to 0.82
IOU. Figure \ref{fig:kb-det-cases} provides examples of keyboard
detections by our system. For more in-depth information, please refer
to the thesis by Sravani Teeparthi \cite{sravani_thesis}.

Our hand detection achieved an average precision of 0.72 at 0.5 IOU on
the dataset described in Section \ref{sec:trn-RPN}. However, the
detection had many false positives, as shown in Figures
\ref{subfig:hand-det-only}. To remove these false positives, we used
the projection technique described in Section
\ref{method-fast-object-tracking}. The projection technique reduced
the detections in the background by at least 75\% across 7 testing
sessions, as shown in Table \ref{tab:pf}. We also demonstrate the
reduction in Figures \ref{fig:hand-dets}.

\subsection{Seperable video activity classification
  results }
\label{sec:results-video-act-calss}

To detect writing and typing from the video activity proposal network,
we employ a separable optimal low-parameter dyadic 3D-CNN model. The
optimal model is selected from our family of models, as described in
Section \ref{sec:method-activity-classifiers}, using the methodology outlined in Section
\ref{sec:optimal-model-selection}. We then compare the performance of
our optimal model against standard activity recognition methods in
Section \ref{sec:trn-act-classifier}.

We employ 3-second representative samples, as detailed in Section
\ref{sec:sampling-procedure}, to evaluate all our experiments. These samples are resized
to have 224 pixels along the longer edge, while the shorter edge is
scaled proportionally to maintain the video's aspect ratio. In
addition to the resolution adjustments, the samples are also
transcoded at 10 and 20 frames per second to accommodate models in our
family that utilize lower frame rate videos. 

\subsubsection{Optimal model selection}
\label{sec:optimal-model-selection}

As described in Section \ref{subsec:model-optimization}, we construct
a family of 12 models that vary in two hyperparameters: the number of
dyads and the input video frame rate. We train each model using the
Adam optimizer with an initial learning rate of 0.001, and use early
stopping and video data augmentation techniques to prevent
overfitting. Specifically, we train each model for a minimum of 50
epochs and a maximum of 100 epochs, with early stopping applied after
50 epochs. The early stopping uses a patience of 5 epochs. This
approach helps to avoid overfitting by stopping the training process
when the model performance no longer improves on the validation set.

We present the results of our optimal model selection experiments in
Table \ref{tab:3d-cnn-opt}. The optimal model is defined based on the
area under the curve (AUC) of the validation set. For typing
classification, models with 4 dyads achieve the highest validation AUC
of 0.95 at both 10 and 30 frames per second (FPS). We choose the 10
FPS model due to its faster inference speed. In the case of writing,
the models with 4 dyads achieve the best validation AUC of 0.84 at 10
FPS.

Both writing and typing classification models attain optimal
performance with 4 dyads. This superior performance is a result of
deeper models more effectively capturing temporal features compared to
their shallow counterparts. As both writing and typing consist of
subtle finger movements, deeper models are better suited for these
tasks. In terms of frame rate, 10 FPS models produce the best outcomes
for both writing and typing. This is due to the limited movement in
consecutive frames. By decreasing the number of frames while keeping
the same duration, the model captures temporal changes in the initial
dyads more accurately. The subsequent dyads then construct more
intricate 3D features based on the initial dyads, ultimately resulting
in enhanced performance.

\begin{table*}[t!]
  \centering
  \caption{\textbf{Low-parameter Dyadic 3D CNN family optimization. This table
      summarizes validation and testing performace at different temporal
      sampling rates and dyads. We mark the optimal model (ours-opt)
      using bold face. The optimal model uses just 18.7 K
      paramters. Also, it is very fast, processing video at just 10
      frames per second.}}
  \label{tab:3d-cnn-opt}
  \begin{tabular}{l l| l l l l l}
    \hline
    \multicolumn{2}{c|}{\bf Hyper parameters} &\textbf{\# Param.}  & \multicolumn{2}{c}{\bf Typing} & \multicolumn{2}{c}{\bf Writing} \\
    
    \textbf{Number of} & \textbf{Frames per} &  & \textbf{Val.} & \textbf{Test} & \textbf{Val.} & \textbf{Test} \\
    \textbf{dyads}     & \textbf{second}     &  & \textbf{AUC}  & \textbf{Acc.} & \textbf{AUC}  & \textbf{Acc.} \\
    \hline
    \hline
    
    1                          & 10                          & 657K                            & 0.5                           & 53.33                               & 0.50                               & 39.93                               \\
    2                          & 10                          & 47K                             & 0.74                          & 61.25                               & 0.50                               & 39.93                               \\
    3                          & 10                          & 7.8K                            & 0.89                          & 61.25                               & 0.58                               & 57.34                               \\
    \colorbox{green}{\bf 4} & \colorbox{green}{\bf 10} & \colorbox{green}{\bf 18.7 K} & \colorbox{green}{\bf 0.95} & \colorbox{green}{\textbf{69.59}} & \colorbox{green}{\textbf{0.84}} & \colorbox{green}{\textbf{63.09}} \\
    1                          & 20                          & 657K                            & 0.5                           & 53.33                               & 0.50                               & 38.90                               \\ 
    2                          & 20                          & 47K                             & 0.5                           & 53.33                               & 0.68                               & 61.09                               \\ 
    3                          & 20                          & 7.8K                            & 0.89                          & 62.08                               & 0.64                               & 59.87                               \\ 
    4                          & 20                          & 18.7K                           & 0.93                          & 65.83                               & 0.69                               & 63.22                               \\
    1                          & 30                          & 657K                            & 0.5                           & 53.33                               & 0.50                               & 39.93                               \\ 
    2                          & 30                          & 47K                             & 0.5                           & 53.33                               & 0.50                               & 39.93                               \\ 
    3                          & 30                          & 7.8K                            & 0.83                          & 62.91                               & 0.52                               & 61.54                               \\         
    4                          & 30                          & 18.7K                           & 0.95                          & 67.91                               & 0.81                               & 64.05                               \\
    \hline
  \end{tabular}
  
\end{table*}

\subsubsection{Performance of our method against 
  State-Of-The-Art (SOTA) methods.}
\label{sec:ours-vs-SOTA}

In the previous section, we determine the optimal classification model
to detect writing and typing to have 4 dyads and use 10 frames per
second activity video samples. In this section, we will compare the
optimal model against SOTA video activity classification systems,
described in Section \ref{subsec:bg-har-systems}.

We evaluate our model in comparison to state-of-the-art (SOTA)
approaches, considering aspects such as classification performance,
model complexity as shown in Table
\ref{tab:results_activity_calssification}.We use Area Under the Curve
(AUC) and accuracy (acc.)  as metrics when evaluating classification
performance.  For model complexity, we use number of trainable
parameters (\# Param.), Graphical Processing Unit memory (GPU mem.)
and inference speed (Inf. speed).  The GPU memory usage and inference
speed are calculated using batch based inference.

The inference speeds and GPU memory usage displayed in the table are
based on the optimal inference batch size. To ensure a fair
comparison, we optimize the inference batch size for both our model
and the SOTA models. The results of these experiments are presented in
Table \ref{tab:optimal_inference_batch_size}. In these results, we
report the inference speed relative to the group interactions video
playback speed, which is standardized to 30 frames per second (FPS),
. When we state that a model can perform inference at $n\times$ speed,
it means that the model can classify $n \times 30$ frames within one
second.

\begin{figure}[t]
  \begin{center}
    \includegraphics[width=\linewidth]{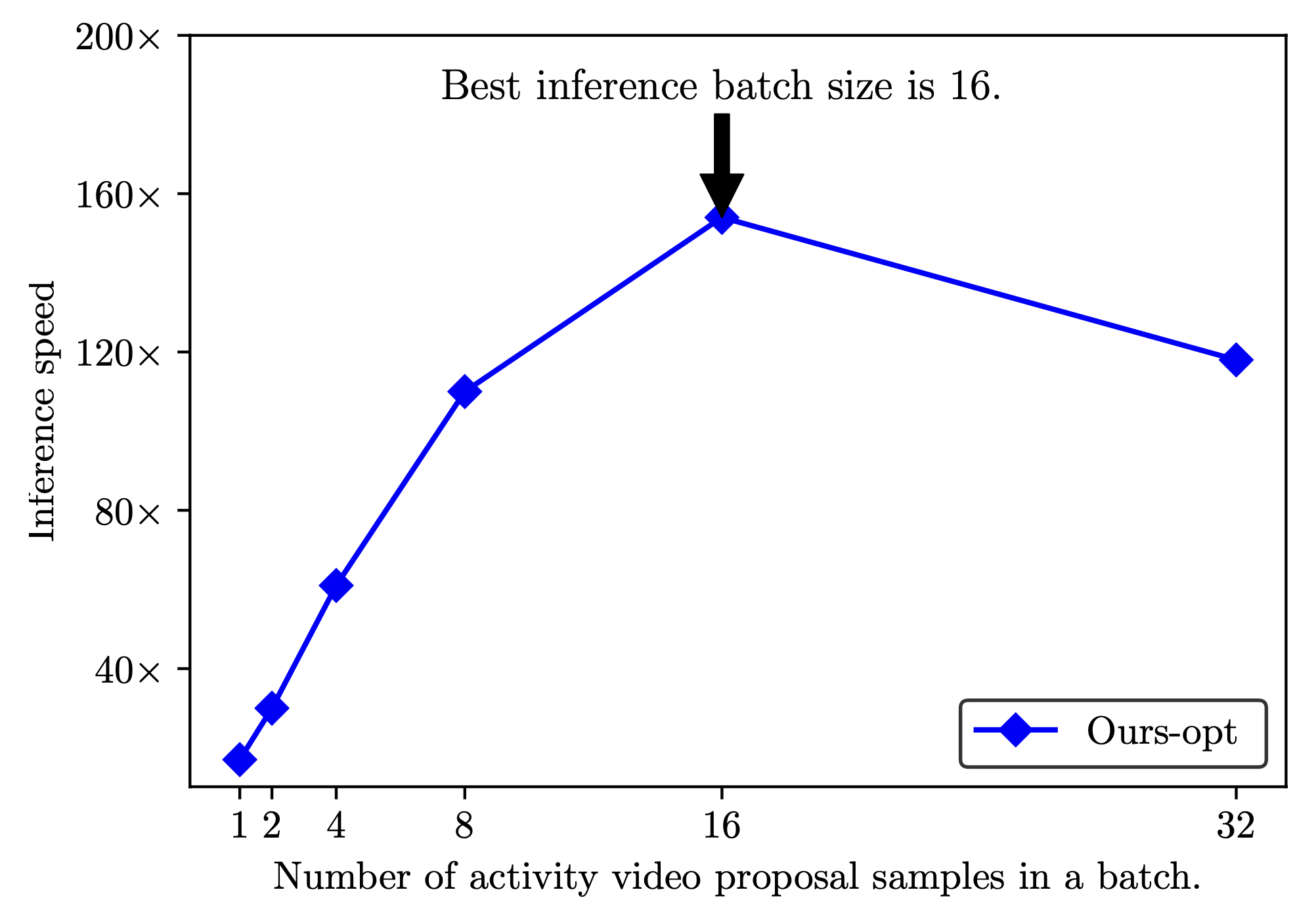}
  \end{center}
  \caption{\textbf{Inference speed for varying activity video sample
      batch sizes of our optimal model (ours-opt). We exponentially
      increase the number of samples as a power of 2. The optimal
      batch size is achieved at 16 samples, as indicated by the
      arrow.}}
  \label{fig:inf-speed-plot}
\end{figure}

From Table \ref{tab:optimal_inference_batch_size}, we observe that
most SOTA models cannot perform inference on more than 4 video
activity samples, except for TSM. This is mainly due to these models
being extremely large and having resource-intensive pre-processing
stages before classification, causing them to run out of GPU memory
when processing more than 4 samples. In contrast, our model does not
require pre-processing, as the spatiotemporal features are captured
within the 3D-CNNs. This enables us to handle more than 4 samples,
with the optimal number being 16 samples. As illustrated in Figure
\ref{fig:inf-speed-plot}, the inference speed decreases after 16
samples due to a bottleneck caused by video decoding.

A video sample must first be decoded and converted into a
floating-point precision 3D numpy array before being fed to the
neural network. In our current model, we have not leveraged the
hardware decoder available in the GPU. As a result, each sample is
decoded in the CPU memory and then copied to the GPU before being fed
into our model, causing a bottleneck. In contrast, SOTA models take
advantage of hardware decoding capabilities using a publicly available
Python library, \texttt{decord} \cite{decord}. We have not explored
this direction for our models; however, we are confident that
utilizing hardware decoding would significantly increase our inference
speed.

We showcase our model's performance against SOTA models using the
optimal batch size in Table
\ref{tab:results_activity_calssification}. Each column of the table
represents either a performance metric or model complexity. The model
with the least complexity and best performance is marked in boldface
and highlighted in green. We compute the validation and testing AUC by
varying the binary classification threshold, and present the
accuracies with a fixed threshold of 0.5.

The proposed approach uses 18.7 K parameters that require 136.32 MB
while running at an 4,620 ($154 \times 30$) frames per second. In terms of
parameters, the proposed approach uses at-least 1,000 less parameters
than any other compared method. In terms of GPU memory, the proposed
method uses 20 times or less memory. In terms of inference, the
proposed method is faster than any other method at 4,620 frames per
second. The proposed method is also more accurate than any other
compared method.

Our model delivers the best performance in typing on
both validation and testing sets. Conversely, it outperforms the SOTA
in the testing set while underperforming only in the validation AUC of
writing. We also observe a significant drop in performance for all
models on the testing set compared to the validation set. This can be
attributed to the nature of the samples in the testing set.

The training and validation samples primarily consist of sessions from
cohort 1 (2017), while the testing sessions are taken from cohort 2
(2018) and cohort 3 (2019). Additionally, the testing sessions have
complete ground truth, meaning they contain more samples without
typing or writing, resulting in an imbalanced dataset. We
intentionally designed the dataset this way to study the performance
of the activity detection system when given a completely new session.

\begin{table}[t]
  \setlength{\tabcolsep}{4pt}
  \centering
  \caption{\textbf{Batch-size optimization for model inference. The following
      table presents inference speed at different batch sizes.  We
      report the inference speed in terms of sample playback time.  We
      highlight the the optimal batch size per method in green and a
      cross (\xmark) to mark that the method failed to perform inference
      due to insufficient GPU memory (our GPU, RTX 5000), has 16 GB of GPU
      memory).}}
  \label{tab:optimal_inference_batch_size}

  \begin{tabular}{l | l l l l l l}
    \hline
    \textbf{Method} & \multicolumn{6}{c}{\bf Inference speed}                                                                                                                                           \\
                    & \multicolumn{6}{c}{\bf (at different batch sizes)}                                                                                                                                \\
                    & \textbf{1}                         & \textbf{2}                         & \textbf{4}  & \textbf{8}                           & \textbf{16}                          & \textbf{32} \\
    \hline
    I3D             & $2\times$                          & \colorbox{green}{\bf 3$\times$} & \xmark      & \xmark                               & \xmark                               & \xmark      \\
    SlowFast        & $2\times$                          & \colorbox{green}{\bf 3$\times$} & \xmark      & \xmark                               & \xmark                               & \xmark      \\    
    TSM             & $37\times$                         & $59\times$                         & $110\times$ & \colorbox{green}{\bf 118$\times$} & $102\times$                          & $90\times$  \\
    TSN             & \colorbox{green}{\bf 4$\times$} & $4\times$                          & $4\times$   & \xmark                               & \xmark                               & \xmark      \\
    ours-opt       & $17\times$                         & $30\times$                         & $61\times$  & $110\times$                          & \colorbox{green}{\bf 154$\times$} & $118\times$ \\
    \hline
    \hline
  \end{tabular}
\end{table}

\begin{table*}[t]
  \centering
  \setlength{\tabcolsep}{12pt}
  \caption{\textbf{Parameter, inference speed, and memory requirements for
    proposed methods and comparative methods. The proposed approach
    uses over a 1000 times less parameters, requires far less memory,
    runs faster than everything, and performs better than all other
    methods. The model performance is presented using Area Under the
    Curve (AUC) and accuracy. We present the model complexity using
    inference speed, number of parameters (\# Param.), and video
    memory (GPU Mem. in MB) used by the model. We use bold face and
    highlight in green to mark the least complexity and best
    performance per column.}}
  \label{tab:results_activity_calssification}
  \begin{tabular}{l | l l l l l l l}
    \hline
    \textbf{Method} & \textbf{\# Param.}                          & \textbf{Inf.}                        & \textbf{GPU Mem.}                            & \textbf{Val.}                 & \textbf{Test}                 & \textbf{Val.}                  & \textbf{Test}                  \\
                    &                                             & \textbf{speed}                       & \textbf{in MB}                           & \textbf{AUC}                  & \textbf{AUC}                  & \textbf{acc.}                  & \textbf{acc.}                  \\
    \hline
    \hline
                    & \multicolumn{7}{c}{\bf Typing and no-typing classification results}                                                                                                                                                                                            \\
    \cline{2-8}
    I3D             & 27.2M ($1437\times$)                       & $3\times$                            & 5051 ($20\times$)                       & 0.73                          & 0.66                          & 77.40                          & 64.58                          \\
    Slowfast        & 33.5M ($1787\times$)                       & $3\times$                            & 6318 ($25\times$)                       & 0.94                          & 0.71                          & \colorbox{green}{\bf 89.79} & 61.25                          \\
    TSM             & 23.5M ($1252\times$)                       & $118\times$                          & 6971 ($28\times$)                       & 0.84                          & 0.59                          & 87.75                          & 58.75                          \\
    TSN             & 23.5M ($1252\times$)                       & $4\times$                            & 5593 ($23\times$)                       & 0.86                          & 0.74                          & 85.71                          & 65                             \\
    Ours-opt       & \colorbox{green}{\bf 18.7K ($1\times$)} & \colorbox{green}{\bf 154$\times$} & \colorbox{green}{\bf 245($1\times$)} & \colorbox{green}{\bf 0.96} & \colorbox{green}{\bf 0.76} & \colorbox{green}{\bf 89.79} & \colorbox{green}{\bf 69.58} \\
    \hline
                    & \multicolumn{7}{c}{\bf Writing and no-writing classification results}                                                                                                                                                                                          \\
    \cline{2-8}
    I3D             & 27.2M ($1437\times$)                       & $3\times$                            & 5051 ($20\times$)                       & 0.73                          & 0.66                          & 77.40                          & 59.58                          \\
    Slowfast        & 33.5M ($1787\times$)                       & $3\times$                            & 6318 ($25\times$)                       & 0.75                          & 0.57                          & 75.96                          & 53.67                          \\
    TSM             & 23.5M ($1252\times$)                       & $118\times$                          & 6971 ($28\times$)                       & 0.73                          & 0.50                          & 72.11                          & 47.60                          \\
    TSN             & 23.5M ($1252\times$)                       & $4\times$                            & 5593 ($23\times$)                       & \colorbox{green}{\bf 0.92} & 0.62                          & \colorbox{green}{\bf 77.88} & 61.66                          \\
    Ours-opt       & \colorbox{green}{\bf 18.7K ($1\times$)} & \colorbox{green}{\bf 154$\times$} & \colorbox{green}{\bf 245($1\times$)} & 0.85                          & \colorbox{green}{\bf 0.67} & \colorbox{green}{\bf 77.88} & \colorbox{green}{\bf 63.09} \\    
    
    \hline
  \end{tabular}
  
\end{table*}

\section{Long-term activity detection performance
  and interactive activity map visualization. \label{sec:results-int-viz}}

In previous sections, we presented the results of the video activity
proposal network and low-parameter, separable video activity
classifications separately. In this section, we will discuss the
activity detections achieved by combining these two approaches using
complete session (1 to 1.5 hours of video playback time). Furthermore,
we will present a novel interactive visualization of these activity
detections using a web based application.

\subsection{Typing detection and visualization}
\label{sec:long-term-ty-det}

\begin{figure}[!t]
  
  \includegraphics[width=0.97\linewidth]{ 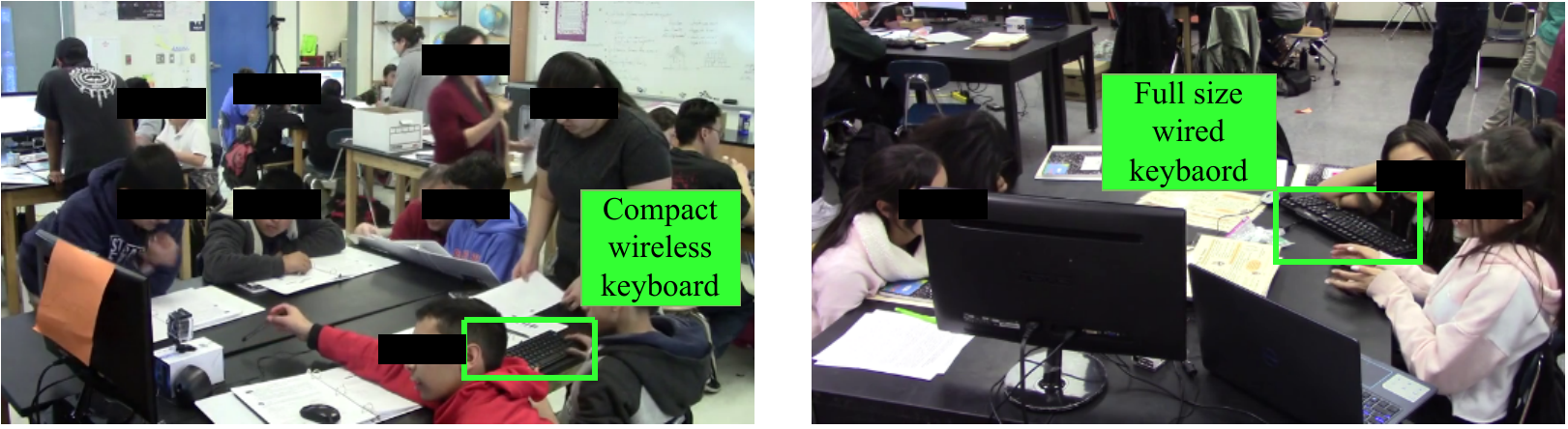 }

  \caption{Sessions used to test our typing detection system. On the
    left we have the first session (TS1) from group E of cohort 1
    (2017) and level 1. On the right we have the second session (TS2)
    from group D of cohort 3 (2019) and level 1.}
\label{fig:end-to-end-ty-sessions}
\end{figure}

We describe the end-to-end performance of our typing detection
system using two sessions. These sessions are chosen from different
cohorts (years apart) and exhibit variations in camera position,
lighting, and keyboard types, as illustrated in Figure
\ref{fig:end-to-end-ty-sessions}. The figure demonstrates that the groups
cover the cases of two different keyboards used in AOLME sessions: a
compact wireless keyboard (on the left) and a full size wired keyboard
(on the right). For brevity, we will refer to these sessions using the
acronyms TS1 (Typing Session 1) for the session from cohort 1 and TS2
(Typing Session 2) for the session from cohort 2.

The duration of TS1 is approximately 1 hour and 25 minutes, while TS2
lasts around 1 hour and 48 minutes. In Table
\ref{tab:end-to-end-ty-time}, we summarize the time taken at different
stages of our system for each session. From the table, it is evident
that the majority of the time is spent in the video activity proposal
network. The classification process is extremely fast, taking only
around 26 seconds for a session of 90 minutes, owing to our use of
optimal batch size and reduced input video frame rate. Upon closer
inspection of our proposal network, we found that the primary speed
bottleneck is caused during the small video extraction phase. The
small video extraction process currently utilizes single-thread
execution, and we are confident that by employing multi-thread
execution, it can become significantly faster.

\begin{figure}
  \includegraphics[width=0.97\linewidth]{ 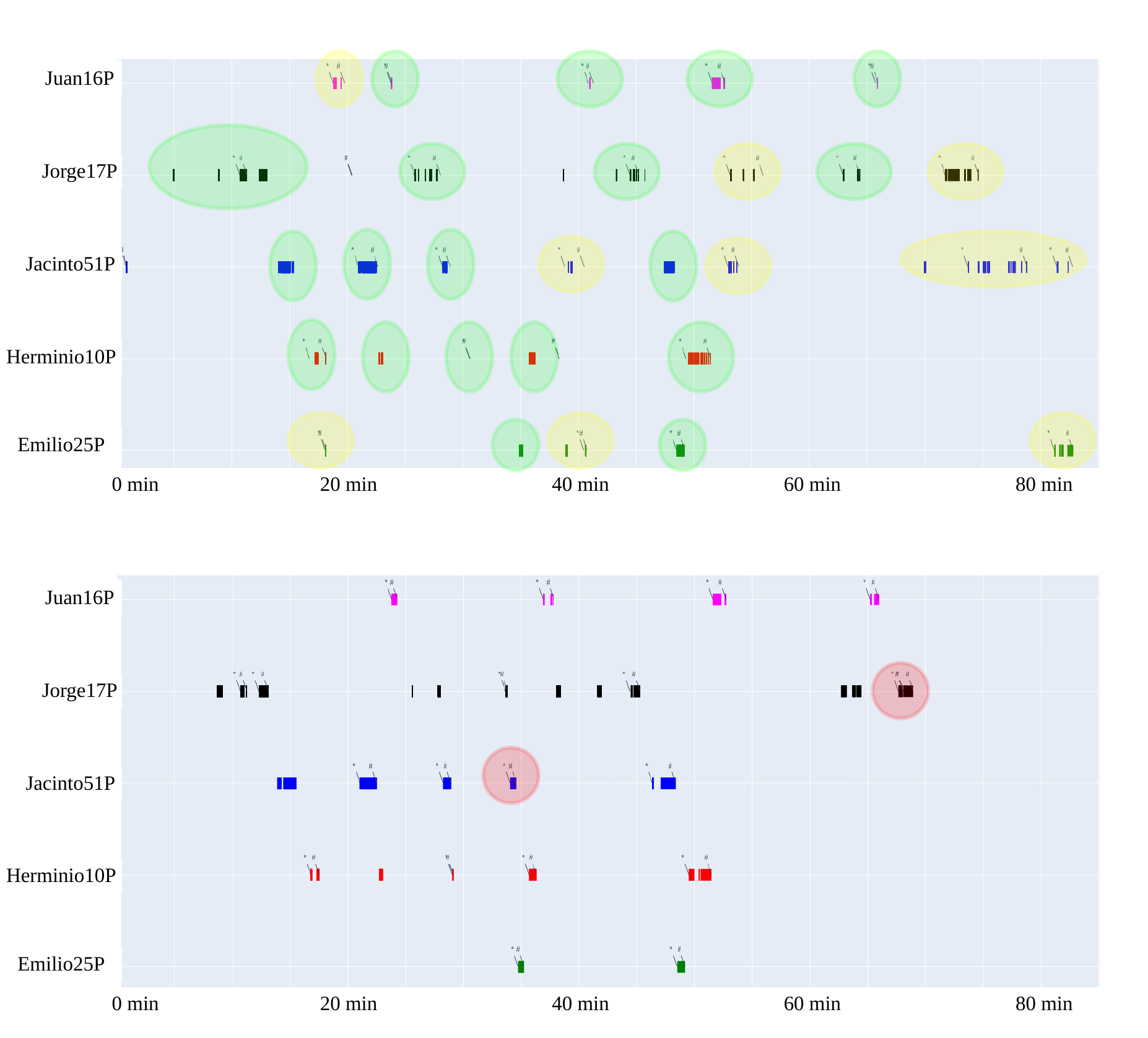}
  \caption{Typing activity map of TS1. Our system's typing detections
    are shown at the top, and the ground truth labels are displayed at
    the bottom. We use green, yellow, and red circles to highlight
    typing detection clusters that represent true positives, false
    positives, and false negatives, respectively.}
  \label{fig:int-ty-act-map}
\end{figure}

We present a visual comparison of typing detections against ground
truth using typing activity maps in Figure
\ref{fig:int-ty-act-map}. In this figure, we display our detections
on the top and the ground truth activity labels on the bottom for
TS1. The maps show not only the occurrence of typing but also the
pseudonym of the person performing it. We group closely occurring
typing detections (less than 3 seconds apart) to form clusters. Typing
clusters with corresponding ground truth labels are highlighted in
green circles, while typing detections by our system without
corresponding ground truth labels are highlighted in yellow. Typing
activities that we failed to detect in the session are highlighted in
red. We trained our model to be sensitive to typing, aiming to have
more false detections and minimize false negatives, and as shown in
the figure, we have achieved this objective effectively.

\begin{table*}[!t]
  \centering
  \caption{Time taken to process typing detection
    system\label{tab:end-to-end-ty-time}. For a one-hour video, our
    system takes only 15 minutes to perform typing detection.}
  \begin{tabular}{p{7.5cm}|ll}
    \hline
    \textbf{Typing detection stage} & \multicolumn{2}{c}{\bf Duraton in HH:MM:SS} \\
                                    & \textbf{TS1 (01:25:06)} & \textbf{TS2 (01:48:00)}     \\
    \hline
    \hline

    \textbf{Typing activity proposal network}              & 00:20:58              & 00:21:55                \\
    \textbf{Low-paramter typing activity classification}   & 00:00:26              & 00:00:26                \\
    \textbf{Interactive visuaization of typing activities} & 00:00:19              & 00:00:20                \\
    \hline
    \textbf{Total inference time}                          & 00:21:33 (4 $\times$) & 00:22:41 (4.7 $\times$) \\
    \hline
  \end{tabular}
\end{table*}

\newpage
\subsection{Writing detection and visualization}
\label{sec:we-det-vis}

\begin{figure}[!t]
  
  \includegraphics[width=0.97\linewidth]{ 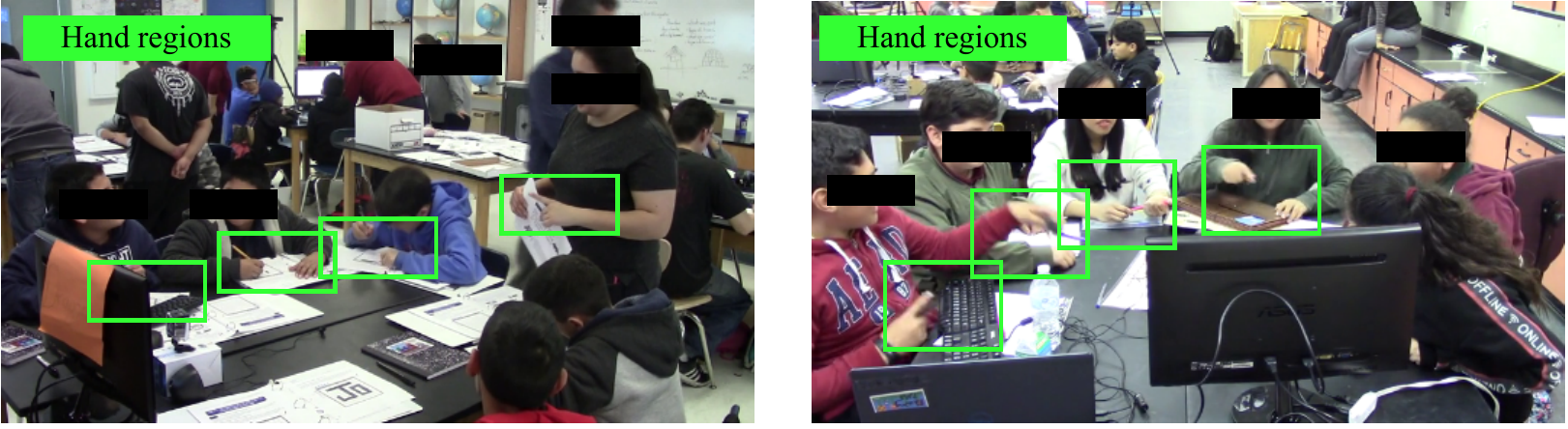 }

  \caption{Sessions used to test our writing detection system. On the
    left we have the first session (WS1) from group E of cohort 1
    (2017) and level 1. On the right we have the second session (TS2)
    from group C of cohort 3 (2019) and level 1. \label{fig:end-to-end-wr-sessions}}
\end{figure}

Similar to typing, we showcase the end-to-end performance of our
writing detection system using two testing sessions, writing session 1
(WS1) and writing session 2 (WS2). We selected these sesions, refer to
Figure \ref{fig:end-to-end-wr-sessions}, to be from different cohorts
and have considerable variability in camera angle and lighting
conditions.

In our typing detection system, we use keyboard tracking and activity
initializations for proposing activity regions. Similarly, in writing
detection, we use hand projection regions and activity
initializations. However, this approach did not effectively improve
our system's performance or speed. The performance was negatively
affected due to the presence of many valid hand regions, as
demonstrated by the green box in Figure
\ref{fig:end-to-end-wr-sessions}. The speed suffered because of the
need to detect hands every second. We provide the time taken for
writing detection in Table \ref{tab:end-to-end-wr-time}. In comparison
to typing detection, we can see that writing detection takes a
significant amount of time in hand region detection using projections.

The failure of writing detection can be attributed to three main
reasons. First, the classifiers are trained on very clean ``no writing''
instances, where the ``no writing'' instances seldom have hands and
pens. In contrast, the majority of ``no writing'' instances observed in
complete sessions have hand movements on a paper, and sometimes they
also have a pen in hand while not writing with the pen. Second, the
writing activity regions extracted from hand projections and activity
initialization typically overlap with the next student. In these
cases, the writing classifier produces a false positive. Third, unlike
typing, where the keyboard does not have hands when there is ``no
typing'', the ``no writing'' instances almost always have hands on the
paper, registering movements similar to writing. Our writing detector
is highly sensitive to hand movement on the paper and produces a lot
of false positives. These issues can be addressed by employing
different approaches, such as detecting a hand holding a pen, using
hand shape-based classifiers, and other techniques.

\begin{table*}[!t]
  \centering
  \caption{Time taken to process writing detection system.  We process
    1 hour writing in approximately 50
    minutes.\label{tab:end-to-end-wr-time}}
  \begin{tabular}{p{7.4cm}|ll}
    \hline
    \textbf{Typing detection stage} & \multicolumn{2}{c}{\bf Duraton in HH:MM:SS}       \\
                                    & \textbf{WS1 (01:25:06)} & \textbf{WS2 (01:47:48)} \\
    \hline
    \hline

    \textbf{Writing activity proposal network}             & 01:34:03                 & 01:39:42                 \\
    \textbf{Low-paramter typing activity classification}   & 00:05:28                 & 00:05:35                 \\
    \textbf{Interactive visulalization of typing activities} & 00:00:19                 & 00:00:20                 \\
    \hline
    \textbf{Total inference time}                          & 01:39:50 (0.86 $\times$) & 01:45:00 (0.93 $\times$) \\
    \hline
  \end{tabular}
\end{table*}

\section{Conclusion and Future Work}
\label{conclusion_and_future_work}

\subsection{Conclusion}

An advanced video activity detection system has been
developed, specifically focusing on the detection of typing and
writing actions in AOLME group interaction videos.  The primary
contributions of this paper include: (i) the creation of a very
fast, separable, low-parameter, and memory-efficient model, employing
3D-CNNs for the purpose of activity classification, (ii) the
implementation of an accurate and fast inference mechanism that
utilizes optimal depth, input video frame rate, and batch size, (iii)
the adoption of a modular and streamlined training methodology that
leverages a limited dataset, (iv) the establishment of interactive
activity maps utilizing web-based technologies for the visualization
of detected activities, and (v) the integration of well-established
deep learning-based object detection methodologies in conjunction with
tracking and projection-based techniques to detect video activity
regions. The classifiers outperform comparable approaches by using over
1,000 times fewer parameters. They achieve a testing AUC of 0.76 and
0.67 for typing and writing activities, respectively.

The typing and writing detection systems offer faster inference speed
compared to end-to-end activity detection systems. The typing
detection system, which uses keyboard tracking and activity
initializations, shows promising results in terms of speed and
effectiveness. However, the writing detection system, based on hand
projections and activity initializations, does not meet the desired
levels of accuracy and efficiency. The classifiers have difficulty
distinguishing between writing and the absence of
writing. Improvements can be made by utilizing more carefully curated
training data or exploring alternative approaches such as hand shape
classification and pen detection.

\subsection{Future work}
\label{sec:future-work}

The developed system employs a modular design and utilizes a limited dataset for
training, as outlined in Section \ref{sec:system}. These design
principles can be harnessed to enhance typing detection and address
the current shortcomings of the writing detection system. In this
section, we summarize these concepts for improving writing detection
and typing detection.

\subsubsection{Improving writing detection}
\label{sec:improving-writing-detection}
Detecting writing activities in group interaction videos presents a
highly complex challenge. The system's performance in identifying this
activity was suboptimal, as discussed in Section \ref{sec:we-det-vis}.
Upon closer examination, there are some open issues that need to be
addressed in future work. In this section, a number of approaches will
be proposed that may overcome these issues.

\textbf{Challenging and diverse training samples  \label{sec:challengin-no-writing insances}}\\

\begin{figure*}[!t]
  \centering  

  \begin{subfigure}{0.97\textwidth}
    \includegraphics[width=\linewidth]{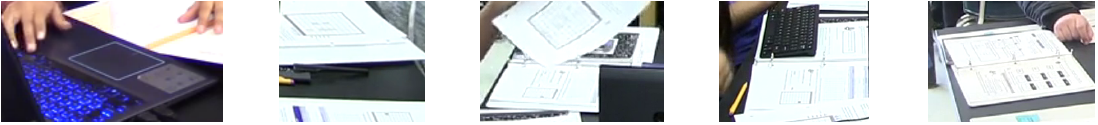}
    \caption{Example images of samples illustrating the absence of writing. It is evident from the figure that the samples consistently display a paper or book on the table. As there are no hands present in these samples, the classifier is likely learning that the presence of a hand with minimal movement signifies writing.}
    \label{fig:easy-no-writing-samples}
  \end{subfigure}

  \begin{subfigure}{0.97\textwidth}
    \includegraphics[width=\linewidth]{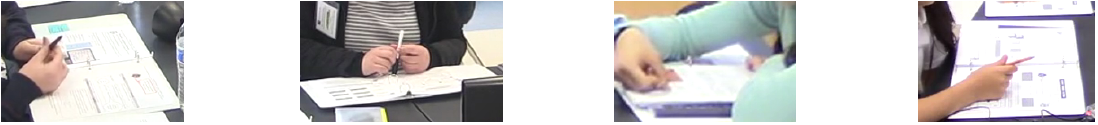}
    \caption{Challenging examples of samples illustrating the absence of writing. As can be observed in the figure, these samples feature hands executing movements similar to those associated with writing.}
    \label{fig:hard-no-writing-samples}
  \end{subfigure}

  \caption{\textbf{Easy and hard cases of samples depicting absence of
    writing. We need more hard samples to train a robust writing
    activity classifier.}}
  \label{diverse-nowriting-samples}
\end{figure*}

The classifier tasked with discerning the presence and absence of
writing activities employs a diverse and robust collection of samples
representing the presence of writing. In contrast, the majority of
samples illustrating the absence of the activity are relatively simple
and unvaried, as shown in Figure
\ref{fig:easy-no-writing-samples}. These samples often display paper
without a hand or a hand without movement. As a result, it is possible
that the classifier is learning to associate the presence of a hand
with small movement as an indication of writing.

To address this issue, we must enhance our manual labeling protocol to
include cases where hands are present on the paper, as illustrated in
Figure \ref{fig:hard-no-writing-samples}. Specifically, we should pay
special attention to incorporating samples that feature hands on the
paper exhibiting movements similar to writing.

\textbf{Filtering writing proposal regions with pen detection\label{sec:pen-detection}}\\

The system employs hand projections and activity region
initializations to propose activity regions, as depicted in Figure
\ref{fig:activity-detection-system-overview}. However, the hand
projections provided limited filtering capabilities for the activity
region proposals, as hands are consistently present. To enhance the
system, we could consider incorporating a pen or pencil detector, as
explored in related research by Jacoby et
al. \cite{jacoby2018context}. The presence of a hand without a pen or
pencil indicates that the proposed region cannot be associated with
writing. Due to the modular design of our system, integrating a pen
detector would be a straightforward process.

\subsubsection{Post-processing techniques to improve typing
  detection }
  \label{sec:improve-typing-det}

We achieved high accuracy for typing classification, and the typing
detection system, as discussed in Section \ref{sec:ours-vs-SOTA} and
Section \ref{sec:long-term-ty-det}, produced satisfactory
results. These outcomes can be further improved using context-based
post-processing techniques. A simple and straightforward approach
would involve filtering out typing instances with durations shorter
than a predetermined threshold. This would help eliminate false
positives and refine the overall performance of the typing detection
system.

Another strategy could be based on the fact that a group typically has
only one keyboard, and since only one student can type at a time, if
typing activities are detected occurring simultaneously across
multiple students, we should consider the activity with the highest
classification probability as the valid one. By incorporating such
context-aware techniques, the system can better differentiate between
true typing instances and potential false detections, leading to more
accurate and reliable results for typing activity identification in
group interaction videos.

  \subsubsection{Faster inferencing using parallel threading and hardware
    video decoder }
    \label{sec:futre-hardware-decoder}

  The already fast inferencing speed of our method, as presented in
  Section \ref{results} of this paper, can be further
  improved by employing multi-threading and hardware video
  decoding. Due to our model's small memory footprint, we can
  effectively utilize parallel threads to load multiple models in the
  GPU and infer on non-overlapping batches, significantly increasing
  our system's classification inferencing speed.

  Furthermore, our method currently does not take advantage of the
  hardware video decoding provided by Nvidia GPUs. In contrast, the
  SOTA methods presented in this paper utilize hardware
  decoding through the \texttt{decord} \cite{decord} Python
  library. Video decoding is a time-consuming process, and the video
  is initially decoded to CPU memory before being loaded into GPU
  memory. By leveraging hardware decoding, we can improve inference
  speed by avoiding the need to copy video data from CPU memory to GPU
  memory.

  \subsection{Implications to education researchers}
  The activity map developed as part of our video activity detection
  system offers valuable insights for education researchers studying
  group interactions and coding activities. By analyzing the activity
  map, researchers can gain a better understanding of student
  engagement and the dynamics of the learning process. Some of the
  questions that can be addressed using typing detection and activiyt
  map are, \textbf{Q1:} When did the students start/stop coding
  (typing)?, \textbf{Q2:} Which student used the keyboard the most?,
  \textbf{Q3:} Did the facilitator interfere with the students while
  coding? Did they do most of the work?, \textbf{Q4:} Which challenge
  engaged the most students?

  Questions Q1, Q2, and Q3 can be answered quantitatively and automated,
  providing researchers with precise information about the timing,
  frequency, and distribution of coding activities among students. This
  data can help identify patterns of student engagement, collaboration,
  and potential areas where facilitator intervention may be required.

  On the other hand, answering question Q4 requires researchers to
  interact with and infer from the activity map. By examining the map,
  researchers can determine which challenges garnered the most
  involvement from students, indicating the effectiveness of the tasks
  in promoting collaboration and active learning. Understanding these
  dynamics allows education researchers to develop more effective
  learning strategies, tailor educational content, and optimize group
  interaction for better learning outcomes.

  \begin{figure*}
    \includegraphics[width=0.97\textwidth]{ 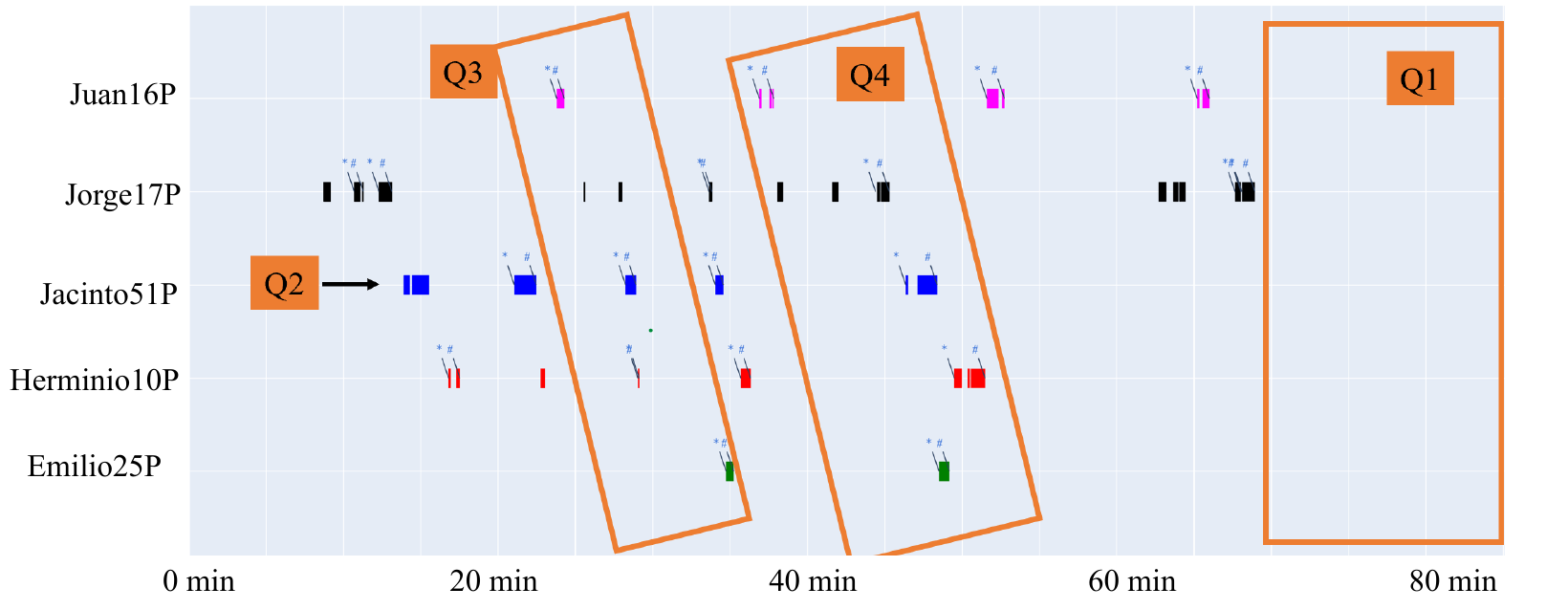 }
    \caption{\textbf{Implications to education researchers. We detect typing
      in this session. From the figure we can clerly see that
      Jacinto51P did most of the typing and most students got involved
      in the sesson around 25 minutes to 55 minutes.}}
      \label{fig:edu-implications}
  \end{figure*}

  \appendices
  
  \section{Video compression}

  \begin{figure}
\begin{lstlisting}[language=bash]
ffmpeg -i <input video> \
  -vf scale=858:480 \
  -c:v libx264 \
  -c:a mp3 -b:a 255k \
  -b:v 2.5M \
  -maxrate 2.5M \
  -bufsize 1.25M \
  -r 30 \
  -x264-params \
  "keyint=30:min-keyint=30:no-scenecut" \
  <output video>
\end{lstlisting}
    \caption{ We utilized \texttt{ffmpeg} to transcode the original
      high-quality videos to lower resolutions and frame rates. This
      transcoding process ensures that the videos maintain quality
      level agreed by education researchers, making them well-suited
      for efficient streaming without compromising on their audio
      integrity.}
      \label{fig:ffmpeg-commands}
\end{figure}

\label{sec:video_compression}
The transcoding command, shown in Fig. \ref{fig:ffmpeg-commands},
is designed to transcode an input video into a specific
output. The command rescales the input video to a resolution of
858x480 pixels, specified by the \texttt{-vf scale=858:480} option.
We use H.264 video codec, which is indicated by \texttt{-c:v
  libx264}. The video bitrate is established at 2.5Mbps with the
\texttt{-b:v 2.5M} option, and the audio bitrate is set to 255Kbps
using \texttt{-b:a 255k}.

The \texttt{-bufsize 1.25M} parameter controls the encoder's buffer
size, crucial for handling variations in the video's bitrate during
playback to ensure smoother video delivery. Setting the frame rate to
30 frames per second with \texttt{-r 30} ensures fluid motion, which
is ideal for most video contents. Further fine-tuning of the encoding
process is achieved through the \texttt{-x264-params
  "keyint=30:min-keyint=30:no-scenecut"} options: \texttt{keyint=30}
sets the maximum interval between keyframes to 30 frames,
\texttt{min-keyint=30} establishes the minimum interval at the same
value to guarantee a keyframe at least once every second, and
\texttt{no-scenecut} disables scene cut detection. This latter option
helps in maintaining a consistent visual quality across the video by
avoiding abrupt changes in bitrate or quality due to scene changes.

\bibliographystyle{IEEEtran}
\bibliography{dissertation}

\begin{thebibliography}{10}
\providecommand{\url}[1]{#1}
\csname url@samestyle\endcsname
\providecommand{\newblock}{\relax}
\providecommand{\bibinfo}[2]{#2}
\providecommand{\BIBentrySTDinterwordspacing}{\spaceskip=0pt\relax}
\providecommand{\BIBentryALTinterwordstretchfactor}{4}
\providecommand{\BIBentryALTinterwordspacing}{\spaceskip=\fontdimen2\font plus
\BIBentryALTinterwordstretchfactor\fontdimen3\font minus
  \fontdimen4\font\relax}
\providecommand{\BIBforeignlanguage}[2]{{%
\expandafter\ifx\csname l@#1\endcsname\relax
\typeout{** WARNING: IEEEtran.bst: No hyphenation pattern has been}%
\typeout{** loaded for the language `#1'. Using the pattern for}%
\typeout{** the default language instead.}%
\else
\language=\csname l@#1\endcsname
\fi
#2}}
\providecommand{\BIBdecl}{\relax}
\BIBdecl

\bibitem{yao2019review}
G.~Yao, T.~Lei, and J.~Zhong, ``A review of convolutional-neural-network-based
  action recognition,'' \emph{Pattern Recognition Letters}, vol. 118, pp.
  14--22, 2019.

\bibitem{yt-8m}
S.~Abu-El-Haija, N.~Kothari, J.~Lee, P.~Natsev, G.~Toderici, B.~Varadarajan,
  and S.~Vijayanarasimhan, ``Youtube-8m: A large-scale video classification
  benchmark,'' \emph{arXiv preprint arXiv:1609.08675}, 2016.

\bibitem{vahdani2022deep}
E.~Vahdani and Y.~Tian, ``Deep learning-based action detection in untrimmed
  videos: a survey,'' \emph{IEEE Transactions on Pattern Analysis and Machine
  Intelligence}, 2022.

\bibitem{ucf101}
K.~Soomro, A.~R. Zamir, and M.~Shah, ``Ucf101: A dataset of 101 human actions
  classes from videos in the wild,'' \emph{arXiv preprint arXiv:1212.0402},
  2012.

\bibitem{tsn}
L.~Wang, Y.~Xiong, Z.~Wang, Y.~Qiao, D.~Lin, X.~Tang, and L.~Van~Gool,
  ``Temporal segment networks for action recognition in videos,'' \emph{IEEE
  Transactions on Pattern Analysis and Machine Intelligence}, vol.~41, no.~11,
  pp. 2740--2755, 2019.

\bibitem{i3d}
J.~Carreira and A.~Zisserman, ``Quo vadis, action recognition? a new model and
  the kinetics dataset,'' 07 2017, pp. 4724--4733.

\bibitem{lin2019tsm}
J.~Lin, C.~Gan, and S.~Han, ``Tsm: Temporal shift module for efficient video
  understanding,'' in \emph{Proceedings of the IEEE International Conference on
  Computer Vision}, 2019.

\bibitem{feichtenhofer2019slowfast}
C.~Feichtenhofer, H.~Fan, J.~Malik, and K.~He, ``Slowfast networks for video
  recognition,'' in \emph{Proceedings of the IEEE international conference on
  computer vision}, 2019, pp. 6202--6211.

\bibitem{ActivePresenter}
\BIBentryALTinterwordspacing
{ Atomi Systems}, ``Activepresenter.'' [Online]. Available:
  \url{https://atomisystems.com/activepresenter/}
\BIBentrySTDinterwordspacing

\bibitem{raspberry_pi}
\BIBentryALTinterwordspacing
{Raspberry Pi Foundation}, ``Raspberry pi.'' [Online]. Available:
  \url{https://www.raspberrypi.org/}
\BIBentrySTDinterwordspacing

\bibitem{hmdb}
H.~Kuehne, H.~Jhuang, E.~Garrote, T.~Poggio, and T.~Serre, ``Hmdb: a large
  video database for human motion recognition,'' in \emph{2011 International
  conference on computer vision}.\hskip 1em plus 0.5em minus 0.4em\relax IEEE,
  2011, pp. 2556--2563.

\bibitem{kinetics}
W.~Kay, J.~Carreira, K.~Simonyan, B.~Zhang, C.~Hillier, S.~Vijayanarasimhan,
  F.~Viola, T.~Green, T.~Back, P.~Natsev \emph{et~al.}, ``The kinetics human
  action video dataset,'' \emph{arXiv preprint arXiv:1705.06950}, 2017.

\bibitem{activitynet}
F.~Caba~Heilbron, V.~Escorcia, B.~Ghanem, and J.~Carlos~Niebles, ``Activitynet:
  A large-scale video benchmark for human activity understanding,'' in
  \emph{Proceedings of the ieee conference on computer vision and pattern
  recognition}, 2015, pp. 961--970.

\bibitem{Tran2021}
P.~Tran, M.~S. Pattichis, S.~Celedón-Pattichis, and C.~L. Leiva, ``Facial
  recognition in collaborative learning videos,'' in \emph{19th International
  Conference CAIP}.\hskip 1em plus 0.5em minus 0.4em\relax Springer, 2021.

\bibitem{KCF}
J.~F. Henriques, R.~Caseiro, P.~Martins, and J.~Batista, ``Exploiting the
  circulant structure of tracking-by-detection with kernels,'' in
  \emph{Computer Vision--ECCV 2012: 12th European Conference on Computer
  Vision, Florence, Italy, October 7-13, 2012, Proceedings, Part IV 12}.\hskip
  1em plus 0.5em minus 0.4em\relax Springer, 2012, pp. 702--715.

\bibitem{sanchez2021bilingual}
L.~Sanchez~Tapia, A.~Gomez, M.~Esparza, V.~Jatla, M.~Pattichis,
  S.~Celed{\'o}n-Pattichis, and C.~L{\'o}pezLeiva, ``Bilingual speech
  recognition by estimating speaker geometry from video data,'' in
  \emph{Computer Analysis of Images and Patterns: 19th International
  Conference, CAIP 2021, Virtual Event, September 28--30, 2021, Proceedings,
  Part I}.\hskip 1em plus 0.5em minus 0.4em\relax Springer, 2021, pp. 79--89.

\bibitem{Antonio2022}
A.~Gomez, M.~S. Pattichis, and S.~Celedón-Pattichis, ``Speaker diarization and
  identification from single channel classroom audio recordings using virtual
  microphones,'' \emph{IEEE Access}, vol.~10, pp. 56\,256--56\,266, 2022.

\bibitem{lopezleiva2020participation}
C.~L{\'o}pezLeiva, S.~Celed{\'o}n-Pattichis, and M.~S. Pattichis,
  ``Participation in the advancing out-of-school learning in mathematics and
  engineering project,'' \emph{Girls and Women of Color In STEM: Navigating the
  Double Bind in K-12 Education}, p. 183, 2020.

\bibitem{celedon2022fake}
S.~Celed{\'o}n-Pattichis, G.~Kussainova, C.~A. L{\'o}pezLeiva, and M.~S.
  Pattichis, ``“fake it until you make it”: Participation and positioning
  of a bilingual latina student in mathematics and computing,'' \emph{Teachers
  College Record}, vol. 124, no.~5, pp. 186--205, 2022.

\bibitem{yanguas2022middle}
J.~A.~L. Yanguas, ``Middle school students communicating computational
  thinking: A systemic functional linguistics-case study of bilingual,
  collaborative teaching/learning of computer programming in python,'' Ph.D.
  dissertation, The University of New Mexico, 2022.

\bibitem{ffmpeg}
\BIBentryALTinterwordspacing
{ FFmpeg Developers}, ``ffmpeg tool.'' [Online]. Available:
  \url{https://ffmpeg.org/}
\BIBentrySTDinterwordspacing

\bibitem{esakki2021adaptive}
G.~Esakki, A.~S. Panayides, V.~Jalta, and M.~S. Pattichis, ``Adaptive video
  encoding for different video codecs,'' \emph{IEEE Access}, vol.~9, pp.
  68\,720--68\,736, 2021.

\bibitem{ren2016faster}
S.~Ren, K.~He, R.~Girshick, and J.~Sun, ``Faster r-cnn: towards real-time
  object detection with region proposal networks,'' \emph{IEEE transactions on
  pattern analysis and machine intelligence}, vol.~39, no.~6, pp. 1137--1149,
  2016.

\bibitem{make-sense}
P.~Skalski, ``{Make Sense},'' \url{https://github.com/SkalskiP/make-sense/},
  2019.

\bibitem{sravani_thesis}
S.~Teeparthi, ``Long-term video object detection and tracking in collaborative
  learning environments,'' 2021.

\bibitem{decord}
D.~D. M.~L. Community, ``{decord},'' \url{https://github.com/dmlc/decord},
  2022.

\bibitem{jacoby2018context}
A.~R. Jacoby, M.~S. Pattichis, S.~Celed{\'o}n-Pattichis, and C.~L{\'o}pezLeiva,
  ``Context-sensitive human activity classification in collaborative learning
  environments,'' in \emph{2018 IEEE Southwest Symposium on Image Analysis and
  Interpretation (SSIAI)}.\hskip 1em plus 0.5em minus 0.4em\relax IEEE, 2018,
  pp. 1--4.

\end{thebibliography}

\begin{IEEEbiography}[{\includegraphics[width=1in,height=1.25in,clip,keepaspectratio]{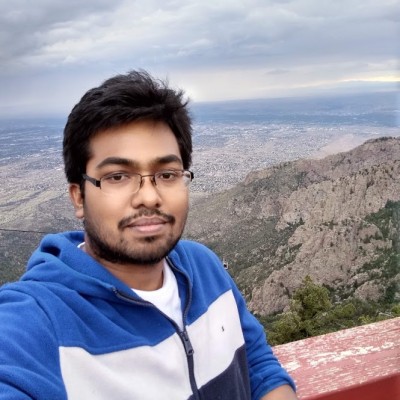}}]{\textbf{Venkatesh Jatla}}
  received his Ph.D. in Electrical and Computer Engineering from the
  University of New Mexico, has a profound background in video
  activity quantification, image processing, machine learning and
  video compression standards. His diverse research experiences,
  including human activity recognition and video compression, are
  well-supported by his work in both academic and industry settings,
  notably at MediaTek and UNM. Jatla's contributions to the field are
  documented through numerous publications in esteemed journals and
  participation in NSF-funded projects, showcasing his technical
  prowess in neural networks, video analysis, and a range of
  programming languages.
\end{IEEEbiography}

\begin{IEEEbiography}[{\includegraphics[width=1in,height=1.25in,clip,keepaspectratio]{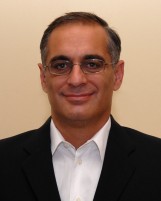}}]{\textbf{Marios S. Pattichis}}
	received the B.Sc. de-
	gree (High Hons. and Special Hons.) in computer
	sciences, the B.A. degree (High Hons.) in math-
	ematics, the M.S. degree in electrical engineering,
	and the Ph.D. degree in computer engineering from
	The University of Texas at Austin, Austin, TX,
	USA, in 1991, 1991, 1993, and 1998, respectively.
	He is currently a Professor and Director of on-
	line programs with the Department of Electrical
	and Computer Engineering at the University of
	New Mexico. At UNM, he is also the Director of the Image and Video
	Processing and Communications Lab (ivPCL). His current research interests
	include digital image and video processing, video communications, dy-
	namically reconfigurable hardware architectures, and biomedical and space
	image-processing applications.
	
	Dr. Pattichis was a fellow of the Center for Collaborative Research and
	Community Engagement, UNM College of Education, from 2019 to 2020.
	He was a recipient of the 2016 Lawton-Ellis and the 2004 Distinguished
	Teaching Awards from the Department of Electrical and Computer Engi-
	neering, UNM. For his development of the digital logic design laboratories
	with UNM, he was recognized by Xilinx Corporation in 2003. He was
	also recognized with the UNM School of Engineering’s Harrison Faculty
	Excellence Award, in 2006.
	
	He was the general chair of the 2008 IEEE Southwest Symposium on
	Image Analysis and Interpretation (SSIAI), general co-chair of the SSIAI,
	in 2020 and 2024. He was also a general chair of the 20th Conference on
	Computer Analysis of Images and Patterns in 2023. He has served as a Senior
	Associate Editor for the IEEE TRANSACTIONS ON IMAGE PROCESS-
	ING and IEEE SIGNAL PROCESSING LETTERS, an Associate Editor for
	IEEE TRANSACTIONS ON IMAGE PROCESSING and IEEE TRANSAC-
	TIONS ON INDUSTRIAL INFORMATICS, and a Guest Associate Editor
	for two additional special issues published in the IEEE TRANSACTIONS
	ON INFORMATION TECHNOLOGY IN BIOMEDICINE, a Special Issue
	published by Teachers College Record, a Special Issue published by the
	IEEE JOURNAL OF BIOMEDICAL AND HEALTH INFORMATICS, and
	a Special Issue published in Biomedical Signal Processing and Control He
	was elected as a fellow of the European Alliance of Medical and Biological
	Engineering and Science (EAMBES) for his contributions to biomedical
	image analysis.
\end{IEEEbiography}

\begin{IEEEbiography}[{\includegraphics[width=1in,height=1.25in,clip,keepaspectratio]{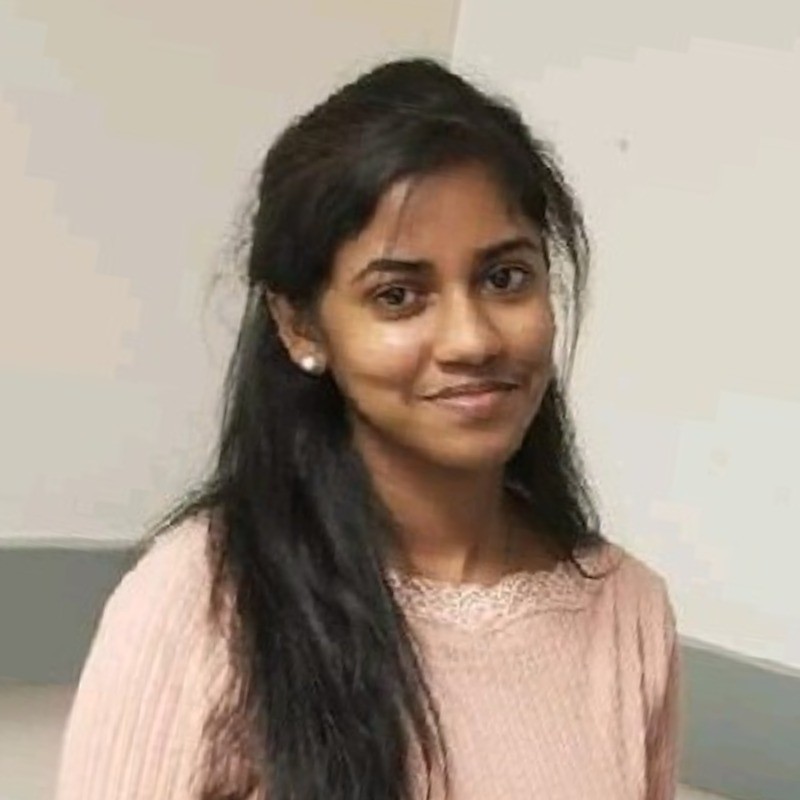}}]{\textbf{Sravani Teeparthi}}
  with a Master of Science in Computer Engineering
  from the University of New Mexico, specializes in image and video
  processing, boasting a CGPA of 4.17/4.0. Her extensive experience
  encompasses roles in data science, machine learning, and data
  engineering across various organizations, including the Fralin
  Biomedical Research Institute and Cadent. Sravani has developed
  innovative machine learning models for computational neuroscience
  and has contributed to the advancement of data pipelines and
  analytics. Her research, recognized for excellence in video object
  detection and tracking within collaborative learning environments,
  has been published in notable conferences and journals.
\end{IEEEbiography}

\begin{IEEEbiography}[{\includegraphics[width=1in,height=1.25in,clip,keepaspectratio]{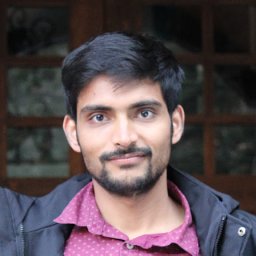}}]{\textbf{Ugesh Egala}}
  received his M.S. degree in Electrical and Computer
  Engineering from the University of New Mexico, Albuquerque, USA,
  specializing in screen activity quantification in collaborative
  learning environments. His academic journey is marked by a strong
  focus on image processing, computer vision, and machine
  learning. Ugesh's research contributions, aimed at enhancing
  learning experiences and analyzing nonverbal communication patterns,
  have led to publications in recognized journals. His professional
  experience spans both academic research and software engineering
  roles, showcasing a commitment to advancing educational technologies
  and collaborative learning analysis.
\end{IEEEbiography}

\begin{IEEEbiography}[{\includegraphics[width=1in,height=1.25in,clip,keepaspectratio]{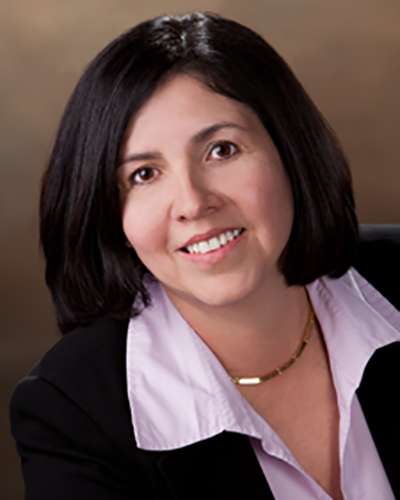}}]{\textbf{SYLVIA CELEDÓN-PATTICHIS}}
	is a professor of
	bilingual/bicultural education in the Department of
	Curriculum and Instruction. She recently served as
	senior associate dean for research and community
	engagement and director of the Center for Collab-
	orative Research and Community Engagement in
	the College of Education at The University of New
	Mexico.
	
	Celedón-Pattichis prepares elementary pre-
	service teachers in the bilingual/ESL cohort to
	teach mathematics and teaches graduate level courses in bilingual education.
	She taught mathematics at Rio Grande City High School in Rio Grande City,
	Texas for four years. Her research interests focus on studying linguistic and
	cultural influences on the teaching and learning of mathematics, particularly
	with bilingual students. She was a co-principal investigator (PI) of the
	National Science Foundation (NSF)-funded Center for the Mathematics
	Education of Latinos/as (CEMELA). She is currently a lead-PI or co-PI of
	three NSF-funded projects that broaden the participation of Latinx students
	in mathematics and computer programming in rural and urban contexts.
	
	She serves as a National Advisory Board member of several NSF-funded
	projects and as an Editorial Board member of the Bilingual Research
	Journal, Journal of Latinos and Education and Teachers College Record.
	Her current work is a special issue on Teaching and Learning Mathematics
	and Computing in Multilingual Contexts through Teachers College Record.
	She co-edited three books published by the National Council of Teachers of
	Mathematics titled Access and Equity: Promoting High Quality Mathematics
	in Grades PreK-2 and Grades 3-5 and Beyond Good Teaching: Advancing
	Mathematics Education for ELLs.
	
	Celedón-Pattichis was a recipient of the Innovation in Research on
	Diversity in Teacher Education Award from the American Educational
	Research Association, and the 2011 Senior Scholar Reviewer Award from
	the National Association of Bilingual Education. She was also a recipient
	of the Regents Lectureship Award, the Faculty of Color Research Award,
	Chester C. Travelstead Endowed Faculty Award, and the Faculty of Color
	Mentoring Award to recognize her research, teaching, and service at The
	University of New Mexico. The accomplishments she is most proud of are
	her two daughters, Rebecca and Antonia Pattichis.
\end{IEEEbiography}

\EOD

\end{document}